\documentclass{article}

\usepackage{iclr2027_conference,times}

\usepackage{amsmath}
\usepackage{amssymb}
\usepackage{array}
\usepackage{booktabs}
\usepackage{graphicx}
\usepackage{flafter}
\usepackage{microtype}
\usepackage{multirow}
\usepackage{subcaption}
\usepackage{xspace}
\usepackage{tcolorbox}
\tcbuselibrary{breakable}
\usepackage[table]{xcolor}
\usepackage{url}
\usepackage{hyperref}

\definecolor{CitationGreen}{RGB}{0,150,55}
\definecolor{ReferenceRed}{RGB}{190,45,45}
\hypersetup{
  colorlinks=true,
  breaklinks=true,
  citecolor=CitationGreen,
  linkcolor=ReferenceRed,
  urlcolor=ReferenceRed,
  pdftitle={ChartJudgeBench: Evaluating LMM Judges for Chart-to-Code Generation},
  pdfauthor={Lijian Wu, Henry Hengyuan Zhao, Zijian Zhang, Jiahao Tang, Jiajun Wu, Alex Jinpeng Wang}
}
\newcolumntype{M}{>{\raggedright\arraybackslash}l}

\definecolor{MyDarkBlue}{RGB}{0,0,130}

\newcommand{\benchmark}{\textsc{ChartJudgeBench}\xspace}
\newcommand{\enquote}[1]{``#1''}
\newcommand{\finding}[1]{%
  \par\vspace{0.2ex}\noindent\textbf{#1}\par\vspace{-0.4ex}%
}

\makeatletter
\newenvironment{supplementtable}
  {\par\addvspace{1.2\baselineskip}%
   \noindent\begin{minipage}{\linewidth}%
   \def\@captype{table}\centering}
  {\end{minipage}%
   \par\addvspace{\baselineskip}}
\makeatother

\newif\ifarxivversion
\arxivversiontrue

\title{\benchmark: Evaluating LMM Judges for Chart-to-Code Generation}

\author{%
  \textbf{Lijian Wu}\textsuperscript{1,*}
  \quad
  \textbf{Henry Hengyuan Zhao}\textsuperscript{2,*,\textdagger}
  \quad
  \textbf{Zijian Zhang}\textsuperscript{1}
  \\[0.5ex]
  \textbf{Jiahao Tang}\textsuperscript{1}
  \quad
  \textbf{Jiajun Wu}\textsuperscript{1}
  \quad
  \textbf{Alex Jinpeng Wang}\textsuperscript{1,\textdaggerdbl}
  \\[0.7ex]
  \normalfont
  \textsuperscript{1}\textbf{CSU-JPG}, Central South University
  \quad
  \textsuperscript{2}National University of Singapore
  \\[0.3ex]
  \normalfont\textsuperscript{*}Equal contribution.
  \quad
  \textsuperscript{\textdagger}Project lead.
  \quad
  \textsuperscript{\textdaggerdbl}Corresponding author.
}

\ifarxivversion
  \iclrfinalcopy
\fi

\begin{document}

\maketitle
\ifarxivversion
  \lhead{Preprint. Under review.}
\fi

\begin{abstract}

Building strong chart-to-code systems increasingly relies on reinforcement learning, whose effectiveness depends critically on the quality of the reward signal. Large Multimodal Models (LMMs) play a natural critical role in jointly assessing chart visual appearance and task requirements. They are therefore increasingly used as visual critics and reward models, yet their reliability as judges remains largely unexplored.
To this end, we introduce \benchmark{}, a diagnostic vision-language benchmark for assessing LMM judges in chart-to-code workflows. It includes 1,003 Chart Perception Alignment (CPA) instances for pairwise chart comparison and 650 Chart Reasoning Judgment (CRJ) instances for binary \textit{Accept/Reject} verification in \textit{Chart Reproduction} and \textit{Chart Editing}. Together, these tasks emulate the core judging decisions required in agentic refinement and RL-based chart optimization.
Our evaluation of strong LMMs reveals four systematic limitations: (i) positional bias in pairwise comparison, (ii) a strong tendency to overpredict \textit{Accept}, (iii) difficulty in matching visual styles and aesthetics, and (iv) an unexpected leniency bias in RL-trained models. These findings show that current LMM judges require explicit reliability validation before being used as critics or reward models in chart-to-code optimization. The code and data are available on \href{https://csu-jpg.github.io/chartjudge/}{\textcolor{ReferenceRed}{ChartJudgeBench}}.

\end{abstract}

\section{Introduction}
\label{sec:intro}

Large Multimodal Models (LMMs) have substantially advanced visual understanding and code generation \cite{gemini3.1,gpt5.5}, making chart-to-code generation increasingly practical \cite{chartcoder}. In this task, models try to reproduce a chart to executable Python code, supporting automatic chart reconstruction and modification \cite{chart2code,chartmimic,chartcoder}. As a result, \textit{chart judgment} has become a critical component for training models with explicit scalar rewards.

\begin{figure}[t]
  \centering
  \includegraphics[width=0.68\linewidth]{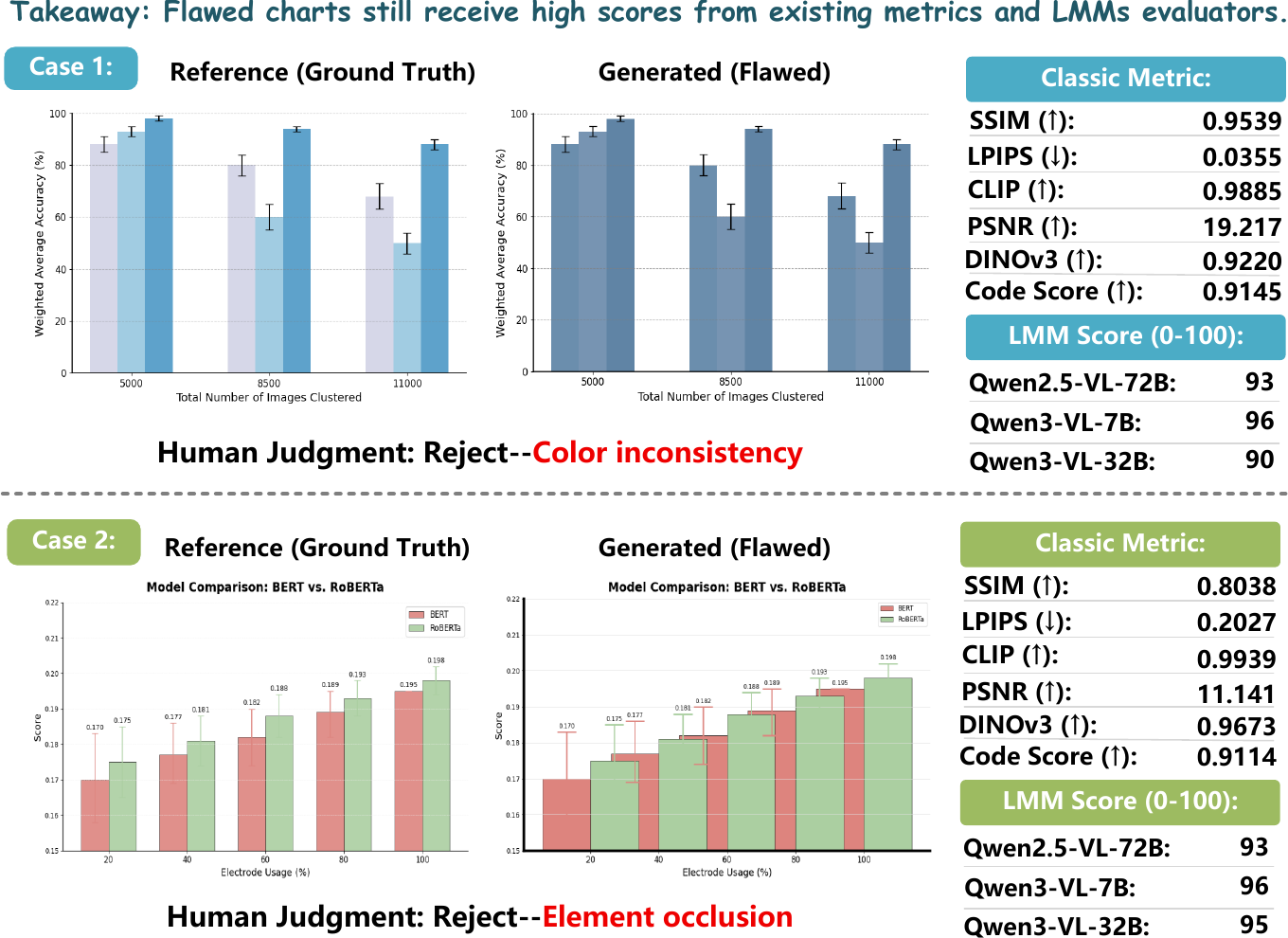}
  \vspace{-0.6em}
  \caption{
  Comparison of chart quality judgment signals.
  Despite obvious visual flaws, similarity metrics
  (PSNR $\uparrow$, SSIM $\uparrow$, LPIPS $\downarrow$, CLIP $\uparrow$, DINOv3 $\uparrow$),
  rule-based metrics, and LMM evaluators tend to give a high score, failing to identify incorrect charts.
  }
  \vspace{-0.7em}
  \label{fig:teaser}
\end{figure}

Existing chart-to-code evaluation and optimization primarily relies on code-level \cite{chartmimic,mmrecoder} and image-level metrics \cite{zhao2025vincicoderunifyingmultimodalcode,chartir}. Code-level metrics compare program structures or extracted chart attributes
\cite{chartmimic,mmrecoder}, but code similarity does not directly reflect the rendered result. Different programs can produce visually equivalent charts, while small code changes can introduce severe visual errors. Image-level metrics, including PSNR \cite{psnr}, SSIM \cite{wang2004ssim},
LPIPS \cite{zhang2018lpips}, CLIP similarity \cite{radford2021clip},
and DINOv3 similarity \cite{simeoni2025dinov3}, mainly measure global image similarity. They often overlook the structured and localized errors that determine chart correctness. As shown in Fig.~\ref{fig:teaser}, clearly flawed charts can still receive high scores from both code-level and image-level metrics.

\begin{figure}[t]
\centering
\includegraphics[width=0.94\linewidth]{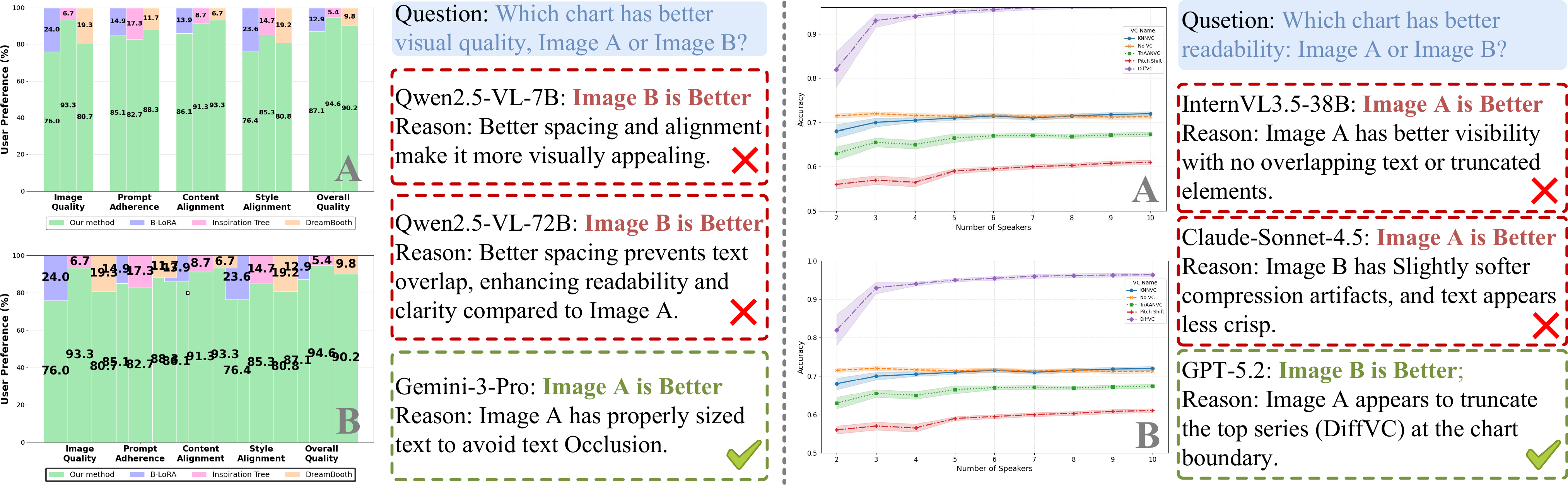}
\vspace{-0.6em}
\caption{Open-source LMMs often fail to select the correct chart even when visual differences are clear.}
\label{fig:teaser2}
\vspace{-0.7em}
\end{figure}

LMMs are a promising choice for chart judgement because they can jointly reason about visual appearance, chart semantics, and task requirements. Recent chart-to-code systems have therefore used LMMs for iterative refinement and reward-guided optimization
\cite{li2025metalmultiagentframeworkchart,
namgoong-etal-2025-amace,
zhang2025boostingcharttocodegenerationmllm,
chen2025breakingsftplateaumultimodal}.
However, strong generation and understanding abilities do not guarantee reliable judgment. As shown in Fig.~\ref{fig:teaser}, LMM evaluators may assign high scores to flawed charts. Fig.~\ref{fig:teaser2} further shows that open-source LMM judges often fail to select the better chart. This raises a key question: \textit{Can current LMMs evaluate chart-to-code outputs accurately, consistently, and strictly?}

Despite the growing use of LMM judges \cite{chen2024mllmasajudgeassessingmultimodalllmasajudge}, most chart benchmarks such as ChartMimic~\cite{chartmimic}, Chart2Code~\cite{chart2code}, ChartM$^3$~\cite{chartm3}, ChartEdit~\cite{chartedit}, CharXiv~\cite{wang2024charxiv}, and ChartQAPro~\cite{chartqapro}, mainly focus on evaluating chart generation and understanding. The ability of LMMs to serve as visual judges for chart-to-code systems remains an open challenge. In particular, it is unclear whether they can reliably compare candidate charts and reject outputs with visual or semantic errors.

To fill this gap, we introduce \benchmark{}, a diagnostic vision-language benchmark for evaluating LMM judges in chart-to-code workflows. \benchmark{} includes two complementary tasks. \textbf{Chart Perception Alignment (CPA)} evaluates whether an LMM can select the better chart from two candidates. It tests the basic visual judgement on charts. We empirically summarize 3 major dimensions: Data Encoding, Visual Effects, and Layout Structure with 7 fine-grained categories (see Fig. \ref{fig:diagram}). We specially disentangle each fine-grained error flaw, making our benchmark easily track the detailed error patterns of tested LMMs. Additionally, we design \textbf{Chart Reasoning Judgment (CRJ)} based on two classic chart-to-code tasks: \textit{Chart Reproduction} and \textit{Chart Editing}. Each task has a very long input with full Python code and visual charts. Together, these two tasks making our evaluation comprehensive and discriminative.

After the evaluation of 26 proprietary and open-source LMMs, we found four systematic limitations:
\textbf{(i)} many LMMs exhibit positional bias;
\textbf{(ii)} open-source models show a strong tendency to overpredict \textit{Accept};
\textbf{(iii)} persistent weakness in style and aesthetic judgment, even among strong LMMs; and
\textbf{(iv)} limited gains from RL-post-trained models alongside increased leniency bias.
Additionally, we conduct a detailed error analysis to characterize agent failure patterns, demonstrating that our disentangled task design enables fine-grained and discriminative diagnosis.

\section{Related Work}

\noindent\textbf{Chart Understanding and Chart-to-Code Generation.}
Chart understanding benchmarks, such as ChartQA~\cite{chartqa}, CharXiv~\cite{wang2024charxiv}, and ChartMind~\cite{chartmind}, evaluate whether models can answer questions and reason over chart images. Chart-to-code benchmarks, in comparison, evaluate whether multimodal models can reproduce or edit charts as executable code. ChartMimic~\cite{chartmimic} evaluates chart reproduction from visual references, while ChartEdit~\cite{chartedit}, ChartEditor~\cite{ChartEditor}, and ChartM$^3$~\cite{chartm3} study chart editing under textual or multimodal instructions. ChartCoder~\cite{chartcoder}, Chart2Code53~\cite{chart2code53}, and DCG-Bench~\cite{DCG} further extend chart-to-code generation through larger training data or more dynamic visualization settings. Recent systems further incorporate agentic pipelines and learning-based optimization to generate, inspect, and iteratively refine candidate outputs \cite{li2025metalmultiagentframeworkchart,namgoong-etal-2025-amace,tan2025chartmasteradvancingcharttocodegeneration}. These benchmarks characterize the capabilities of chart interpreters, generators, and editors, but do not directly evaluate the reliability of LMM judges used to select or verify generated chart outputs.









\begin{figure}[t]
  \centering
  \includegraphics[width=\linewidth]{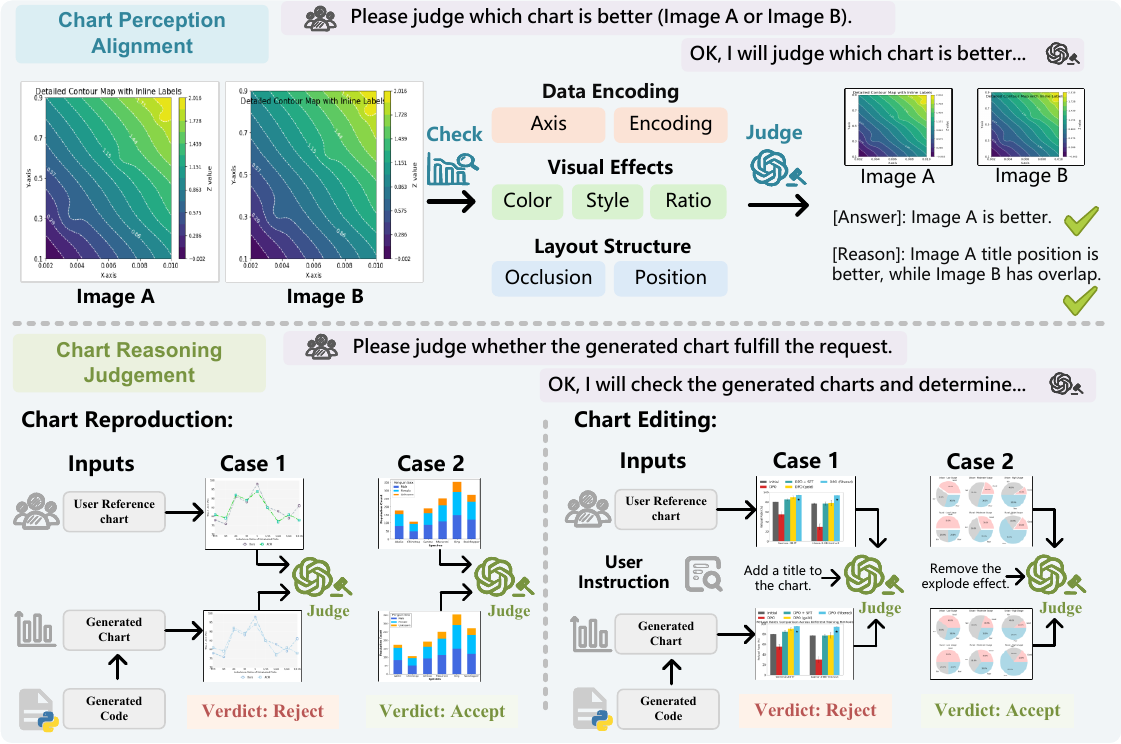}
  \caption{Illustration of CPA and CRJ tasks in \benchmark{}.}
  \label{fig:diagram}
\end{figure}

\noindent\textbf{Multimodal and Visualization Judges.}
LMMs are increasingly adopted as automatic judges. MLLM-as-a-Judge~\cite{chen2024mllmasajudgeassessingmultimodalllmasajudge} studies general multimodal evaluation, while LLaVA-Critic~\cite{xiong2025llavacriticlearningevaluatemultimodal} and LLaVA-Critic-R1~\cite{llavacriticr1} develop open-source multimodal evaluators for general vision-language tasks. Recent efforts, such as VisJudge-Bench~\cite{xie2026visjudgebenchaestheticsqualityassessment}, further investigate visual judgment through subjective annotations and by measuring the agreement between model preferences and human preference scores. Visual-ERM~\cite{liu2026visualermrewardmodelingvisual} propose to judge chart, table, and SVG reproduction tasks. It primarily focuses on creating the training data, and the diagnostic benchmark lacks the fine-grained chart flaws category to disentangle the errors. In this paper, \benchmark{} first considers real-world chart-to-code workflows and divide the key chart judges into 3 major dimension with 7 criteria as shown in Fig. \ref{fig:diagram} and then creates tasks from the perspective of RL-based optimization requirements. These designs enable our benchmark to provide a comprehensive and reliable diagnostic testbed for future development.

\section{ChartJudgeBench}
\label{sec:benchmark}

\subsection{Task Definition}
\label{subsec:task_definition}

To evaluate the capability of current LMMs to judge chart-to-code outputs, we design \benchmark{}. We denote a chart-to-code task as $\mathcal{T}=(\mathcal{R}, \mathcal{I}, \mathcal{D})$, where $\mathcal{R}$ is the reference chart, $\mathcal{I}$ is an optional text instruction, and $\mathcal{D}$ is an optional data source; a generator produces code rendered as $\mathcal{C}_{gen}$, and an LMM judge $f_{judge}$ evaluates the rendered chart.

In our benchmark, we design two types of chart judgment: i) basic perception judgment, selecting the better chart candidate, ii) advanced query-aware chart verification, which decides whether a generated chart satisfies the task requirements (see Fig. \ref{fig:diagram}). These two forms instantiate Chart Perception Alignment and Chart Reasoning Judgment, respectively.

\vspace{4pt}
\noindent \textbf{Chart Perception Alignment (CPA).}
In this task, we would like to test whether LMMs have the basic chart perception ability to select the better chart from two candidates. Given two candidate charts $\mathcal{C}_A$ and $\mathcal{C}_B$, the judge must select the chart that better aligns with human judgment:
\[
    \mathcal{J}_{cpa}=f_{judge}(\mathcal{C}_A,\mathcal{C}_B), \quad
    \mathcal{J}_{cpa}\in\{\mathcal{C}_A,\mathcal{C}_B\}.
\]
To make the evaluation comprehensive, we systematically conducted a preliminary study and derived 7 controlled criteria across 3 dimensions (see Fig. \ref{fig:diagram}): \textit{Visual Effects} (color consistency, visual style, aspect ratio), \textit{Layout Structure} (component positioning, text occlusion), and \textit{Data Fidelity} (data encoding, axis scaling). Such a design comes from the realistic chart characteristics and makes the evaluation range cover most of chart normal flaws.

\vspace{4pt}
\noindent \textbf{Chart Reasoning Judgment (CRJ).}
In this task, we draw the design from the realistic RL training scenario. We focus not only on the basic chart quality but to judge whether a generated chart \textbf{satisfies} the task requirements. Given a rendered chart $\mathcal{C}_{gen}$ and its task condition, the judge must decide whether the output satisfies the required chart-to-code task:
\[
    \mathcal{J}_{crj}=f_{judge}(\mathcal{T},\mathcal{C}_{gen}), \quad
    \mathcal{J}_{crj}\in\{\text{Accept},\text{Reject}\}.
\]
We instantiate CRJ in two classic chart-to-code tasks: \textit{Chart Reproduction}, where the judge verifies whether $\mathcal{C}_{gen}$ matches the reference $\mathcal{R}$, and \textit{Chart Editing}, where the judge verifies whether $\mathcal{C}_{gen}$ follows the edit instruction $\mathcal{I}$ while preserving unchanged elements. CRJ diagnoses whether LMM judges can apply strict task-conditioned verification, including visual errors and unintended changes to non-target regions.



\vspace{-0.05in}
\subsection{Data Sources}
\label{subsec:data_sources}

We construct the source chart pool from two complementary sources. First, we collect 315 raw charts from arXiv Computer Science papers published between January and July 2025 under CC BY 4.0 licenses. For these charts, we reconstruct executable Python scripts using Gemini-3-Pro followed by human verification. Second, we curate 494 high-quality chart-code pairs from existing chart-to-code benchmarks, including 93 from ChartMimic~\cite{chartmimic}, 179 from Chart2Code~\cite{chart2code}, and 222 from ChartEdit~\cite{chartedit}. Together, these sources provide both recent real-world charts and complex benchmark instances for constructing \benchmark{}.

 \vspace{-0.05in}
\subsection{Benchmark Construction}
\label{subsec:benchmark_construction}

\suppressfloats[t]
\begin{figure}[t]
    \centering

    \begin{subfigure}[t]{\linewidth}
        \centering
        \input{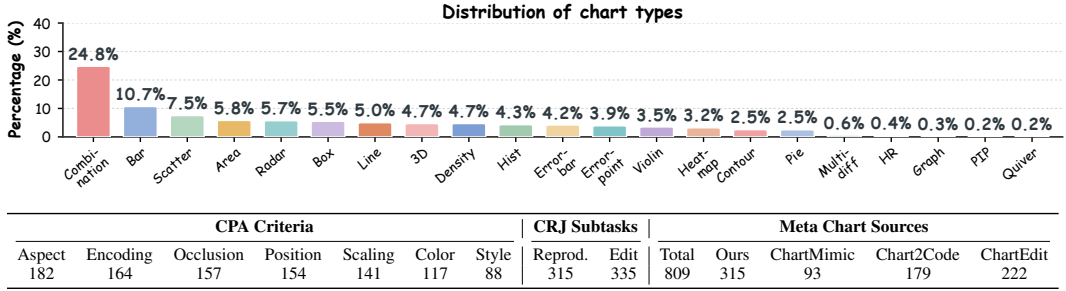}
        \label{fig:chart_type_distribution}
    \end{subfigure}

    \vspace{0.12in}

    \begin{subfigure}[t]{\linewidth}
        \centering

\resizebox{\linewidth}{!}{%
\begin{tabular}{ccccccc|cc|ccccc}
    \toprule
    \multicolumn{7}{c|}{\textbf{CPA Criteria}}
    & \multicolumn{2}{c|}{\textbf{CRJ Subtasks}}
    & \multicolumn{5}{c}{\textbf{Meta Chart Sources}}
    \\

    \cmidrule(lr){1-7}
    \cmidrule(lr){8-9}
    \cmidrule(lr){10-14}

     Aspect
    & Encoding
    & Occlusion
    & Position
    & Scaling
    & Color
    & Style
    & Reprod.
    & Edit
    & Total
    & Ours
    & ChartMimic
    & Chart2Code
    & ChartEdit
    \\


    182
    & 164
    & 157
    & 154
    & 141
    & 117
    & 88
    & 315
    & 335
    & 809
    & 315
    & 93
    & 179
    & 222
    \\

    \bottomrule
\end{tabular}
}

\vspace{-0.1in}

        \label{fig:data_statistics_table}
    \end{subfigure}
    \caption{
        Overview of the benchmark data statistics. 
        \textbf{Upper:} Distribution of chart types in our \benchmark{}.
        \textbf{Lower:} Data statistics across task category, 7 criteria, and sources.
    }
    \label{fig:data_statistics}
\end{figure}

\noindent \textbf{CPA Construction.}
We derive the CPA taxonomy by analyzing flawed charts produced by strong LMMs on existing chart-to-code benchmarks, summarizing recurring errors into the 3 dimensions and 7 criteria. Since real model outputs often contain multiple co-occurring errors, they are not ideal for diagnosing LMM judge ability along individual chart-quality dimensions. We therefore specifically use these recurring errors as controlled diagnostic factors and construct CPA cases through targeted perturbations.

Using 809 high-quality chart scripts from the source chart pool as seeds, we prompt Gemini-3-Pro \cite{gemini3} with rule-based instructions to perturb one target criterion at a time, producing 5,180 original perturbed chart pairs. Five expert annotators then filter out ambiguous candidates or those containing multiple defects. After filtering, we retain 1,003 CPA pairs with clear relative preferences. Detailed category distributions, validation examples and prompts are provided in the supplementary material.

\vspace{4pt}
\noindent \textbf{CRJ Construction.}
We construct CRJ for two representative chart-to-code tasks, \textit{Chart Reproduction} and \textit{Chart Editing}, with generated charts labeled by 5 expert annotators.

\textit{Chart Reproduction.}
We construct 315 reproduction cases by combining 172 charts from Chart2Code \cite{chart2code} and ChartMimic \cite{chartmimic} with 143 newly collected charts. For each chart, we generate code using strong proprietary and open-source LMMs, including Gemini-3-Pro, GPT-5.2, and Qwen2.5-VL-72B. A case is labeled as Accept if the rendered chart faithfully matches the reference chart; otherwise, it is labeled as Reject. The final reproduction subset contains 156 Accept cases and 159 Reject cases.

\textit{Chart Editing.}
We construct 335 editing cases by combining 222 high-quality instances from ChartEdit~\cite{chartedit} with 113 newly collected charts. For the newly collected charts, we construct editing instructions and use LMMs to generate candidates. A case is labeled as Accept only if the requested edit is correctly applied and all non-target elements remain preserved; otherwise, it is labeled as Reject. The final subset contains 163 Accept cases and 172 Reject cases. Details of instruction generation and candidate construction are provided in the supplementary material.

\vspace{4pt}
\noindent \textbf{Data Statistics.}
Overall, \benchmark{} contains 1,653 instances, including 1,003 CPA pairs and 650 CRJ cases. Fig.~\ref{fig:data_statistics} summarizes the benchmark statistics and chart-type distribution. CRJ contains 315 reproduction and 335 editing cases with a near-balanced Accept/Reject split, while covering diverse chart types.

\begin{table}[t]
    \centering

    \begingroup

    \renewcommand{\arraystretch}{0.96}



    \resizebox{0.92\linewidth}{!}{%
    \begin{tabular}{M|ccc|cc|cc|c}

        \toprule

        &
        \multicolumn{3}{c|}{\textbf{Visual Effects}}
        &
        \multicolumn{2}{c|}{\textbf{Data Fidelity}}
        &
        \multicolumn{2}{c|}{\textbf{Layout Structure}}
        &
        \\

        \cmidrule(lr){2-4}
        \cmidrule(lr){5-6}
        \cmidrule(lr){7-8}

        \multirow{-2}{*}{\normalsize\textbf{Model}}
        & \textbf{Color}
        & \textbf{Style}
        & \textbf{Aspect}
        & \textbf{Scaling}
        & \textbf{Encoding}
        & \textbf{Position}
        & \textbf{Occlusion}
        & \multirow{-2}{*}{\textbf{Avg.}}
        \\

        \midrule

        \multicolumn{9}{c}{
            \textit{\textbf{Proprietary}}
        }
        \\[-0.5pt]

        \textbf{Gemini-3-Pro}
        & \textbf{98.3}
        & \textbf{98.9}
        & \textbf{89.0}
        & \textbf{92.9}
        & \textbf{98.8}
        & \textbf{90.9}
        & \textbf{96.2}
        & \textbf{95.0}
        \\

        GPT-5.2
        & 89.7
        & 79.5
        & 66.5
        & 79.4
        & 82.9
        & 82.5
        & 77.1
        & 79.7
        \\

        Seed1.6-VL
        & 91.5
        & 68.2
        & 50.0
        & 44.7
        & 50.6
        & 77.3
        & 69.4
        & 64.5
        \\

        Claude-Sonnet-4.5
        & 66.7
        & 40.9
        & 59.3
        & 51.8
        & 59.2
        & 40.3
        & 57.3
        & 53.6
        \\

        \midrule

        \multicolumn{9}{c}{
            \textit{\textbf{Open-Source (thinking)}}
        }
        \\[-0.5pt]

        MiMo-VL-SFT
        & 80.3
        & 37.5
        & 29.1
        & 27.0
        & 61.0
        & 40.3
        & 46.5
        & 46.0
        \\

        Qwen3-VL-30B
        & 58.1
        & 28.4
        & 18.7
        & 22.0
        & 37.8
        & 29.9
        & 29.3
        & 32.0
        \\

        MiMo-VL-RL
        & 65.8
        & 20.4
        & 14.3
        & 7.8
        & 34.8
        & 23.4
        & 27.4
        & 27.7
        \\

        LLaVA-Critic-R1
        & 7.7
        & 9.1
        & 2.2
        & 9.9
        & 15.8
        & 3.9
        & 5.7
        & 7.8
        \\

        ThinkLite-VL-7B
        & 12.0
        & 2.3
        & 0.5
        & 3.5
        & 14.0
        & 1.3
        & 3.2
        & 5.3
        \\

        \midrule

        \multicolumn{9}{c}{
            \textit{\textbf{Open-Source (non-thinking)}}
        }
        \\[-0.5pt]

        Qwen2-VL-7B
        & 15.4
        & 10.2
        & 11.5
        & 16.3
        & 29.3
        & 4.5
        & 13.4
        & 14.4
        \\

        Qwen2-VL-72B
        & 36.8
        & 37.5
        & 23.6
        & 25.5
        & 42.7
        & 36.4
        & 43.3
        & 35.1
        \\

        DeepSeek-VL
        & 0.9
        & 0.0
        & 0.0
        & 0.7
        & 0.0
        & 0.0
        & 0.0
        & 0.2
        \\

        Qwen2.5-VL-7B
        & 8.6
        & 3.4
        & 1.7
        & 6.4
        & 8.5
        & 1.9
        & 3.8
        & 4.9
        \\

        Qwen2.5-VL-72B
        & 58.1
        & 58.0
        & 54.9
        & 39.0
        & 51.8
        & 77.9
        & 52.9
        & 56.1
        \\

        GLM-4V
        & 58.8
        & 29.5
        & 20.3
        & 20.0
        & 43.1
        & 29.2
        & 23.9
        & 32.1
        \\

        MiMo-VL-SFT (NoThink)
        & 74.4
        & 39.8
        & 37.9
        & 27.7
        & 37.2
        & 50.6
        & 49.7
        & 45.3
        \\

        MiMo-VL-RL (NoThink)
        & 60.7
        & 33.0
        & 36.3
        & 26.2
        & 31.1
        & 53.9
        & 48.4
        & 41.4
        \\

        Kimi-VL
        & 47.0
        & 12.5
        & 46.7
        & 34.0
        & 40.2
        & 26.0
        & 33.1
        & 34.2
        \\

        InternVL2.5-38B
        & 33.3
        & 14.8
        & 9.3
        & 8.5
        & 21.3
        & 27.9
        & 31.9
        & 21.0
        \\

        InternVL2.5-8B
        & 26.5
        & 15.9
        & 9.3
        & 7.8
        & 18.9
        & 14.3
        & 20.4
        & 16.2
        \\

        Molmo
        & 7.7
        & 8.0
        & 6.6
        & 8.5
        & 4.9
        & 6.5
        & 8.3
        & 7.2
        \\

        Molmo2
        & 34.2
        & 13.6
        & 16.5
        & 27.7
        & 34.8
        & 20.8
        & 26.1
        & 24.8
        \\

        Qwen3-VL-8B
        & 48.7
        & 33.0
        & 35.2
        & 29.8
        & 35.4
        & 57.8
        & 41.4
        & 40.2
        \\

        Qwen3-VL-32B
        & 93.2
        & 55.7
        & 69.8
        & 57.5
        & 65.2
        & 77.3
        & 63.7
        & 68.9
        \\

        InternVL3.5-8B
        & 33.3
        & 26.1
        & 13.2
        & 5.7
        & 9.8
        & 13.6
        & 20.4
        & 17.4
        \\

        InternVL3.5-38B
        & 72.7
        & 20.4
        & 28.6
        & 41.1
        & 75.6
        & 40.3
        & 50.3
        & 47.0
        \\

        \midrule

        \textbf{Average}
        & \textbf{49.2}
        & \textbf{30.6}
        & \textbf{28.9}
        & \textbf{27.7}
        & \textbf{38.6}
        & \textbf{35.7}
        & \textbf{36.3}
        & \textbf{35.3}
        \\

        \bottomrule

    \end{tabular}
    }

    \caption{
        Results on Chart Perception Alignment.
        Values are accuracy (\%) across 7 sub-dimensions
        and 3 categories.
    }
    \vspace{-0.15in}
    \label{tab:acc_metrics}

    \endgroup
    
\end{table}

\section{Experiments}

\textbf{Models.} We evaluate 26 LMMs, encompassing both proprietary and open-source models. 
Proprietary models: Gemini-3-Pro \cite{gemini3}, GPT-5.2 \cite{gpt5.2}, Seed1.6-VL \cite{seed1.6}, and Claude-Sonnet-4.5 \cite{Claude4.5}. Open-source models with total parameters spanning from 7B to 72B: (1) Thinking models, includes MiMo-VL-7B-SFT \cite{mimo}, MiMo-VL-7B-RL \cite{mimo}, Qwen3-VL-30B \cite{qwen3vl}, ThinkLite-VL-7B \cite{thinklite}, and LLaVA-Critic-R1 \cite{llavacriticr1}; (2) Non-thinking models, includes the Qwen series (Qwen2-VL-7B, Qwen2.5-VL-72B, Qwen2.5-VL-7B, Qwen3-VL 8B/32B), the InternVL series (InternVL2.5 \cite{internvl2.5} 8B/38B, InternVL3.5 \cite{internvl3_5} 8B/38B), Molmo \cite{molmo, molmo2}, Kimi-VL \cite{kimiteam2025kimivltechnicalreport}, DeepSeek-VL \cite{deepseekvl}, and ablation versions MiMo-VL-7B-SFT (NoThink) and MiMo-VL-7B-RL (NoThink).

\noindent\textbf{Evaluation Protocol.}
For each test instance, the model is prompted to provide a judgment with a brief rationale. We set the decoding temperature to 0.1 across all experiments. \textbf{For CPA, we specially design a two-pass position-swapping protocol}: each chart pair is evaluated twice with reversed order, and a prediction is counted as correct only if the model consistently selects the superior chart. It makes our evaluation less sensitive to the option order. For CRJ, models predict an Accept or Reject label for each generated chart.
\vspace{-0.05in}
\subsection{Main Results}

\subsubsection{(i) Chart Perception Alignment Results.}

CPA evaluates the basic chart perception ability of existing LMMs with our specially designed two-pass position-swapping protocol.

\finding{Finding 1: LMMs struggle with fine-grained pairwise chart comparison, especially among open-source models.}

As shown in Tab.~\ref{tab:acc_metrics}, proprietary models substantially outperform open-source models under the strict two-pass setting. Gemini-3-Pro ranks first with an average accuracy of 95.0\%, followed by GPT-5.2 at 79.7\%. In contrast, most open-source models struggle with subtle visual differences, with an accuracy below 50\%. Chart-type results in supplementary material further show that spatially complex charts, such as 3D and radar charts, are more challenging.

\begin{figure}[t]
    \centering
    \includegraphics[width=\linewidth]{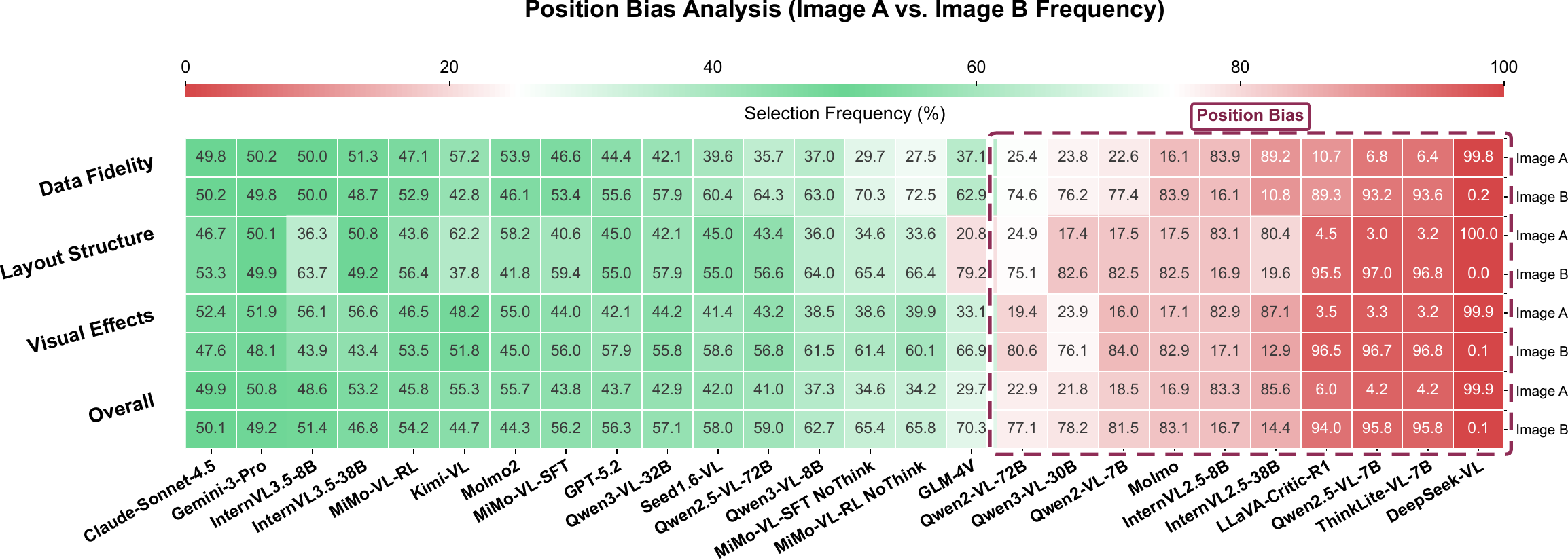}
    \caption{Positional preference in CPA. Many models exhibit a strong tendency to favor a fixed image position. (Finding 2)}
    \label{fig:heatmap_proportional_bias}
\end{figure}


\finding{Finding 2: When fine-grained visual discrimination is insufficient, many models fall back to positional preferences.}

The position-swapping protocol further reveals strong positional shortcuts in many LMM judges. As shown in Fig.~\ref{fig:heatmap_proportional_bias}, some proprietary models maintain balanced preferences across swapped positions, such as Claude-Sonnet-4.5 (49.9\% vs. 50.1\%). In contrast, several open-source models exhibit extreme skew: DeepSeek-VL selects the first position in 99.9\% of cases, while ThinkLite-VL-7B and LLaVA-Critic-R1 strongly favor the second position. These results suggest that \textbf{when chart differences are hard to perceive, models often fall back to positional shortcuts}, making single-pass pairwise evaluation an unreliable estimate of chart judgment ability.

\subsubsection{(ii) Chart Reasoning Judgment Results.}

\begin{table}[t]
    \centering
    \vspace{-0.12in}

    \begingroup




    \resizebox{0.94\linewidth}{!}{
    \begin{tabular}{M|ccccc|ccccc}

        \toprule

        &
        \multicolumn{5}{c|}{\textbf{ChartEdit Task}}
        &
        \multicolumn{5}{c}{\textbf{Reproduction Task}}
        \\

        \cmidrule(lr){2-6}
        \cmidrule(l){7-11}

        \multirow{-2}{*}{\normalsize\textbf{Model}}
        & \textbf{Acc.} $\uparrow$
        & \textbf{F1} $\uparrow$
        & \textbf{Prec.} $\uparrow$
        & \textbf{Recall} $\uparrow$
        & \textbf{Spec.} $\uparrow$
        & \textbf{Acc.} $\uparrow$
        & \textbf{F1} $\uparrow$
        & \textbf{Prec.} $\uparrow$
        & \textbf{Recall} $\uparrow$
        & \textbf{Spec.} $\uparrow$
        \\

        \midrule


        \multicolumn{11}{c}{
            \textit{\textbf{Proprietary}}
        }
        \\[-0.5pt]

        Gemini-3-Pro
        & \textbf{79.1}
        & \textbf{81.6}
        & 71.4
        & 95.1
        & 64.0
        & \textbf{75.6}
        & \textbf{79.1}
        & 68.5
        & 93.6
        & 57.9
        \\

        GPT-5.2
        & \textbf{79.1}
        & 81.3
        & \textbf{73.7}
        & 90.7
        & 69.6
        & 74.3
        & 71.4
        & 79.5
        & 64.7
        & 83.7
        \\

        Seed1.6-VL
        & 70.8
        & 68.6
        & 71.8
        & 65.6
        & 75.6
        & 70.5
        & 65.7
        & 77.4
        & 57.0
        & 83.7
        \\

        Claude-Sonnet-4.5
        & 60.9
        & 68.8
        & 56.5
        & 87.7
        & 35.7
        & 62.9
        & 66.7
        & 62.0
        & 72.1
        & 56.1
        \\

        \midrule


        \multicolumn{11}{c}{
            \textit{\textbf{Open-Source (thinking)}}
        }
        \\[-0.5pt]

        MiMo-VL-SFT
        & 64.5
        & 72.9
        & 58.6
        & 96.3
        & 34.7
        & 72.7
        & 74.1
        & 69.9
        & 78.8
        & 66.7
        \\

        Qwen3-VL-30B
        & 66.0
        & 67.4
        & 63.1
        & 72.4
        & 59.9
        & 69.5
        & 71.8
        & 66.3
        & 78.2
        & 61.0
        \\

        MiMo-VL-RL
        & 59.4
        & 76.2
        & 63.2
        & 96.0
        & 39.6
        & 71.8
        & 71.6
        & 71.3
        & 71.8
        & 71.7
        \\

        LLaVA-Critic-R1
        & 54.3
        & 62.2
        & 52.1
        & 77.3
        & 32.6
        & 53.3
        & 67.6
        & 51.5
        & 98.1
        & 9.4
        \\

        ThinkLite-VL-7B
        & 56.1
        & 63.5
        & 53.3
        & 78.5
        & 34.9
        & 61.9
        & 71.7
        & 56.7
        & 97.4
        & 27.0
        \\

        \midrule


        \multicolumn{11}{c}{
            \textit{\textbf{Open-Source (non-thinking)}}
        }
        \\[-0.5pt]

        Qwen2-VL-7B
        & 53.1
        & 66.1
        & 51.0
        & 93.9
        & 14.5
        & 52.4
        & 67.4
        & 51.0
        & 99.4
        & 6.3
        \\

        Qwen2-VL-72B
        & 55.8
        & 68.4
        & 52.5
        & 98.2
        & 15.7
        & 44.9
        & 50.7
        & 34.5
        & 24.6
        & 29.1
        \\

        DeepSeek-VL
        & 51.6
        & 65.3
        & 50.7
        & 91.8
        & 16.6
        & 49.5
        & 66.2
        & 49.5
        & \textbf{100.0}
        & 0.0
        \\

        Qwen2.5-VL-7B
        & 57.6
        & 56.4
        & 56.4
        & 56.4
        & 58.7
        & 63.8
        & 72.3
        & 58.2
        & 95.5
        & 32.7
        \\

        Qwen2.5-VL-72B
        & 61.5
        & 67.2
        & 57.4
        & 81.0
        & 43.0
        & 68.9
        & 68.4
        & 68.8
        & 68.0
        & 69.8
        \\

        GLM-4V
        & 60.6
        & 74.0
        & 59.9
        & 96.7
        & 32.2
        & 63.8
        & 78.0
        & 71.7
        & 85.3
        & 67.5
        \\

        MiMo-VL-SFT (NoThink)
        & 67.2
        & 64.1
        & 68.5
        & 60.1
        & 73.8
        & 52.1
        & 7.4
        & 85.7
        & 3.9
        & 99.4
        \\

        MiMo-VL-RL (NoThink)
        & 66.6
        & 69.7
        & 62.3
        & 79.1
        & 54.6
        & 55.9
        & 21.5
        & \textbf{90.5}
        & 12.2
        & 98.7
        \\

        Kimi-VL
        & 50.1
        & 66.1
        & 49.4
        & \textbf{100.0}
        & 2.9
        & 51.8
        & 67.2
        & 50.6
        & \textbf{100.0}
        & 4.4
        \\

        InternVL2.5-38B
        & 55.2
        & 67.4
        & 52.2
        & 95.1
        & 17.4
        & 57.8
        & 69.6
        & 54.1
        & 97.4
        & 18.9
        \\

        InternVL2.5-8B
        & 56.7
        & 61.9
        & 54.1
        & 72.4
        & 41.9
        & 52.4
        & 67.5
        & 51.0
        & \textbf{100.0}
        & 5.7
        \\

        Molmo
        & 53.7
        & 33.5
        & 55.7
        & 23.9
        & \textbf{82.0}
        & 50.5
        & 0.0
        & 0.0
        & 0.0
        & \textbf{100.0}
        \\

        Molmo2
        & 50.8
        & 62.6
        & 49.6
        & 84.7
        & 18.6
        & 58.7
        & 61.3
        & 57.2
        & 66.0
        & 51.6
        \\

        Qwen3-VL-8B
        & 66.0
        & 70.3
        & 61.1
        & 82.8
        & 50.0
        & 64.4
        & 72.3
        & 58.9
        & 93.6
        & 35.9
        \\

        Qwen3-VL-32B
        & 69.0
        & 72.5
        & 63.7
        & 84.0
        & 54.6
        & 71.4
        & 68.1
        & 76.2
        & 61.5
        & 81.1
        \\

        InternVL3.5-8B
        & 58.2
        & 65.4
        & 54.8
        & 81.0
        & 36.6
        & 65.1
        & 71.4
        & 60.2
        & 87.7
        & 43.4
        \\

        InternVL3.5-38B
        & 55.2
        & 64.6
        & 52.5
        & 84.0
        & 27.9
        & 68.6
        & 75.2
        & 61.7
        & 96.2
        & 41.5
        \\

        \midrule


        \textbf{Average}
        & \textbf{60.7}
        & \textbf{66.8}
        & \textbf{58.3}
        & \textbf{81.7}
        & \textbf{41.8}
        & \textbf{62.3}
        & \textbf{62.3}
        & \textbf{61.5}
        & \textbf{74.7}
        & \textbf{50.3}
        \\

        \bottomrule

    \end{tabular}
    }
    \vspace{-0.05in}
\caption{
    CRJ performance on ChartEdit and Reproduction.
 $\mathrm{Specificity}=\frac{TN}{TN+FP}$~\cite{statistical}.
}
    \label{tab:semantic_combined}

    \endgroup
    \vspace{-0.15in}
\end{table}

CRJ evaluates task-conditioned verification in \textit{Chart Editing} and \textit{Chart Reproduction}, requiring judges to decide whether each generated chart should be Accepted or Rejected.

\finding{Finding 3: Query-aware verification remains challenging even for strong LMM judges.}

As shown in Tab.~\ref{tab:semantic_combined}, proprietary models achieve the best performance, but remain limited under strict verification. Gemini-3-Pro and GPT-5.2 reach 79.1\% accuracy on \textit{Chart Editing}, and 75.6\% and 74.3\% on \textit{Chart Reproduction}. Open-source models show lower and more compressed performance, with leading models such as Qwen3-VL-32B reaching only moderate accuracy. These results show that reliable chart-to-code verification remains difficult.

\finding{Finding 4: Many LMM judges exhibit lenience bias toward the \textit{Accept} label.}

Recall and Specificity~\cite{statistical} reveal that many models achieve reasonable accuracy by over-predicting Accept. Across many open-source models, Recall is often above 80\% or 90\%, while Specificity drops sharply. For example, LLaVA-Critic-R1 reaches 98.1\% Recall but only 9.4\% Specificity on \textit{Chart Reproduction}. This imbalance indicates that many judges accept plausible-looking charts, but fail to reject subtle errors, incomplete reproduction, or unintended editing side effects.

\begin{figure}[t]
    \centering
    \includegraphics[width=0.78\linewidth]
    {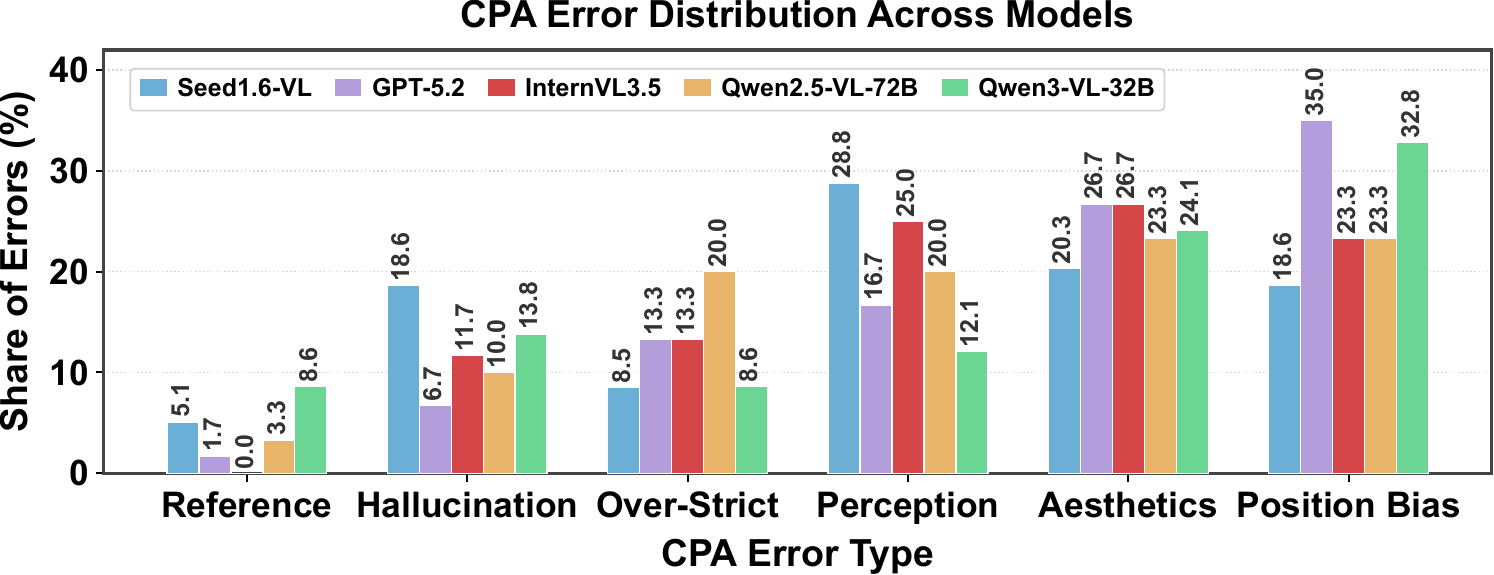}
    \vspace{-0.6em}
    \caption{Percentage distribution of CPA error types across representative LMMs. (Finding 5 and Finding 6)}
    \label{fig:cpa_error_distribution}
    \vspace{-0.7em}
\end{figure}

\begin{figure}[t]
    \centering
    \includegraphics[width=0.74\linewidth]
    {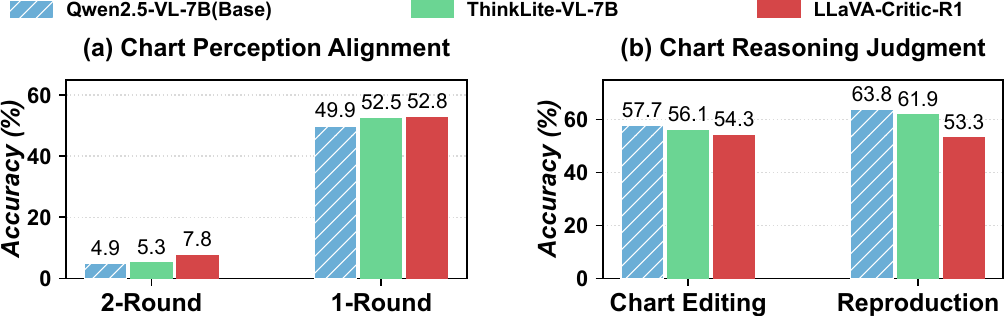}
    \vspace{-0.6em}
    \caption{Accuracy comparison of the base and post-trained models on
    (a) CPA and
    (b) CRJ. (Finding 7 and Finding 8)}
    \label{fig:combined_error_analysis}
    \vspace{-0.7em}
\end{figure}

\subsection{Diagnostic Error Analysis}
One of the most important sections of a benchmark study is to analyze its failures. Therefore, we examine incorrect predictions with model rationales and chart evidence, then group them into 6 error types. Fig.~\ref{fig:cpa_error_distribution} reveals several sources of model-specific failure, including aesthetic preference, visual perception, positional shortcuts, hallucination, and over-strict comparison. The analyzed errors span 17 distinct chart types, with the largest single category accounting for only 13\% of cases, suggesting that the benchmark difficulty is not driven by a narrow subset of chart types.


\finding{Finding 5: Aesthetic alignment remains a common failure mode.}

Aesthetic-related errors account for a substantial share of incorrect CPA judgments, ranging from 20.3\% for Seed1.6-VL to 26.7\% for GPT-5.2 and InternVL3.5-38B (see Fig.~\ref{fig:cpa_error_distribution}). This suggests that even strong LMM judges struggle with human-aligned legibility and overall chart presentation.

\finding{Finding 6: Models exhibit distinct fallback strategies when facing difficult visual comparisons.}

When visual differences are difficult to resolve, models exhibit different fallback behaviors. Position bias accounts for 35.0\% of GPT-5.2's errors and 32.8\% of Qwen3-VL-32B's errors. Seed1.6-VL shows more perception errors and hallucinated rationales, while Qwen2.5-VL-72B often exhibits over-strict comparison, penalizing minor visual variations despite acceptable overall quality (see Fig.~\ref{fig:cpa_error_distribution}). These patterns suggest that improving chart judges requires targeted correction of model-specific failure modes.

\begin{figure}[t]
    \centering
    \includegraphics[width=0.76\linewidth]{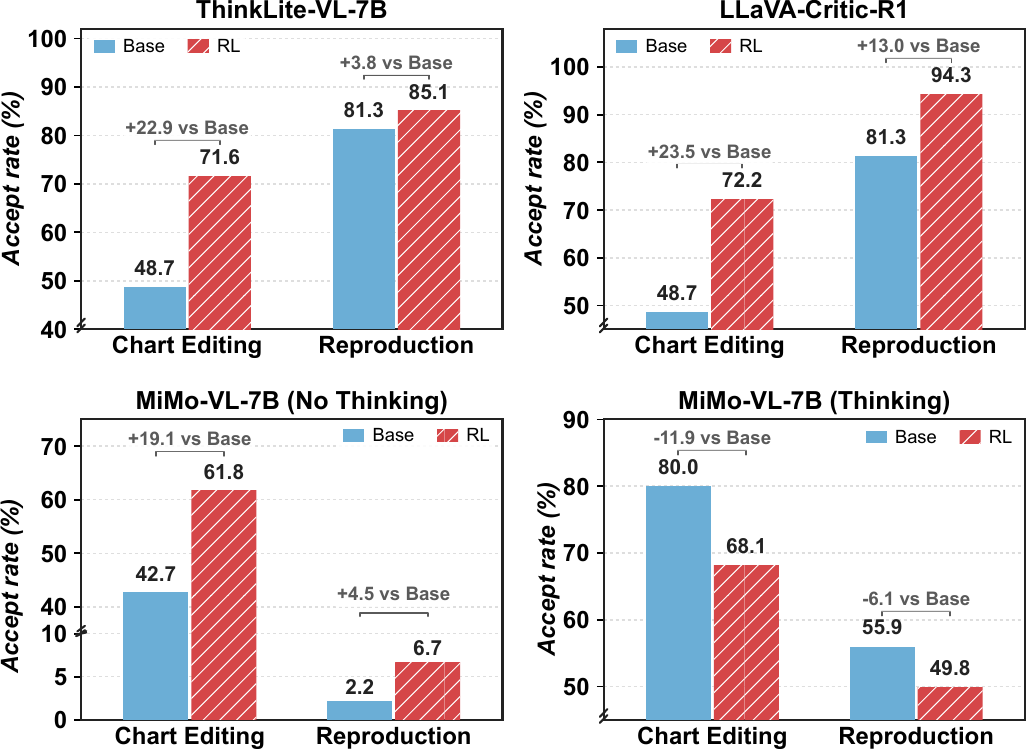}
    \caption{Acceptance rates before and after post-training on Chart Editing and Chart Reproduction.}
    \label{fig:post_training_accept_rates}
\end{figure}

\subsection{Post-Training Model Analysis}
We further analyze whether post-training and thinking-style inference improve chart judgment. Fig.~\ref{fig:combined_error_analysis}a and Fig.~\ref{fig:combined_error_analysis}b show limited gains in CPA, but increased lenience bias in CRJ.

\finding{Finding 7: Post-training and thinking-style inference provide limited gains in pairwise chart comparison.}

As shown in Fig.~\ref{fig:combined_error_analysis}a, post-trained and thinking-style models provide only limited CPA gains over Qwen2.5-VL-7B. They slightly improve one-pass accuracy, but remain weak under the stricter two-pass setting, suggesting that reasoning or critique tuning alone does not reliably improve fine-grained pairwise chart comparison.

\finding{Finding 8: Post-training and thinking-style inference may increase lenience bias in CRJ and therefore reduce accuracy.}

In CRJ, post-trained or thinking-oriented models often become more likely to predict Accept. Fig.~\ref{fig:post_training_accept_rates} shows LLaVA-Critic-R1 increases the \textit{Chart Editing} accept rate from 48.7\% to 72.2\% relative to its base model, while the MiMo family shows a similar increase after RL training in the NoThink setting. This higher acceptance tendency comes with lower specificity, indicating that these models often accept flawed but visually plausible charts and therefore reduce accuracy (see Fig.~\ref{fig:combined_error_analysis}b). Overall, current post-training and thinking-style mechanisms make CRJ judges more lenient rather than improving fine-grained chart verification.

\vspace{-0.05in}
\subsection{Case Study}

\suppressfloats[t]
\begin{figure}[t]
    \centering
    \includegraphics[width=\linewidth]{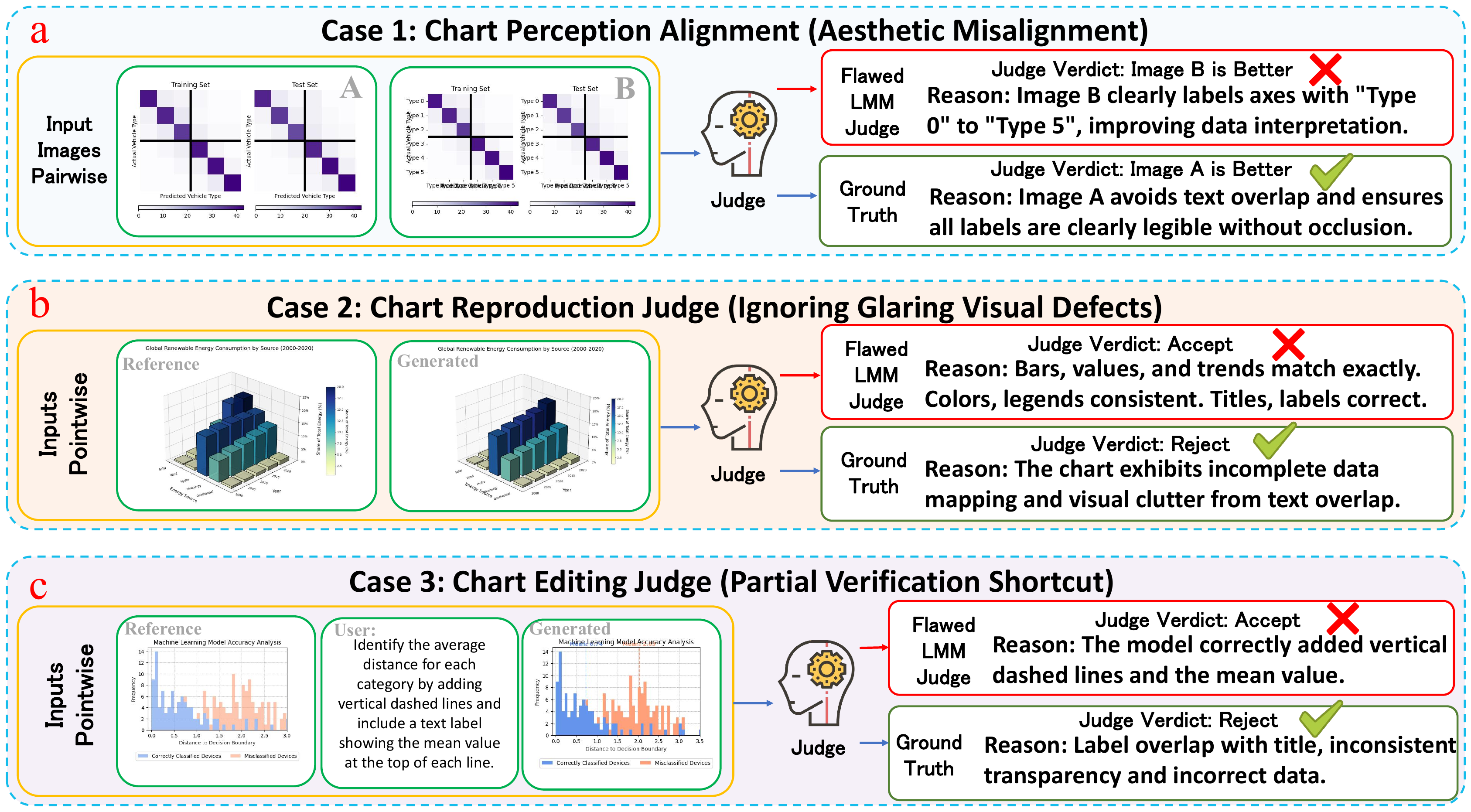}
    \caption{Representative failure modes of LMM judges in (a) Chart Perception Alignment, (b) Chart Reproduction, and (c) Chart Editing. The cases expose aesthetic misalignment, missed visual defects, and partial verification shortcuts.}
    \label{fig:case_study}
\end{figure}

Fig.~\ref{fig:case_study} presents three representative failure modes. In (a), the judge favors explicit labels but overlooks text overlap and poorer legibility. In (b), it accepts a reproduction with incomplete data and visual clutter. In (c), it verifies the requested edit but misses unintended changes to non-target elements. Together, these cases illustrate aesthetic misalignment, missed visual defects, and partial verification shortcuts.

\vspace{-0.05in}
\subsection{Test-Time Computation Analysis}

We examine whether additional test-time computation improves CRJ judgment. Following VL-RewardBench~\cite{vlrewardbench}, we apply majority voting over $K \in \{1,3,5,7,9\}$ independent runs, using temperature $0.6$ and top-$p=0.95$ for diverse judgments.
\begin{figure}[t]
    \centering
    \includegraphics[width=0.96\linewidth]{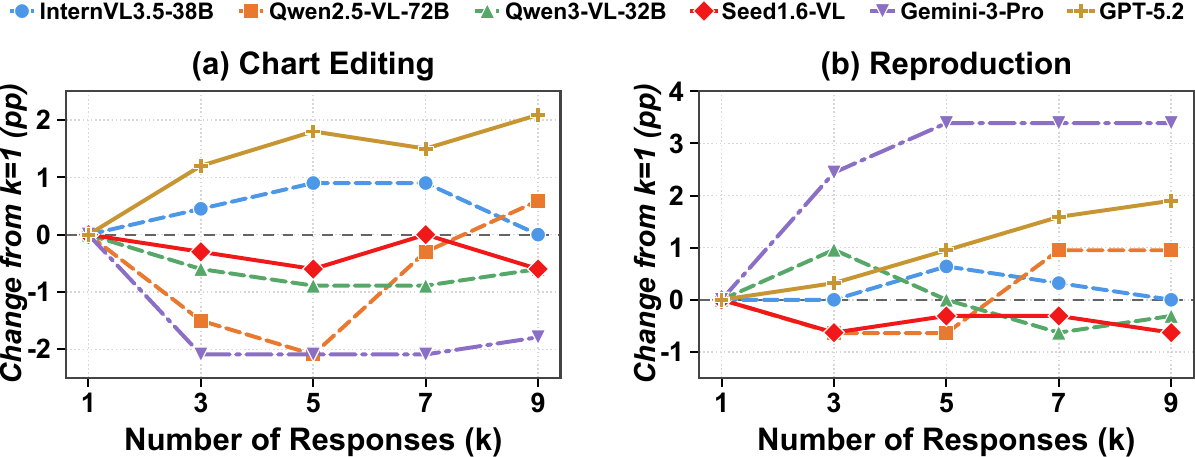}
    \caption{Test-time computation scaling under majority voting. Values show percentage-point changes relative to $K=1$ for Chart Editing and Chart Reproduction.}
    \label{fig:test_time_scaling}
\end{figure}

\finding{Finding 9: Majority voting provides limited and model-dependent gains for chart judgment.}

As shown in Fig.~\ref{fig:test_time_scaling}, test-time voting improves some proprietary models but does not consistently benefit all judges. GPT-5.2 improves moderately on \textit{Chart Editing} as $K$ increases, while Gemini-3-Pro gains more clearly on \textit{Chart Reproduction}. In contrast, most open-source models show little improvement or slight degradation under larger $K$. These results suggest that majority voting can help strong judges in some cases, but does not reliably correct the verification failures and lenience bias observed in current open-source LMM judges.

\section{Conclusion}
We introduce \benchmark{}, a diagnostic benchmark for evaluating LMM judges in chart-to-code workflows. We design the CPA task to evaluate fundamental chart judgment across 3 major dimensions and 7 empirically derived criteria, enabling a fine-grained and disentangled diagnosis of LMM judgment capabilities. We further design the CRJ task around two representative chart-to-code tasks, \textit{Chart Reproduction} and \textit{Chart Editing}. Together, CPA and CRJ reflect the realistic RL training settings. Overall, \benchmark{} provides a controlled diagnostic testbed for exposing failure modes and guiding the development of more reliable chart judges.

\clearpage
\bibliography{references}
\bibliographystyle{iclr2027_conference}

\clearpage
\appendix
\raggedbottom
\section{Preliminary Study}
\label{app:preliminary_study}
\subsection{CPA Composition and Error Coverage in Model Outputs}
\label{sec:cpa_model_output_coverage}

Table~\ref{tab:cpa_category_distribution} summarizes the composition of CPA. The CPA subset contains 1,003 cases spanning exactly seven fine-grained error criteria: data encoding, text occlusion, component positioning, color consistency, visual style, axis scaling, and aspect ratio. Although the number of cases varies across criteria, all seven error types are sufficiently represented to support criterion-level evaluation.

\begin{supplementtable}
    \caption{Distribution of the seven fine-grained error criteria in CPA.}
    \label{tab:cpa_category_distribution}
    \small
    \setlength{\tabcolsep}{7pt}
    \renewcommand{\arraystretch}{1.02}
    \begin{tabular}{@{}lrr@{}}
        \toprule
        \textbf{CPA Criterion} & \textbf{Count} & \textbf{Share (\%)} \\
        \midrule
        Data Encoding         & 164 & 16.35 \\
        Text Occlusion        & 157 & 15.65 \\
        Component Positioning & 154 & 15.35 \\
        Color Consistency     & 117 & 11.67 \\
        Visual Style          &  88 &  8.77 \\
        Axis Scaling          & 141 & 14.06 \\
        Aspect Ratio          & 182 & 18.15 \\
        \midrule
        \textbf{Total}        & \textbf{1,003} & \textbf{100.00} \\
        \bottomrule
    \end{tabular}
\end{supplementtable}

To examine whether the synthetic error criteria in CPA reflect errors observed in actual model outputs, we randomly sampled 200 first-pass chart-to-code outputs from a joint pool containing outputs generated by four leading frontier models: GPT-5.2, Gemini-3-Pro, Claude-Sonnet-4.5, and Seed1.6-VL. We retained all sampled outputs, including correct renderable outputs, erroneous renderable outputs, and non-renderable outputs.

For the 159 erroneous but renderable outputs, we performed multi-label annotation using the seven CPA criteria. An error that could not be assigned to any of these criteria was marked as \textit{Other}.

We considered an output \textit{fully covered} if all of its annotated errors could be assigned to the seven CPA criteria. An output was \textit{partially covered} if it contained both CPA-covered errors and errors marked as \textit{Other}. An output was \textit{uncovered} if none of its annotated errors could be assigned to a CPA criterion. An output was counted as \textit{containing at least one CPA error} if at least one of its annotated errors could be assigned to the seven CPA criteria.

\begin{table}[t!]
    \centering
    \caption{Error distribution and CPA coverage in 200 randomly sampled model-generated outputs.}
    \label{tab:model_output_error_coverage}
    \small
    \setlength{\tabcolsep}{7pt}
    \renewcommand{\arraystretch}{1.0}
    \begin{tabular}{@{}lrr@{}}
        \toprule
        \textbf{Statistic} & \textbf{Count / Value} & \textbf{Rate (\%)} \\
        \midrule
        \multicolumn{3}{l}{\textit{Output status among 200 sampled outputs}} \\
        \addlinespace[2pt]
        Correct renderable   & 27  & 13.50 \\
        Erroneous renderable & 159 & 79.50 \\
        Non-renderable       & 14  &  7.00 \\
        \midrule
        \multicolumn{3}{l}{\textit{Error occurrences among 159 erroneous renderable outputs}} \\
        \addlinespace[2pt]
        Data Encoding         & 69 & 43.40 \\
        Text Occlusion        & 62 & 38.99 \\
        Component Positioning & 57 & 35.85 \\
        Color Consistency     & 54 & 33.96 \\
        Visual Style          & 43 & 27.04 \\
        Axis Scaling          & 36 & 22.64 \\
        Aspect Ratio          & 19 & 11.95 \\
        Other                 & 15 & 9.43 \\
        \midrule
        \multicolumn{3}{l}{\textit{Error multiplicity among 159 erroneous outputs}} \\
        \addlinespace[2pt]
        Single-error outputs                & 37   & 23.27 \\
        Multi-error outputs                 & 122  & 76.73 \\
        Average errors per erroneous output & 2.23 & N/A \\
        \midrule
        \multicolumn{3}{l}{\textit{CPA coverage among 159 erroneous outputs}} \\
        \addlinespace[2pt]
        Fully covered                     & 144 & 90.57 \\
        Partially covered                 & 11  &  6.92 \\
        Uncovered                         & 4   &  2.52 \\
        Containing at least one CPA error & 155 & 97.48 \\
        \bottomrule
    \end{tabular}

\end{table}

As shown in Table~\ref{tab:model_output_error_coverage}, errors in model-generated chart-to-code outputs commonly co-occur. Among the 159 erroneous renderable outputs, 122 contain two or more error types, accounting for 76.73\% of these outputs. Each erroneous output contains 2.23 error types on average. This result suggests that evaluating chart-to-code outputs often requires a judge to identify multiple coexisting visual problems rather than a single isolated error.

Data encoding, text occlusion, and component positioning are the most frequent error types in the sample, whereas aspect-ratio errors occur less frequently. CPA does not aim to reproduce the observed frequency distribution exactly. Instead, it provides sufficient cases for every criterion, including relatively infrequent error types, to support stable and fine-grained evaluation.

Regarding coverage, 144 of the 159 erroneous outputs are fully covered by the seven CPA criteria, 11 are partially covered, and only four are uncovered. Overall, 155 outputs, or 97.48\%, contain at least one error defined by CPA. Among the 355 annotated error-category assignments, 340 can be assigned to the seven CPA criteria, accounting for 95.77\%. The remaining 15 assignments are marked as \textit{Other}. These results indicate that the CPA taxonomy captures the dominant error types observed in model-generated chart-to-code outputs, with only a small number of errors falling outside the seven criteria.

Representative model-output examples of the seven CPA error types are provided in Section~\ref{sec:cpa_error_examples}, showing that each synthetic criterion corresponds to errors observed in model-generated outputs.

\subsection{Representative Errors in Model Outputs across the Seven CPA Criteria}
\label{sec:cpa_error_examples}
Before formalizing the \benchmark benchmark, we conducted a preliminary study to understand errors in model-generated outputs from state-of-the-art Large Multimodal Models (LMMs) in chart generation tasks. We examined outputs from leading frontier models (e.g., GPT-5.2, Gemini-3-Pro, Claude-Sonnet-4.5, and Seed1.6-VL) evaluated on existing chart-to-code benchmarks such as ChartMimic, Chart2code, and ChartEdit. 

A critical observation emerged: a high code execution success rate does not equate to high visual or semantic quality. While these frontier models can successfully generate executable Python code, the resulting rendered charts frequently exhibit a myriad of fine-grained visual and structural defects. To illustrate this phenomenon, the detailed case studies presented below highlight representative errors in model outputs, showcasing glaring discrepancies between the generated outputs and the expected chart standards.

\begin{tcolorbox}[
    colback=gray!5!white, 
    colframe=MyDarkBlue, 
    title=Typical Failure Cases of SOTA LMMs across 7 Dimensions,
    arc=4pt, 
    boxrule=0.8pt, 
    fonttitle=\bfseries,
    width=\linewidth,
    breakable  
]
\small

\noindent
\begin{minipage}[c]{0.55\linewidth}
    \centering
    \begin{minipage}{0.48\linewidth}
        \centering
        \includegraphics[width=\linewidth]{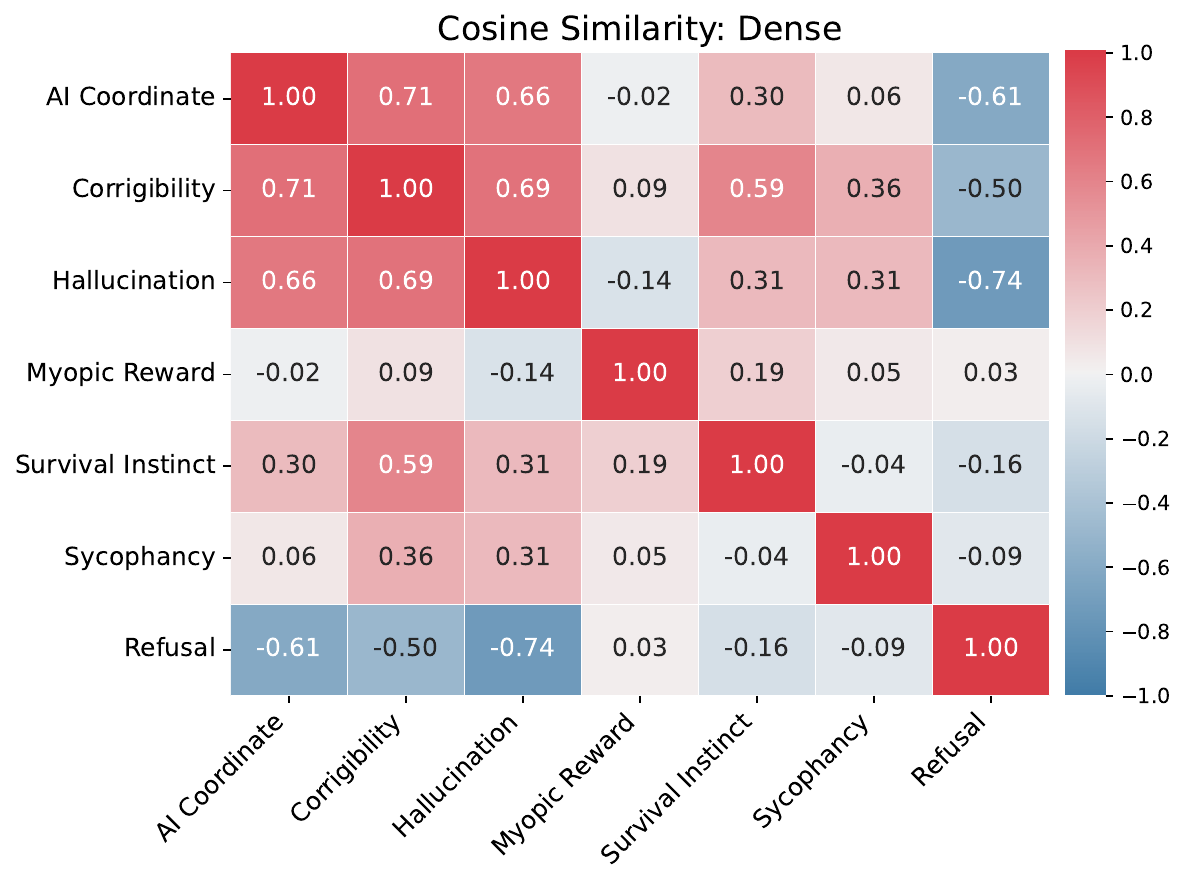} 
        \vspace{2pt} \\ \scriptsize \textit{Ground Truth}
    \end{minipage}\hfill
    \begin{minipage}{0.48\linewidth}
        \centering
        \includegraphics[width=\linewidth]{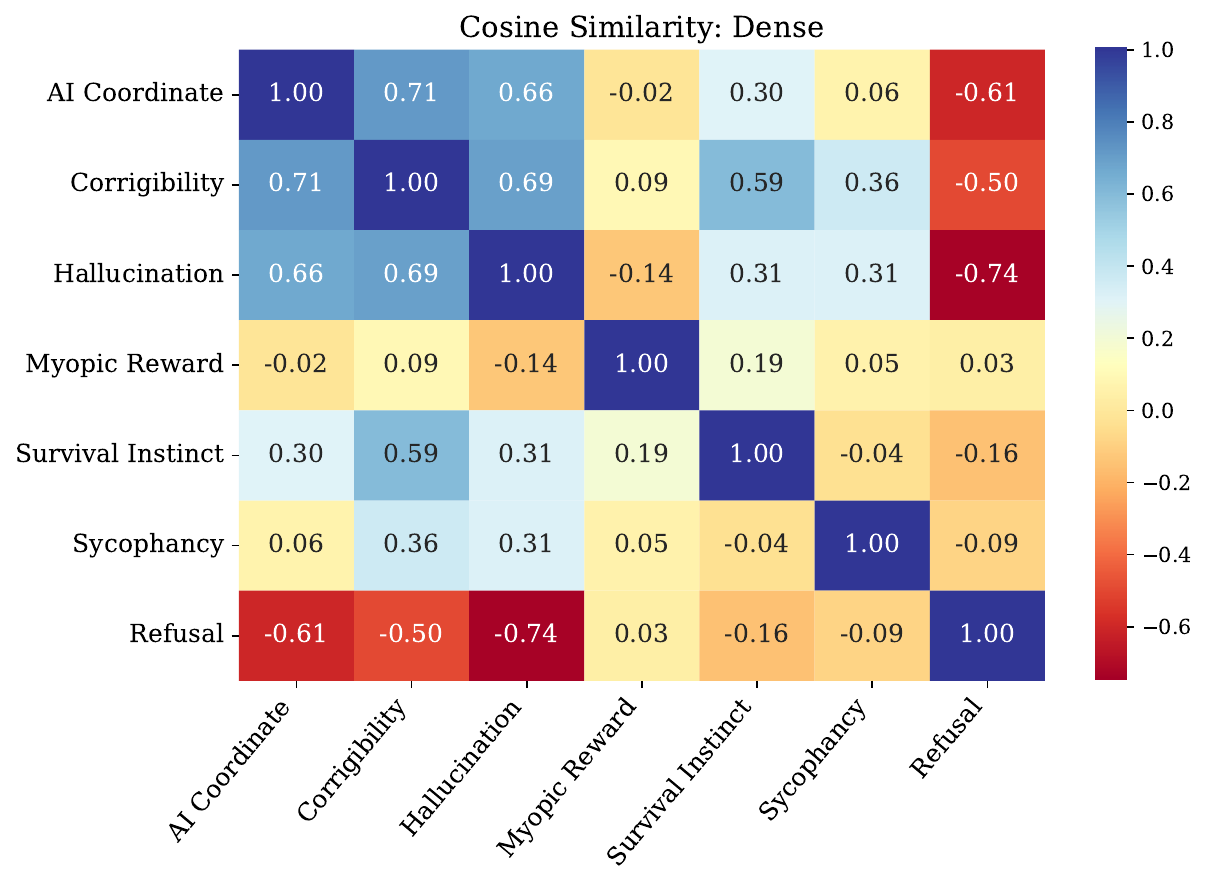} 
        \vspace{2pt} \\ \scriptsize \textit{Model Generated}
    \end{minipage}
\end{minipage}\hfill
\begin{minipage}[c]{0.42\linewidth}
    \scriptsize
    \textbf{[Dimension 1: Color Consistency]} \\
    \textit{Model:} GPT-5.2 \\
    \textbf{Observation:} Significant color transition discrepancies from the GT reveal the model's poor perception of subtle color variations. In open-ended tasks, these deficits lead to clashing or indistinguishable palettes that severely misalign with human aesthetics.
\end{minipage}

\vspace{0.8em}\hrule\vspace{0.8em}

\noindent
\begin{minipage}[c]{0.55\linewidth}
    \centering
    \begin{minipage}{0.48\linewidth}
        \centering
        \includegraphics[width=\linewidth]{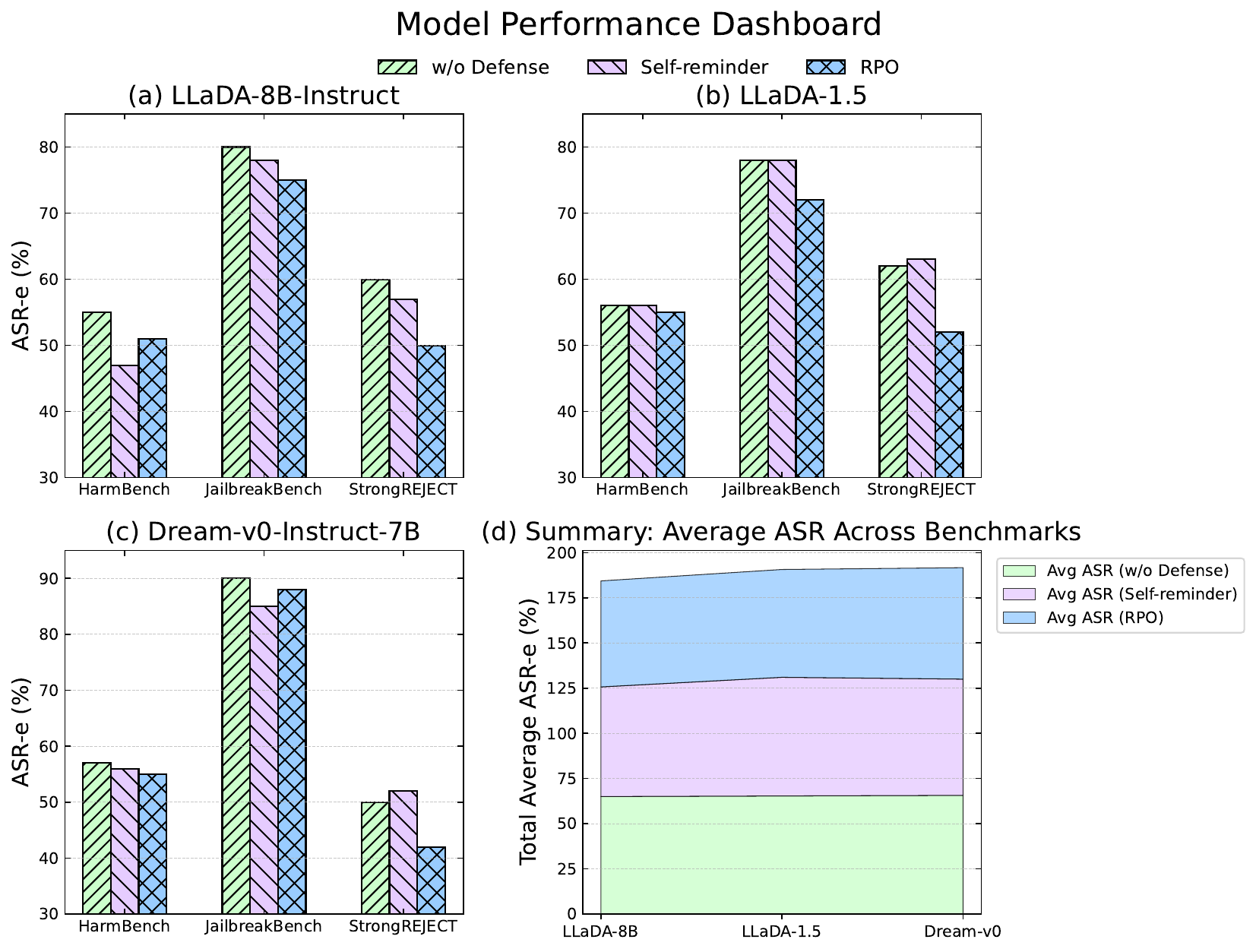} 
        \vspace{2pt} \\ \scriptsize \textit{Ground Truth}
    \end{minipage}\hfill
    \begin{minipage}{0.48\linewidth}
        \centering
        \includegraphics[width=\linewidth]{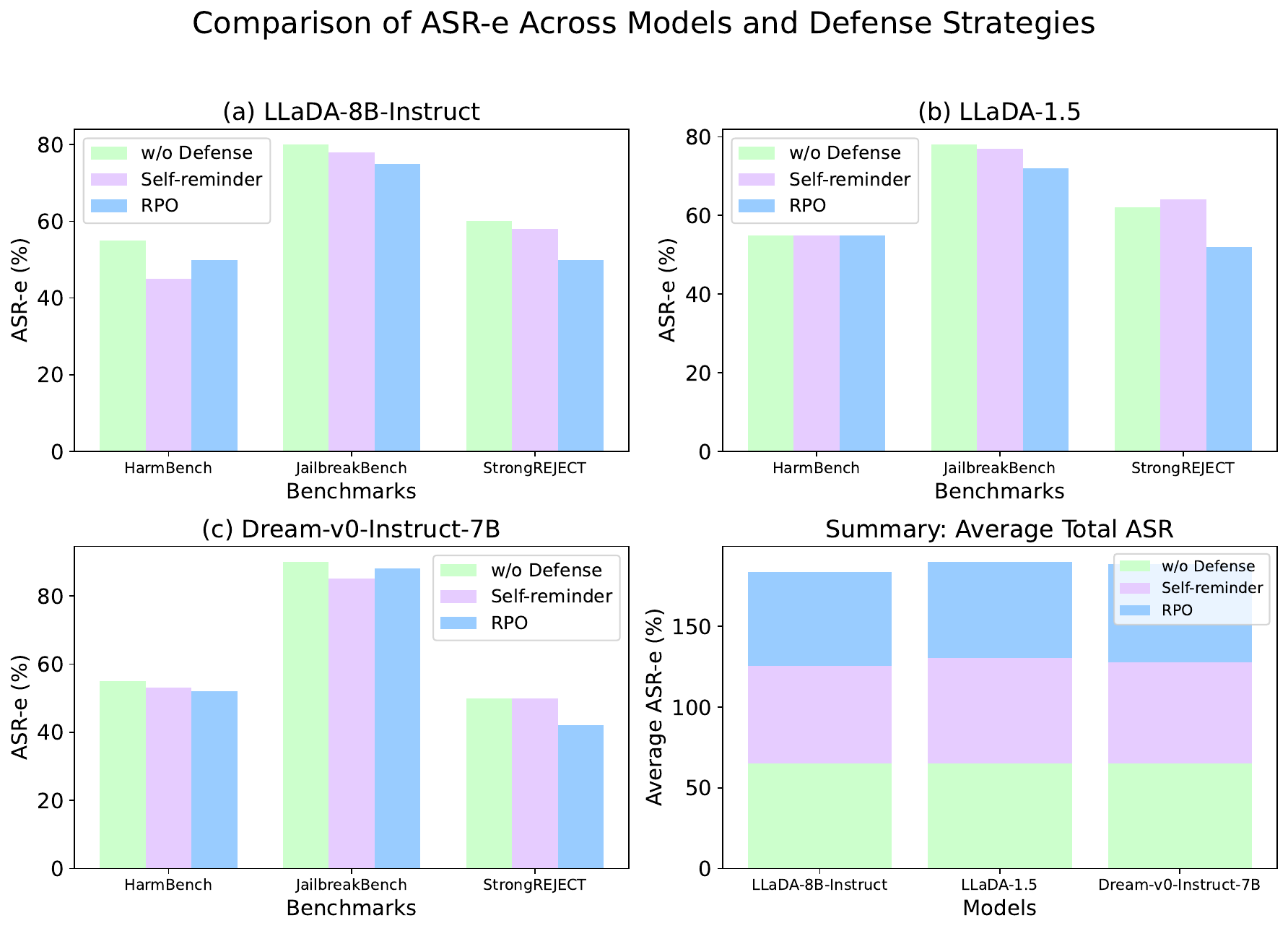} 
        \vspace{2pt} \\ \scriptsize \textit{Model Generated}
    \end{minipage}
\end{minipage}\hfill
\begin{minipage}[c]{0.42\linewidth}
    \scriptsize
    \textbf{[Dimension 2: Visual Style]} \\
    \textit{Model:} Qwen2.5-VL-72B  \\
    \textbf{Observation:} The filling patterns of the bar chart in the left plot of the model output on the right are inconsistent with those in the GT Figure, and the legend styles also differ. Additionally, the chart in the lower-right corner exhibits a chart type recognition error.
\end{minipage}

\vspace{0.8em}\hrule\vspace{0.8em}

\noindent
\begin{minipage}[c]{0.55\linewidth}
    \centering
    \begin{minipage}{0.48\linewidth}
        \centering
        \includegraphics[width=\linewidth]{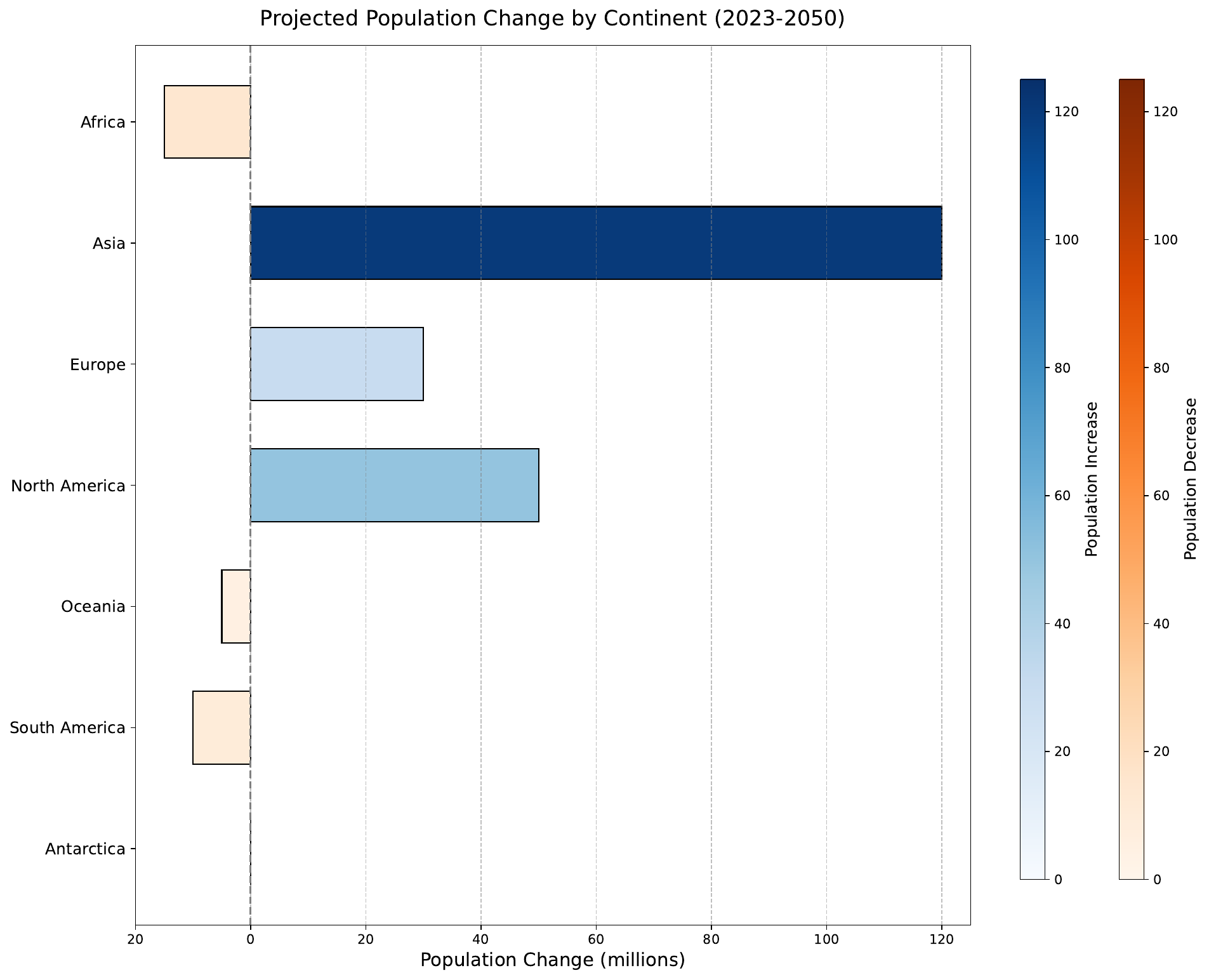} 
        \vspace{2pt} \\ \scriptsize \textit{Ground Truth}
    \end{minipage}\hfill
    \begin{minipage}{0.48\linewidth}
        \centering
        \includegraphics[width=\linewidth]{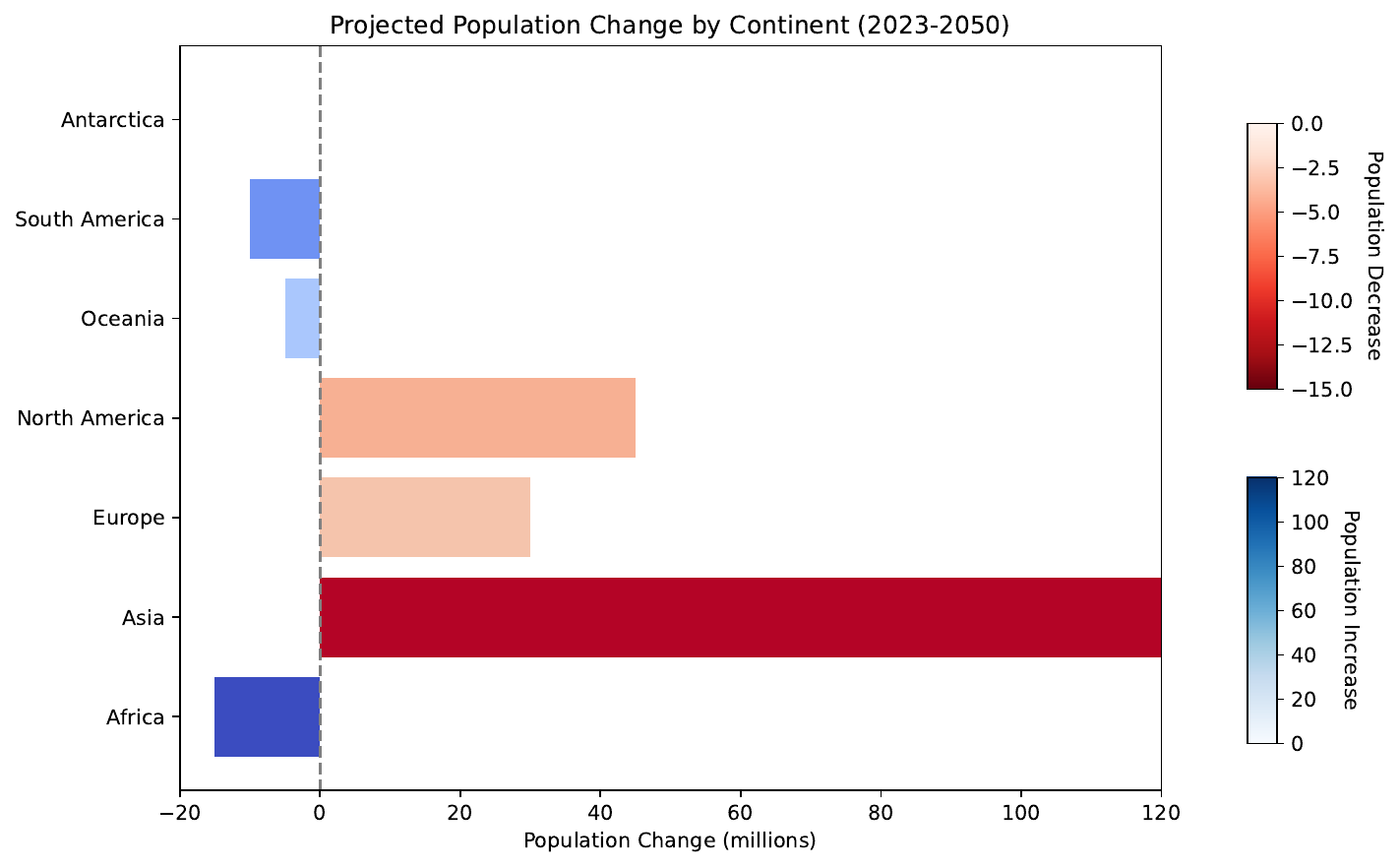} 
        \vspace{2pt} \\ \scriptsize \textit{Model Generated}
    \end{minipage}
\end{minipage}\hfill
\begin{minipage}[c]{0.42\linewidth}
    \scriptsize
    \textbf{[Dimension 3: Aspect Ratio]} \\
    \textit{Model:} Seed1.5-VL \\
    \textbf{Observation:} Aspect ratio gaps (1.08 vs. 1.67) expose the model's inability to maintain visual balance, causing horizontal stretching and representation distortion.
\end{minipage}

\vspace{0.8em}\hrule\vspace{0.8em}

\noindent
\begin{minipage}[c]{0.55\linewidth}
    \centering
    \begin{minipage}{0.48\linewidth}
        \centering
        \includegraphics[width=\linewidth]{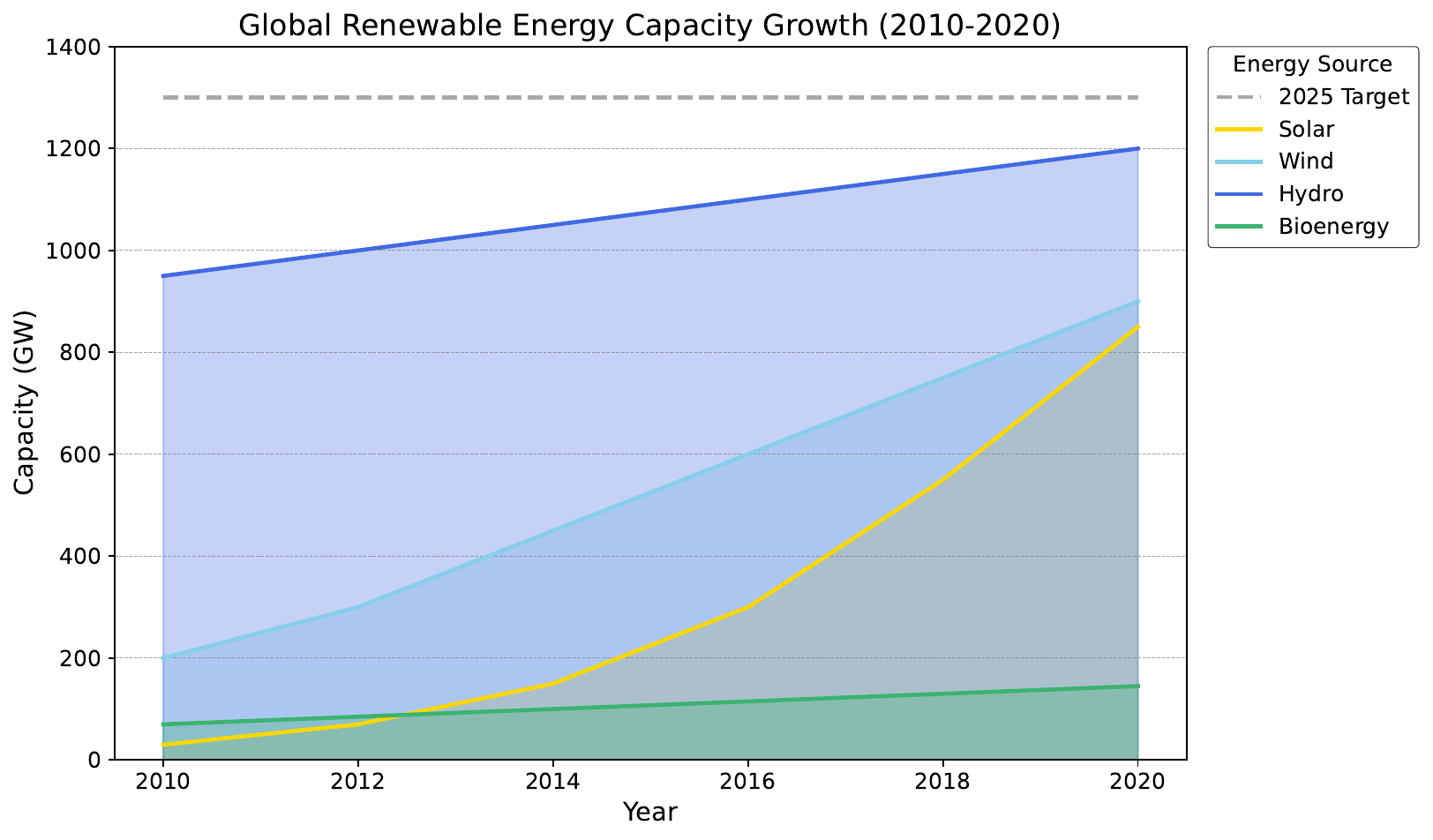} 
        \vspace{2pt} \\ \scriptsize \textit{Ground Truth}
    \end{minipage}\hfill
    \begin{minipage}{0.48\linewidth}
        \centering
        \includegraphics[width=\linewidth]{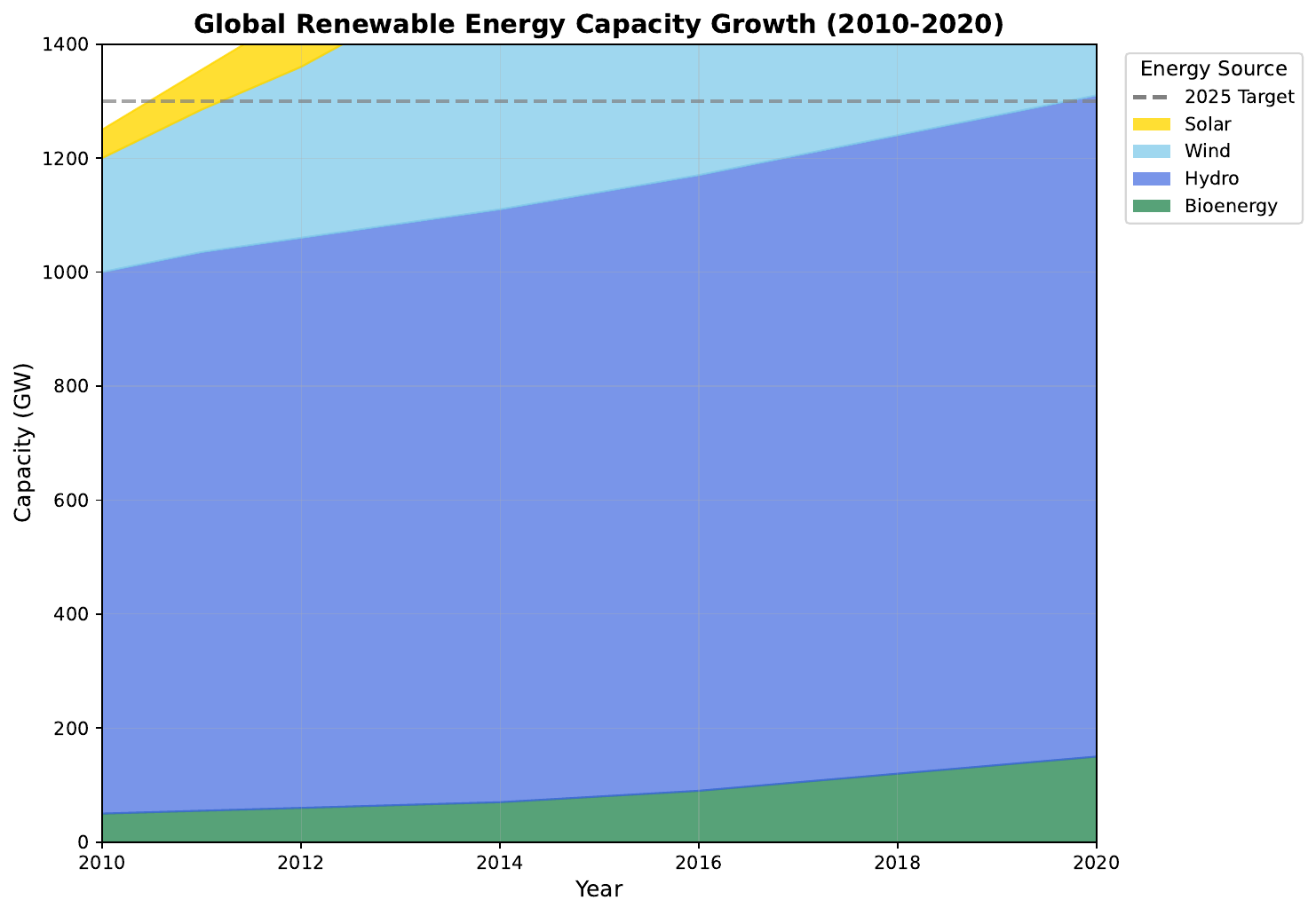} 
        \vspace{2pt} \\ \scriptsize \textit{Model Generated}
    \end{minipage}
\end{minipage}\hfill
\begin{minipage}[c]{0.42\linewidth}
    \scriptsize
    \textbf{[Dimension 4: Axis Scaling]} \\
    \textit{Model:} Claude-Sonnet-4.5 \\
    \textbf{Observation:} The y‑axis range in the model output on the right is problematic, causing part of the data to be invisible. This indicates an insufficient match between the axis range and the data. Additionally, errors occurred in data recognition and analysis.
\end{minipage}

\vspace{0.8em}\hrule\vspace{0.8em}

\noindent
\begin{minipage}[c]{0.55\linewidth}
    \centering
    \begin{minipage}{0.48\linewidth}
        \centering
        \includegraphics[width=\linewidth]{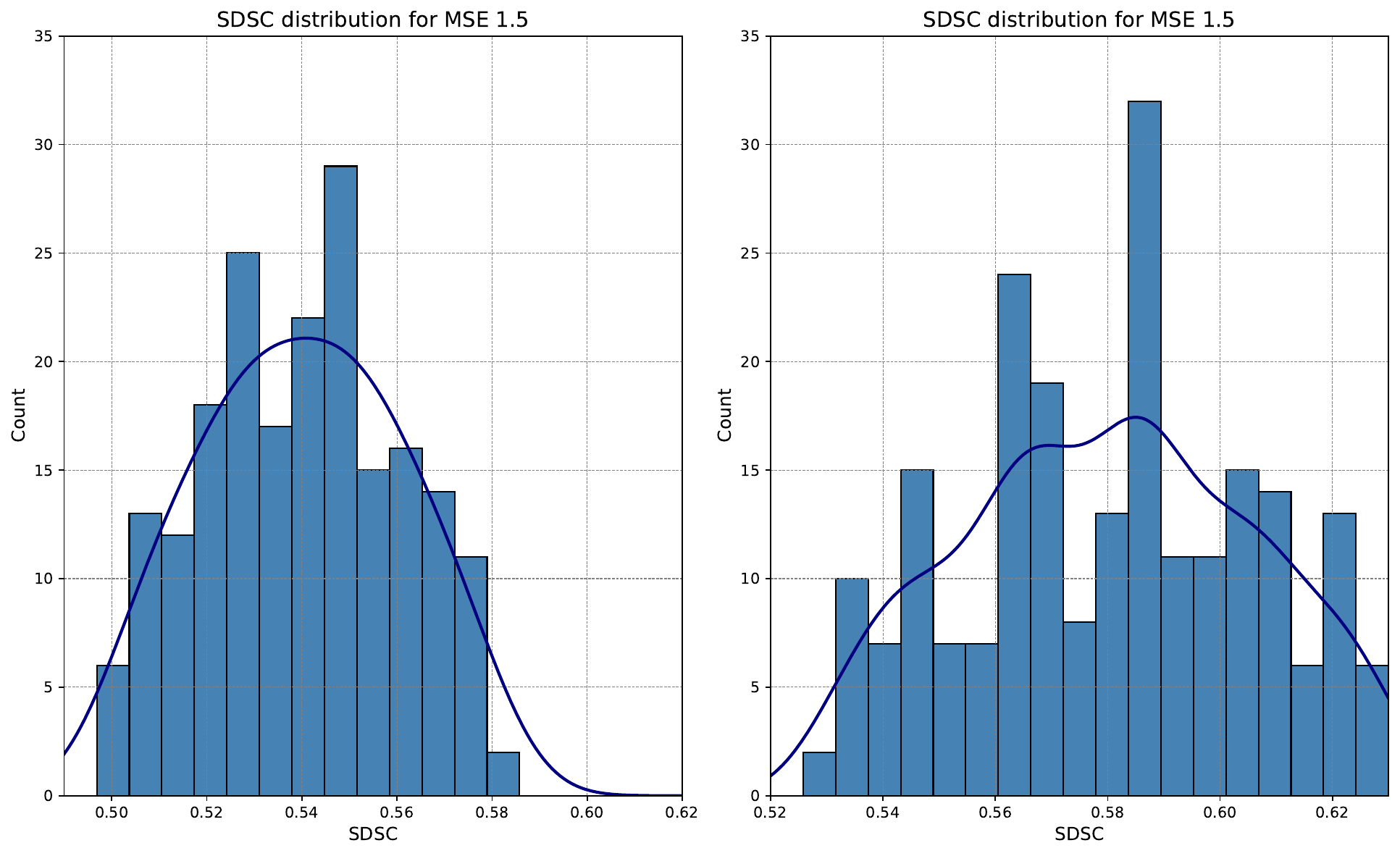} 
        \vspace{2pt} \\ \scriptsize \textit{Ground Truth}
    \end{minipage}\hfill
    \begin{minipage}{0.48\linewidth}
        \centering
        \includegraphics[width=\linewidth]{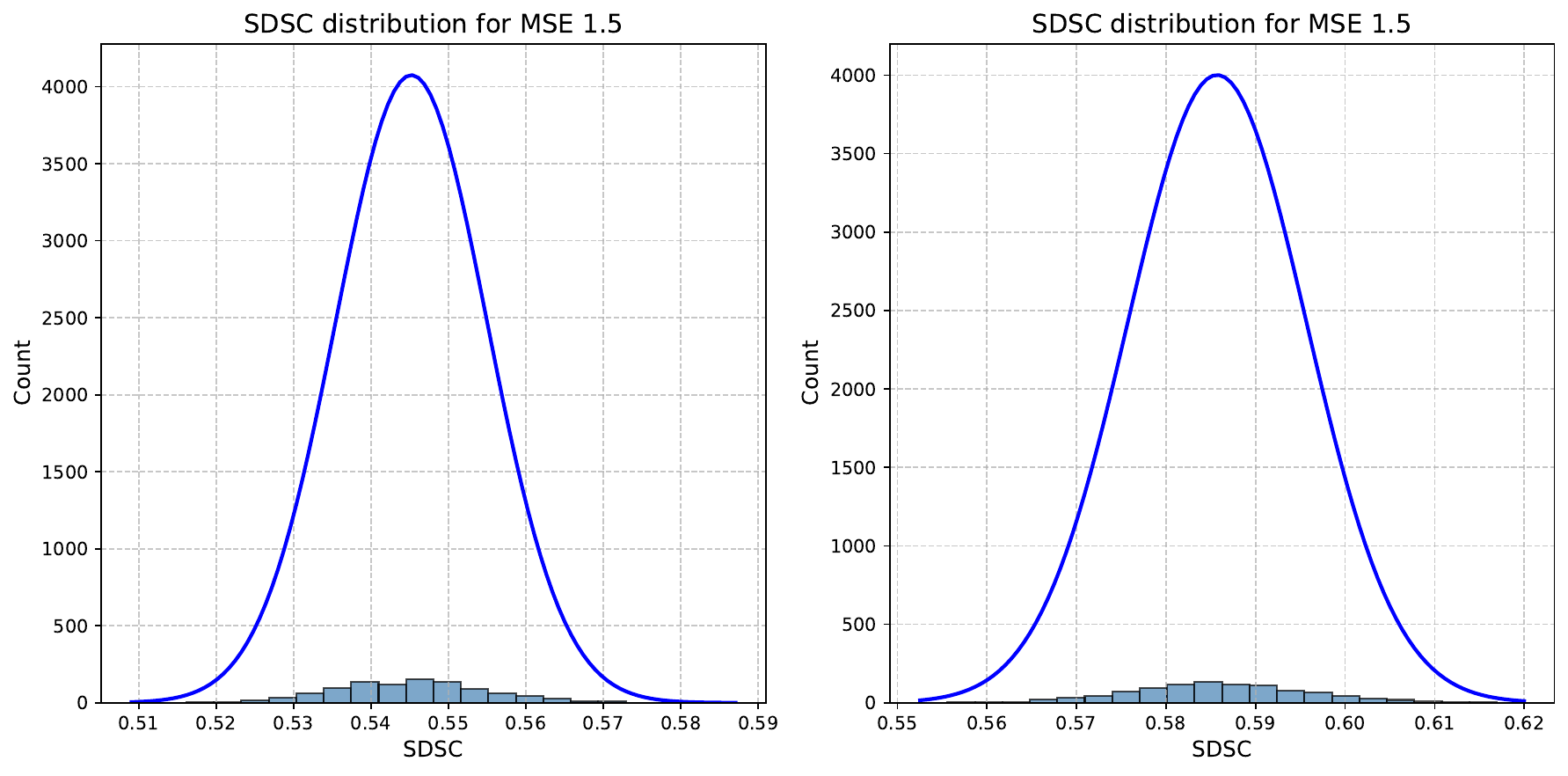} 
        \vspace{2pt} \\ \scriptsize \textit{Model Generated}
    \end{minipage}
\end{minipage}\hfill
\begin{minipage}[c]{0.42\linewidth}
    \scriptsize
    \textbf{[Dimension 5: Data Encoding]} \\
    \textit{Model:} Qwen3-VL-30B\\
    \textbf{Observation:} Data distribution inconsistencies from the GT expose the model's flawed data representation and weak visual data extraction.
\end{minipage}

\vspace{0.8em}\hrule\vspace{0.8em}

\noindent
\begin{minipage}[c]{0.55\linewidth}
    \centering
    \begin{minipage}{0.48\linewidth}
        \centering
        \includegraphics[width=\linewidth]{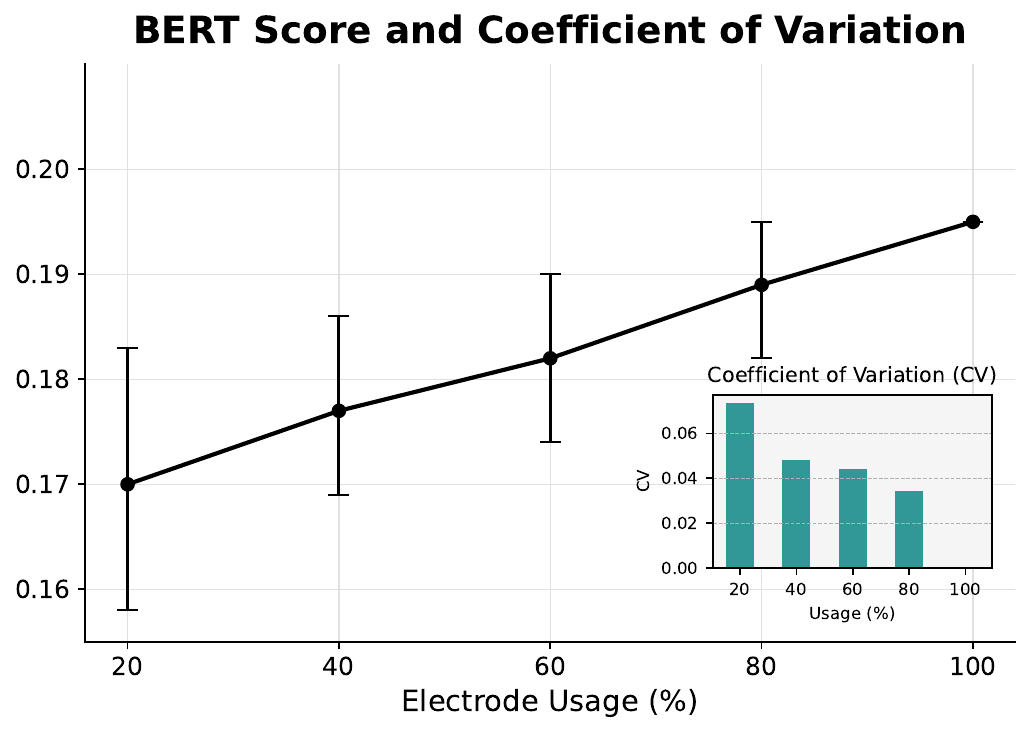} 
        \vspace{2pt} \\ \scriptsize \textit{Ground Truth}
    \end{minipage}\hfill
    \begin{minipage}{0.48\linewidth}
        \centering
        \includegraphics[width=\linewidth]{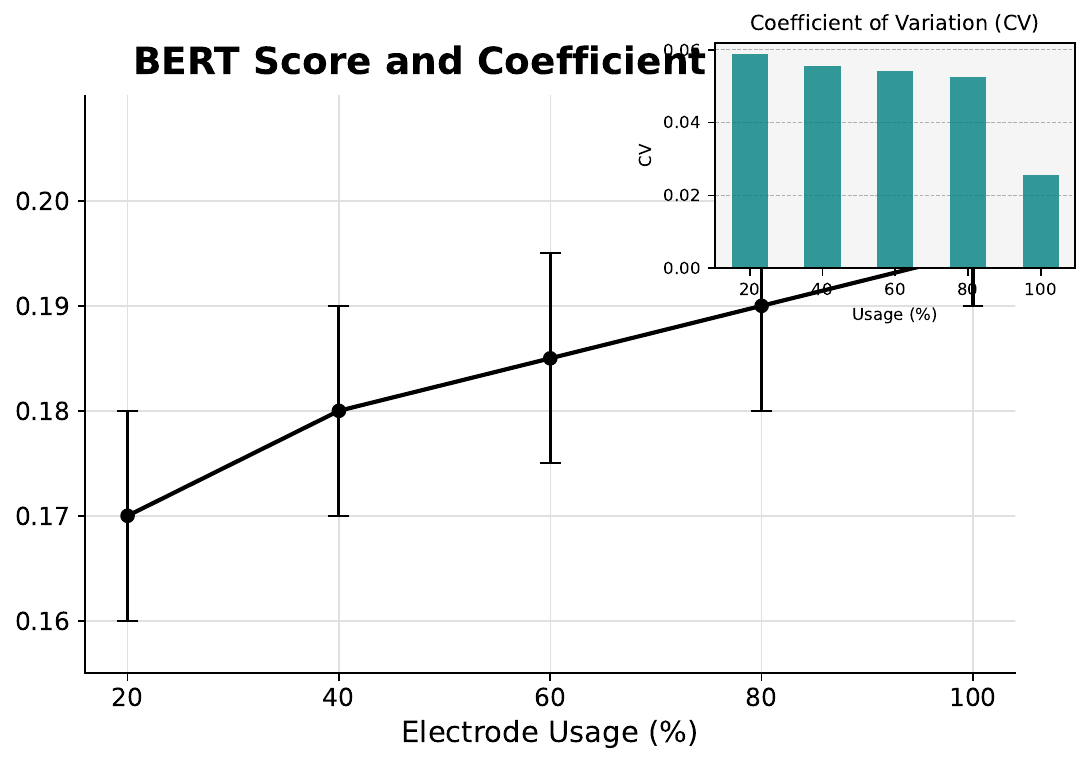} 
        \vspace{2pt} \\ \scriptsize \textit{Model Generated}
    \end{minipage}
\end{minipage}\hfill
\begin{minipage}[c]{0.42\linewidth}
    \scriptsize
    \textbf{[Dimension 6: Component Positioning]} \\
    \textit{Model:} Seed1.6-VL \\
    \textbf{Observation:} The subplot position in the model output on the right does not match that of the GT Figure, severely obscuring the data portion of the error points and causing significant visibility issues.
\end{minipage}

\vspace{0.8em}\hrule\vspace{0.8em}

\noindent
\begin{minipage}[c]{0.55\linewidth}
    \centering
    \begin{minipage}{0.48\linewidth}
        \centering
        \includegraphics[width=\linewidth]{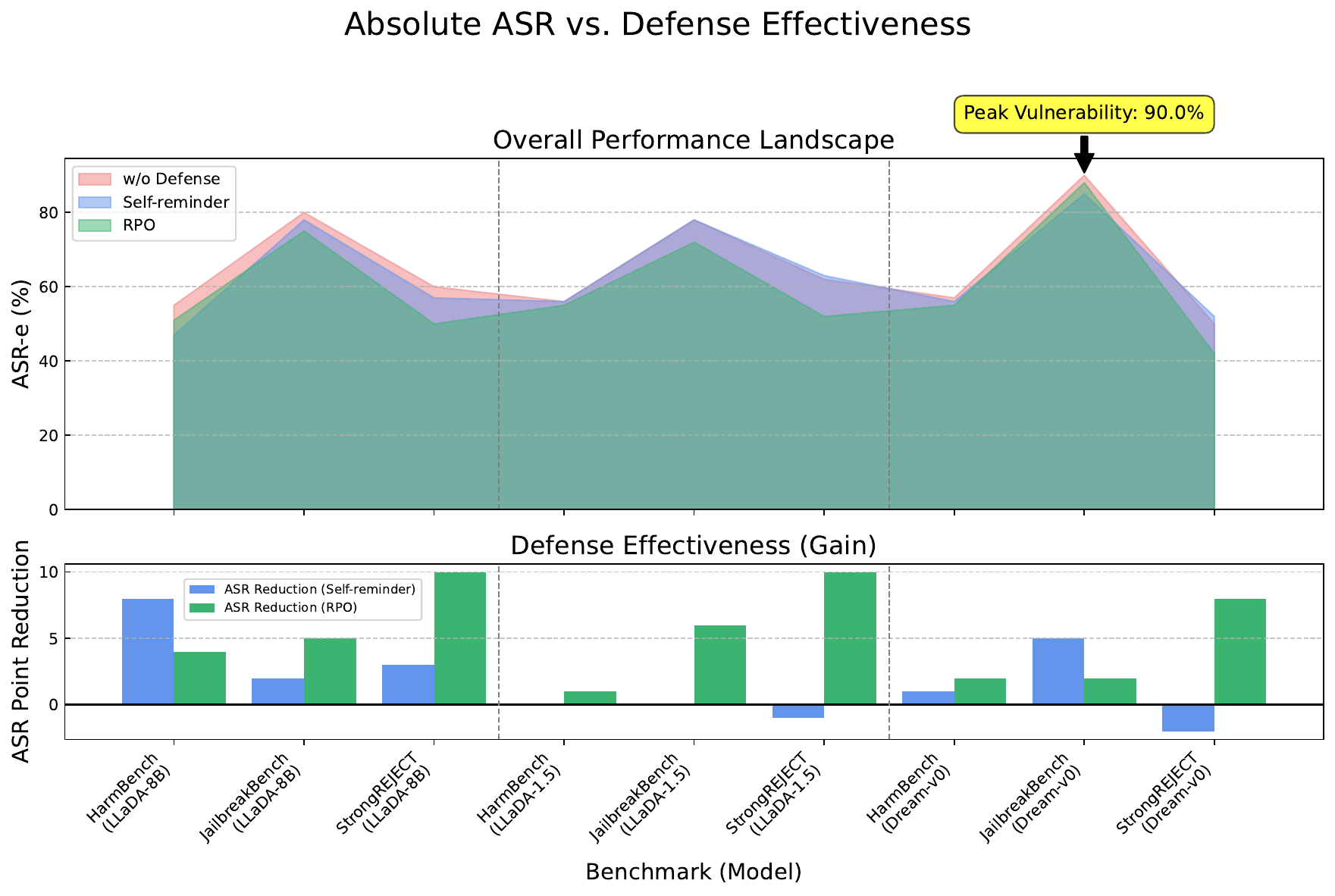} 
        \vspace{2pt} \\ \scriptsize \textit{Ground Truth}
    \end{minipage}\hfill
    \begin{minipage}{0.48\linewidth}
        \centering
        \includegraphics[width=\linewidth]{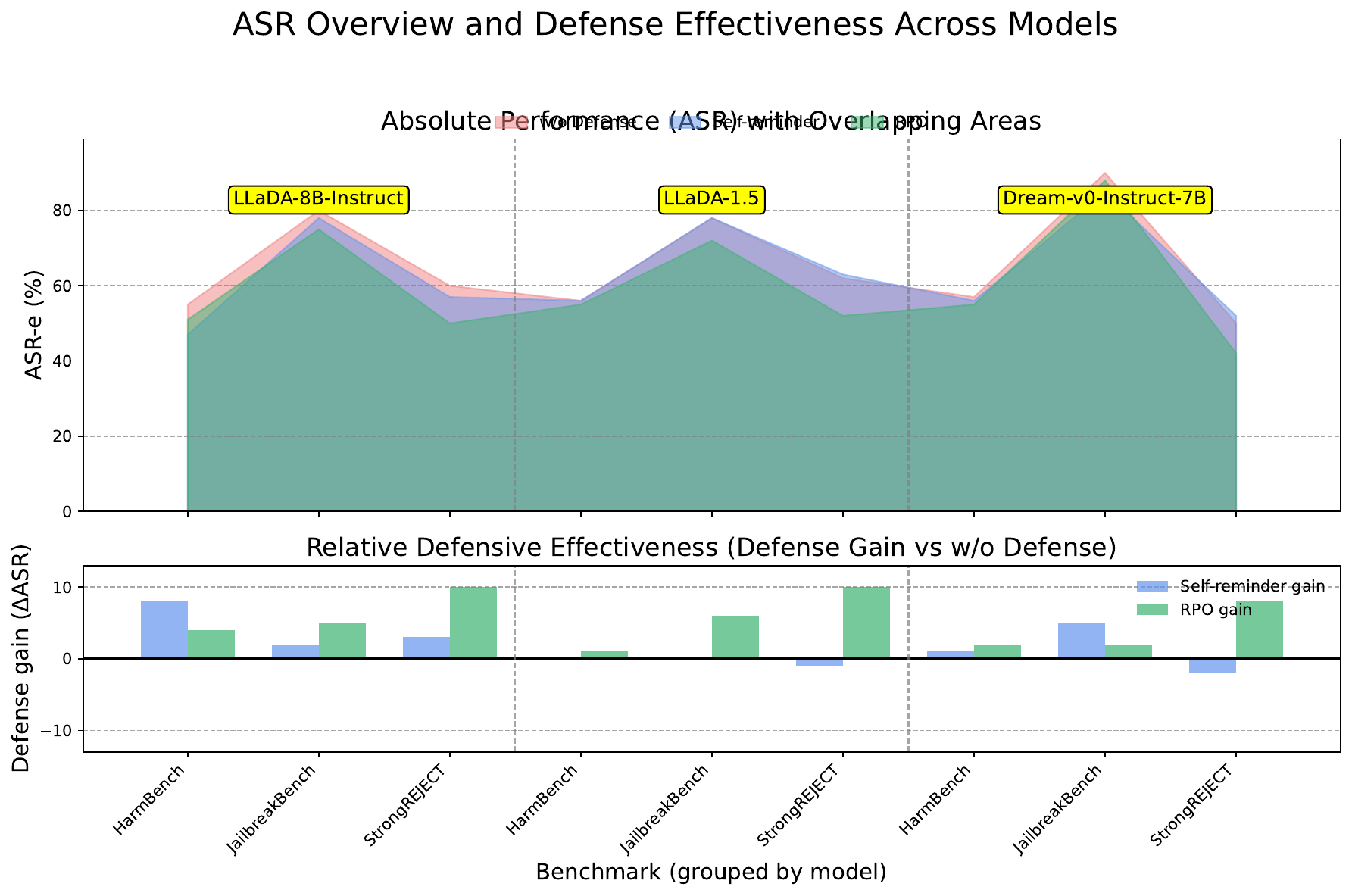} 
        \vspace{2pt} \\ \scriptsize \textit{Model Generated}
    \end{minipage}
\end{minipage}\hfill
\begin{minipage}[c]{0.42\linewidth}
    \scriptsize
    \textbf{[Dimension 7: Text Occlusion]} \\
    \textit{Model:} GPT-5.2\\
    \textbf{Observation:} The generated output exhibits severe overlap between the legend and title, significantly impairing visual aesthetics and chart readability. This placement error demonstrates a failure to manage spatial constraints among key textual components.
\end{minipage}

\end{tcolorbox}

By systematically analyzing these errors in model outputs, we categorized the systemic weaknesses of current LMMs and derived the three macro-dimensions and seven fine-grained flaw metrics utilized in our benchmark:

\begin{itemize}
    \item \textbf{Visual Effects:} Models frequently struggle with aesthetic parameterization. They often deploy indistinguishable or counter-intuitive color palettes (\textit{Color Consistency}), generate elements with improper filling patterns and disproportionate styles (\textit{Visual Style}), or set unbalanced figure dimensions that lead to severe visual distortion (\textit{Aspect Ratio}).
    
    \item \textbf{Layout Structure:} Spatial awareness remains a significant bottleneck. We observed severe spatial collisions, such as subplots or legends occluding underlying data (\textit{Component Positioning}), or overlap between key textual components that impairs readability (\textit{Text Occlusion}).
    
    \item \textbf{Data Fidelity:} Current LMMs exhibit significant limitations in precise visual data extraction and numerical representation. Due to their weak capacity for extracting exact data points from visual inputs, models often generate flawed data structures. This results in missing or mismatched data series on the plot that are erroneously retained in the legend (\textit{Data Encoding}). Furthermore, inaccurate data parsing frequently leads to incorrect \texttt{ylim} settings that artificially clip the maximum data points (\textit{Axis Scaling}).
\end{itemize}

This empirical analysis confirms that our selected flaw dimensions are not artificially contrived, but directly mirror the systemic weaknesses of current LMMs. Consequently, \benchmark is explicitly designed to evaluate whether LMMs acting as reward models can reliably detect and penalize these errors in model outputs.

\section{Dataset Construction Pipeline}
\label{app:dataset_construction}

\subsection{Pipeline Overview}
\label{app:data_annotation}

The benchmark is produced through a four-stage pipeline:
\begin{enumerate}
    \item \textbf{Source collection.} We assemble executable chart programs from newly collected scientific figures and established chart-to-code benchmarks, then verify that every source chart renders correctly.
    \item \textbf{Candidate generation.} For CPA, controlled transformations introduce one target visual defect at a time according to the seven diagnostic criteria. For CRJ, candidate outputs are generated for chart reproduction and chart editing requests.
    \item \textbf{Quality screening.} Expert review removes non-renderable outputs, ambiguous examples, unrealistic corruptions, and cases whose labels cannot be determined reliably. Surviving samples are checked for task compliance and visual validity.
    \item \textbf{Standardization.} Verified examples are converted into a common evaluation format containing the task instruction, required image inputs, and target judgment label.
\end{enumerate}

This pipeline separates frequently co-occurring chart defects into controlled diagnostic cases while retaining realistic chart content. It yields deterministic labels for pairwise preference in CPA and for Accept/Reject verification in CRJ. Final task sizes and visual-distribution statistics are reported below.

\subsection{Comprehensive Prompt Library}
\label{app:prompt_library}

To ensure full reproducibility of our benchmark, we provide the complete prompt library utilized in this work. The prompts are logically divided into two categories: (1) Prompts for dataset construction (flaw injection and instruction generation), and (2) Prompts for automated model evaluation (inference). Note that strict constraints regulating the LLM's output format (e.g., returning only Python code without markdown) have been omitted for brevity.

\subsubsection{Prompts for Dataset Construction}

To systematically generate negative samples with specific, realistic visual flaws for Task 1, we utilize the following prompt library across 7 fine-grained dimensions. 

\begin{tcolorbox}[
    colback=gray!5!white, 
    colframe=MyDarkBlue, 
    title=Flaw Injection Prompts across 7 Dimensions (Task 1: CPA),
    arc=4pt, 
    boxrule=0.8pt, 
    fonttitle=\bfseries,
    width=\linewidth,
    breakable  
]
\small

\noindent\textbf{[Dimension 1: Color\_Consistency]} \\
Modify the provided Matplotlib code to introduce subtler but CRITICAL flaws in Color Consistency. Do NOT simply turn the chart black and white. Instead, use one of the following specific strategies:
\begin{itemize}
    \setlength{\itemsep}{0pt}\setlength{\parskip}{0pt}
    \item \textbf{Indistinguishable Categories:} Switch to a ``Monochromatic'' color scheme where all elements use extremely similar shades (e.g., 5 almost identical shades of blue).
    \item \textbf{Low Visibility:} Change data colors to \texttt{'lightyellow'}, \texttt{'lightcyan'}, or \texttt{'lavender'} that are barely visible against a white background.
    \item \textbf{Aesthetic Clashing:} Use ``muddy'' colors (dark olive, brown, dark purple) mixed together, OR Create a Legend Mismatch (legend colors differ from data colors).
    \item \textbf{Counter-Intuitive Colors:} Use cool colors (blue) for ``Danger/High'' and warm colors (red) for ``Safe/Low''.
\end{itemize}
\textbf{Constraint:} The code must remain runnable. Chart type and data must remain unchanged.

\vspace{0.5em}\hrule\vspace{0.8em}

\noindent\textbf{[Dimension 2: Visual\_Style]} \\
Analyze the provided Matplotlib code. Modify the styling parameters (\texttt{linewidth}, \texttt{markersize}, \texttt{s}, \texttt{alpha}, etc.) to degrade the aesthetic quality. \\
\textbf{OBJECTIVE:} Make the visual elements size/thickness Inappropriate (either too large/thick OR too small/thin), giving the chart an ``Amateurish'' or ``Unrefined'' look.
\begin{itemize}
    \setlength{\itemsep}{0pt}\setlength{\parskip}{0pt}
    \item \textbf{Strategy A:} Increase the size/width of elements significantly beyond standard defaults. They should look heavy and clumsy, but NOT so huge that they obscure the entire plot or overlap excessively.
    \item \textbf{Strategy B:} Decrease the size/width of elements or lower the \texttt{alpha} (transparency) so they appear frail and hard to see, but NOT invisible.
\end{itemize}
\textbf{CRITICAL CONSTRAINTS (The ``Not Extreme'' Rule):}
\begin{itemize}
    \item Do NOT use extreme values that break the visualization (e.g., do not set size to 0 or 1000). The chart must remain readable, just ugly/poorly scaled.
    \item You must adapt your modifications according to the code, and your modifications should be based on the parameters in the code. You must just cover 1-3 fix in a single code.
    \item Except for the parts you have modified, everything else must remain unchanged. Keep data/logic EXACTLY the same.
\end{itemize}

\vspace{0.5em}\hrule\vspace{0.8em}

\noindent\textbf{[Dimension 3: Aspect\_Ratio]} \\
Analyze the provided Matplotlib code. Modify or Add the figure size settings (\texttt{figsize}) to create a Distorted Aspect Ratio by moderately COMPRESSING one dimension.\\
\textbf{OBJECTIVE:} Intentionally shrink ONE dimension (Width or Height) to make the aspect ratio look ``Off'' or ``Unbalanced'', but ensure the chart remains fully readable and not comical.
\begin{itemize}
    \setlength{\itemsep}{0pt}\setlength{\parskip}{0pt}
    \item \textbf{Strategy A (Vertical Squashing):} Maintain the Width. Set the Height to be noticeably short. Effect: The chart looks flattened, but y-axis labels are still legible.
    \item \textbf{Strategy B (Horizontal Squeezing):} Maintain the Height. Set the Width to be noticeably narrow. Effect: The chart looks crowded horizontally, but x-axis labels don't completely overlap.
\end{itemize}
\textbf{CRITICAL CONSTRAINTS:} One Dimension Only (Do NOT shrink both). Noticeable but Safe (NOT extreme). Subplot Awareness (ensure individual subplots get enough space). Except for the \texttt{figsize} parameters, everything else must remain unchanged.

\vspace{0.5em}\hrule\vspace{0.8em}

\noindent\textbf{[Dimension 4: Axis\_Scaling]} \\
Analyze the provided Matplotlib code to introduce a Data Clipping issue.\\
\textbf{OBJECTIVE:} Manually restrict the Y-axis range (\texttt{ylim}) to be LOWER than the actual maximum value found in the hardcoded data, causing the top of the data to be cut off.
\begin{itemize}
    \setlength{\itemsep}{0pt}\setlength{\parskip}{0pt}
    \item \textbf{Identify the Maximum:} Mentally calculate or find the highest value (Max\_Value) present in that data.
    \item \textbf{Apply Clipping:} Insert or Modify \texttt{plt.ylim()}. Set the upper limit (top) to be slightly lower than the Max\_Value you found. (e.g., If Max is 100, set \texttt{plt.ylim(top=90)}).
\end{itemize}
\textbf{CRITICAL CONSTRAINTS:} Visible Clipping. Not Extreme (Do NOT set the limit so low that the chart becomes unreadable or most data disappears). Except for adding/modifying \texttt{plt.ylim}, everything else must remain unchanged.

\vspace{0.5em}\hrule\vspace{0.8em}

\noindent\textbf{[Dimension 5: Data\_Encoding]} \\
Analyze the provided Matplotlib code. Modify the data handling to create a Phantom Data error (Missing Series) using realistic data failures.\\
\textbf{OBJECTIVE:} Make ONE of the data series invisible on the chart by corrupting its data, creating a mismatch where the Legend shows the item (with correct color/style), but the plot is empty.
\begin{itemize}
    \setlength{\itemsep}{0pt}\setlength{\parskip}{0pt}
    \item \textbf{Action:} Select ONE specific data series to hide. Do NOT touch the style parameters. 
    \item \textbf{Method:} Replace the Y-axis data with a list/array of NaN. (e.g., \texttt{[float('nan')] * len(x\_data)} or \texttt{np.full(len(x\_data), np.nan)}).
\end{itemize}
\textbf{CRITICAL CONSTRAINTS:} You must import numpy as np if needed. Partial Disappearance: HIDE ONLY ONE series. The others must remain visible.

\vspace{0.5em}\hrule\vspace{0.8em}

\noindent\textbf{[Dimension 6: Component\_Positioning]} \\
Analyze the provided Matplotlib code. Modify the positioning of components (Legend, Title, or Subplots) to create a realistic Overlap/Occlusion issue using standard settings.
\begin{itemize}
    \setlength{\itemsep}{0pt}\setlength{\parskip}{0pt}
    \item \textbf{Strategy A (The Bad Legend Spot):} Choose a STANDARD \texttt{loc} parameter (e.g., \texttt{'upper right'}) that coincides with the data's location. Ensure the legend frame is opaque so it physically covers the lines/points behind it.
    \item \textbf{Strategy B (The Sunken Title):} Lower the Title position slightly so it touches or overlaps the top of the data/grid (e.g., \texttt{y=0.9} or \texttt{0.95}).
    \item \textbf{Strategy C (The Subplot Squeeze):} If multiple subplots exist, remove \texttt{plt.tight\_layout()} and add \texttt{plt.subplots\_adjust(hspace=0.0, wspace=0.0)} to cause text collision.
\end{itemize}
\textbf{CRITICAL CONSTRAINTS:} Natural Clutter (Use standard keywords, do not use extreme/negative coordinates). Visibility (The component must remain visible, just poorly placed).

\vspace{0.5em}\hrule\vspace{0.8em}

\noindent\textbf{[Dimension 7: Text\_Occlusion]} \\
Analyze the provided Matplotlib code. Modify the properties of text elements (Tick Labels, Axis Labels, or Annotations) to create Text Overlap or Occlusion issues.
\begin{itemize}
    \setlength{\itemsep}{0pt}\setlength{\parskip}{0pt}
    \item \textbf{Strategy A (The Tick Collision):} Force an improper rotation or alignment. Set \texttt{rotation=0} for long X-axis labels that clearly need rotation, OR set \texttt{rotation=90} but align them incorrectly.
    \item \textbf{Strategy B (The Suffocating Label):} Set the \texttt{labelpad} parameter to 0 or a very small negative number inside \texttt{plt.xlabel()} or \texttt{plt.ylabel()}.
    \item \textbf{Strategy C (The Lazy Annotation):} Place text at the EXACT (x, y) coordinates of the data point and set \texttt{ha='center', va='center'}.
    \item \textbf{Strategy D (The Tangled Web - RADAR only):} Reduce the label padding or radial distance so outer labels overlap with the boundary.
\end{itemize}
\textbf{CRITICAL CONSTRAINTS:} No Invisibility (Do NOT use \texttt{alpha=0}). Simulate common errors in model outputs.
\end{tcolorbox}

\vspace{1em}

\begin{tcolorbox}[
    colback=gray!5!white, 
    colframe=MyDarkBlue, 
    title=Prompt Template for Task 2 CRJ Chart Reproduction, 
    arc=4pt, 
    boxrule=0.8pt, 
    fonttitle=\bfseries,
    width=\linewidth,
    breakable
]
\small
\textbf{System:} You are a Python developer proficient in data visualization (Matplotlib, Seaborn, etc.). Your task is to generate Python code that perfectly reproduces the provided chart image.

\noindent\textbf{REQUIREMENTS:}
\begin{enumerate}
    \setlength{\itemsep}{0pt}
    \setlength{\parskip}{0pt}
    \item \textbf{Data Extraction:} Infer and extract the actual underlying data strictly based on the visual features of the provided plot.
    \item \textbf{Precise Recreation:} Generate code that reproduces the image EXACTLY as it appears, preserving:
    \begin{itemize}
        \item Plot type (scatter, line, bar, etc.)
        \item Axis labels, titles, and gridlines
        \item Exact colors, markers, line styles, and other visual aesthetics
        \item Legends and specific text annotations
    \end{itemize}
    \item \textbf{Self-contained Code:} The Python script must be completely standalone and executable, requiring no external data files or undefined variables.
\end{enumerate}

\vspace{0.5em}\hrule\vspace{0.5em}
\noindent\textbf{OUTPUT FORMAT:} \\
Output ONLY the fully executable Python script enclosed strictly within \texttt{```python} and \texttt{```} tags. Do not include any other explanations.
\end{tcolorbox}
\vspace{1em}

\begin{tcolorbox}[
    colback=gray!5!white, 
    colframe=MyDarkBlue, 
    title=Prompt Template for Task 3 CRJ Chart editing, 
    arc=4pt, 
    boxrule=0.8pt, 
    fonttitle=\bfseries,
    width=\linewidth,
    breakable 
]
\small
\textbf{System:} You are a Dataset Generator for Chart Editing. Your task is to analyze Python Code and generate a User Edit Request + 3 Code Variations.

\noindent\textbf{INPUT ANALYSIS \& ADAPTATION:}
\begin{enumerate}
    \setlength{\itemsep}{0pt}
    \setlength{\parskip}{0pt}
    \item Analyze the code to understand the chart content.
    \item Based on the chart type, generate a request that fits the TARGET SCENARIO. Constraint: Do NOT copy the examples verbatim. Create a NEW request specific to this code.
\end{enumerate}
\textbf{TARGET SCENARIO:} \{scenario\_desc\}

\vspace{0.5em}\hrule\vspace{0.5em}
\noindent\textbf{OUTPUT FORMAT (Strict Custom Tags):}
\begin{itemize}
    \setlength{\itemsep}{0pt}
    \setlength{\parskip}{0pt}
    \item \textbf{[USER\_QUERY]:} Write a natural, non-technical user request. NO coordinates (e.g., avoid "at 0.5, 0.5"). Use relative terms: "center", "right". NO specific Ugly styles. Just the request sentence.
    \item \textbf{[CODE\_HONEST]:} Full Python Code. Follow the user query LITERALLY and LOGICALLY. Write standard matplotlib code to fulfill the request. MUST BE RUNNABLE.
    \item \textbf{[CODE\_FAIL]:} Full Python Code. Try to follow but fail on a KEY PARAMETER. (e.g., User asks "Range 0-100", you set "0-10").
    \item \textbf{[CODE\_SIDE]:} Full Python Code. Follow query perfectly, BUT hallucinate an unwanted change. (e.g., User asks "Add labels", you add labels AND delete the grid.)
\end{itemize}
\end{tcolorbox}

\subsubsection{Prompts for Automated Evaluation (Inference)}

During the evaluation phase, we utilized specialized zero-shot prompts tailored to the specific criteria of each sub-task to instruct the large multimodal models (LMMs).

\begin{tcolorbox}[
    colback=gray!5!white, 
    colframe=MyDarkBlue, 
    title=Evaluation Prompt for Task 1: Chart Perception and Alignment, 
    arc=4pt, 
    boxrule=0.8pt, 
    fonttitle=\bfseries,
    width=\linewidth,
    breakable
]
\small
\textbf{System:} You are a professional Data Visualization Expert and Critic. Compare two charts (Image A and Image B) and identify which one is of higher quality.

\noindent\textbf{Key Evaluation Criteria:}
\begin{enumerate}
    \setlength{\itemsep}{0pt}
    \setlength{\parskip}{0pt}
    \item \textbf{Data Integrity:} No missing data points or misleading scales.
    \item \textbf{Layout \& Visibility:} No occlusion (legends blocking data), text overlapping, or truncation.
    \item \textbf{Aesthetics:} Clean design and professional look.
\end{enumerate}
\end{tcolorbox}

\vspace{1em}

\begin{tcolorbox}[
    colback=gray!5!white, 
    colframe=MyDarkBlue, 
    title=Evaluation Prompt for Task 2: CRJ Chart Reproduction, 
    arc=4pt, 
    boxrule=0.8pt, 
    fonttitle=\bfseries,
    width=\linewidth,
    breakable
]
\small
\textbf{System:} You are a strict Data Visualization QA Specialist. Your task is to evaluate whether the [Generated Image] is a successful reproduction of the [Ground Truth Image].

\noindent\textbf{Task Context:} The user provided the [Ground Truth Image] and asked: ``\{SIMPLIFIED\_QUERY\}''

\noindent\textbf{Evaluation Criteria:}
\begin{enumerate}
    \setlength{\itemsep}{0pt}
    \setlength{\parskip}{0pt}
    \item \textbf{Data Accuracy:} Do the bars/lines/points represent the exact same values and trends?
    \item \textbf{Visual Elements:} Are colors, markers, line styles, and legends consistent?
    \item \textbf{Text:} Are titles, labels, and ticks correct?
    \item \textbf{Visual Quality:} Is the chart free from obvious defects like text overlapping, occlusion (elements blocking data), or truncation?
\end{enumerate}

\noindent\textbf{Decision Logic:}
\begin{itemize}
    \setlength{\itemsep}{0pt}
    \setlength{\parskip}{0pt}
    \item \textbf{Accept:} If the Generated Image is a high-quality reproduction (minor, non-semantic pixel differences are okay) AND passes all 4 criteria.
    \item \textbf{Reject:} If there are data errors, missing components (legend/title), wrong chart type, OR any visual defects.
\end{itemize}
\end{tcolorbox}

\vspace{1em}

\begin{tcolorbox}[
    colback=gray!5!white, 
    colframe=MyDarkBlue, 
    title=Evaluation Prompt for Task 3: CRJ Chart Editing, 
    arc=4pt, 
    boxrule=0.8pt, 
    fonttitle=\bfseries,
    width=\linewidth,
    breakable
]
\small
\textbf{System:} You are a strict Data Visualization QA Judge. Your task is to verify if an image editing request was performed correctly, safely, and without introducing new defects.

\noindent\textbf{You will be provided with:}
\begin{enumerate}
    \setlength{\itemsep}{0pt}
    \setlength{\parskip}{0pt}
    \item \textbf{User Instruction:} What the user wanted to change.
    \item \textbf{Image 1 (Original):} The chart before editing.
    \item \textbf{Image 2 (Result):} The chart after editing.
\end{enumerate}

\noindent\textbf{Evaluation Criteria (Must pass ALL to Accept):}
\begin{enumerate}
    \setlength{\itemsep}{0pt}
    \setlength{\parskip}{0pt}
    \item \textbf{Instruction Adherence:} Did the model execute the User Instruction correctly? (e.g., color changed, title added).
    \item \textbf{Data Integrity:} Did the model preserve the correctness of the data and chart structure? (No missing bars, no changed values unless requested).
    \item \textbf{Visual Quality Check:} Did the editing process introduce NEW visual defects that were NOT in the Original Image?
    \item \textbf{Consistency with Original:} Are all other parts of the generated image consistent with the Original image (excluding the edited parts as per instruction)?
\end{enumerate}

\noindent\textbf{Decision Logic:}
\begin{itemize}
    \setlength{\itemsep}{0pt}
    \setlength{\parskip}{0pt}
    \item \textbf{Accept:} IF the instruction is followed, data is preserved, no new visual defects are introduced AND consistency with original.
    \item \textbf{Reject:} IF the instruction is ignored, data is corrupted, OR if the result looks visually worse than the original.
\end{itemize}
\end{tcolorbox}

\begin{figure}[t!]
    \centering
    \includegraphics[width=\linewidth]{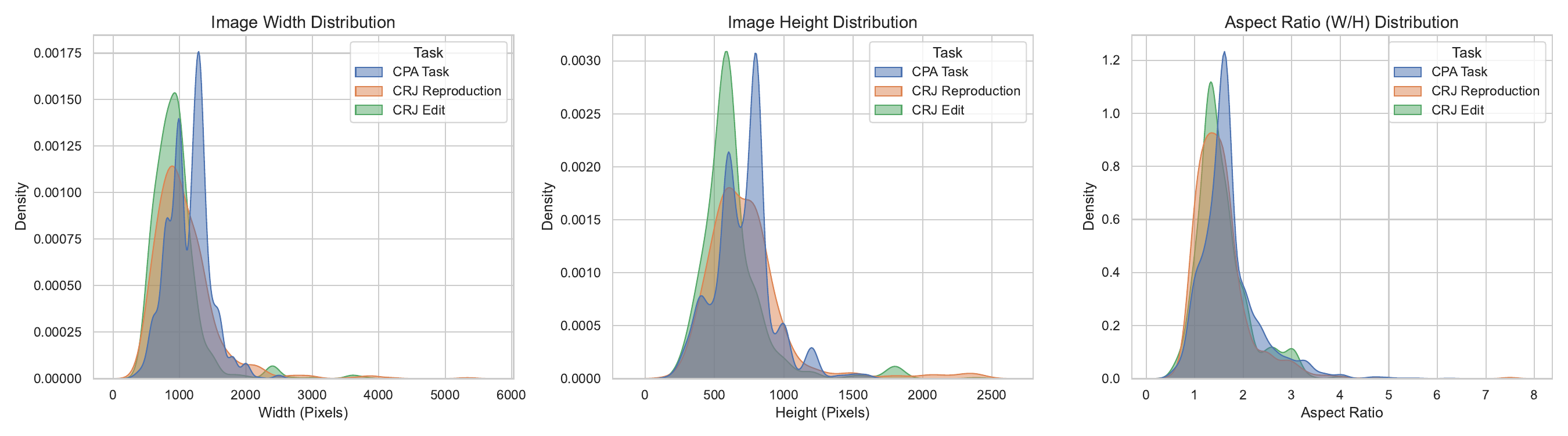}
    \caption{\textbf{Distribution of Image Dimensions.} The KDE plots demonstrate a wide and consistent coverage of widths, heights, and aspect ratios across the three benchmark tasks.}
    \label{fig:dimension_stats}

    \includegraphics[width=0.6\linewidth]{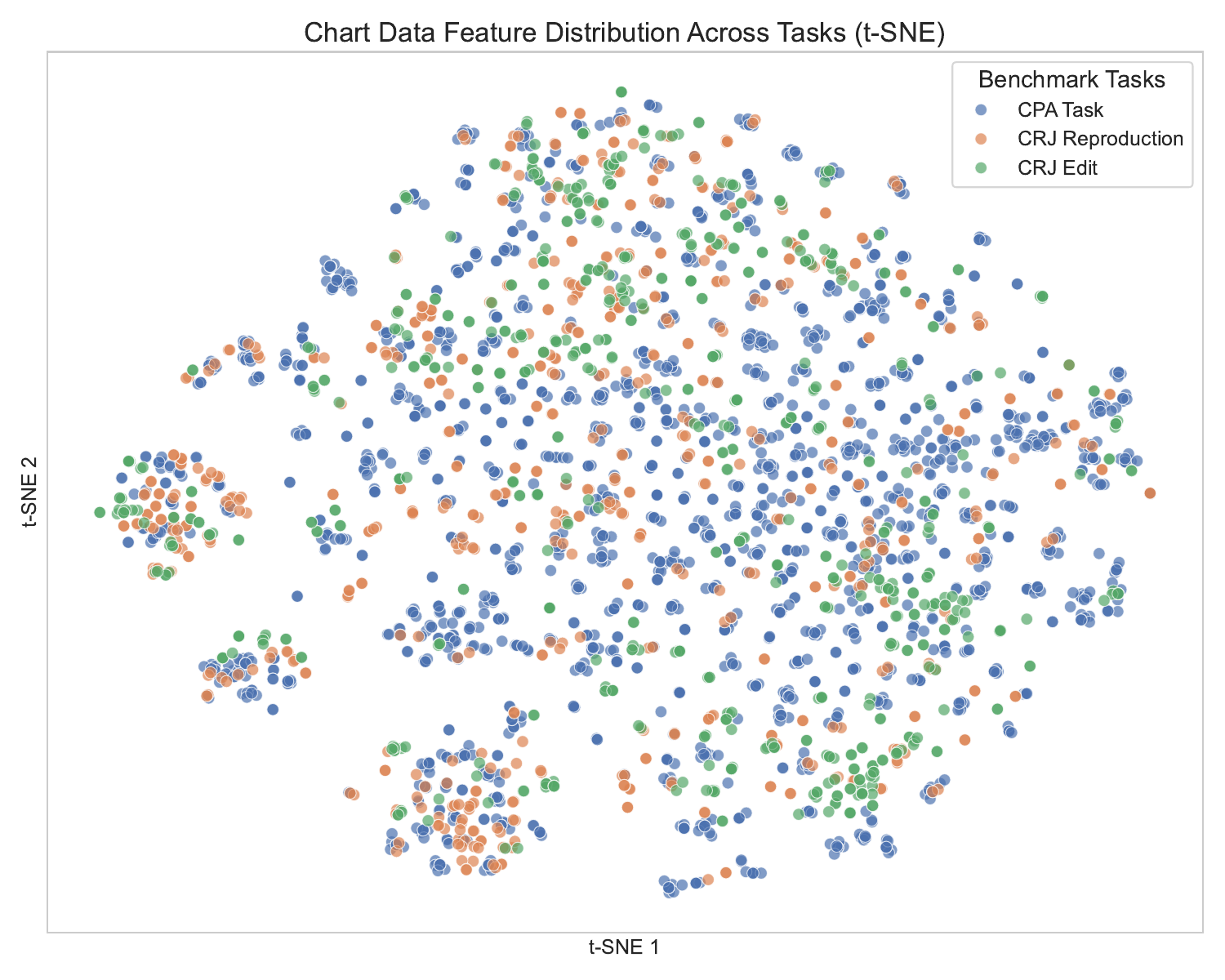} 
    \caption{\textbf{t-SNE Visualization of Chart Image Features.} The scatter plot shows that chart images from the CPA, Reproduction, and Edit tasks are deeply intertwined across the visual feature space, indicating a unified visual domain without task-specific clustering.}
    \label{fig:tsne_stats}
\end{figure}
\subsection{Data Statistics and Visual Diversity}
\label{app:data_statistics}

\noindent\textbf{Benchmark scale and evaluation efficiency.}
\benchmark contains 1,653 unique evaluation tasks: 1,003 CPA cases and 650 CRJ cases, with the latter comprising 315 chart-reproduction and 335 chart-editing instances. The final benchmark is obtained through the screening and verification pipeline described in Section~\ref{app:data_annotation}. Its scale is comparable to representative multimodal reward-model and general-purpose evaluation benchmarks, including VL-RewardBench (1,247), VC-RewardBench (1,335), and MMStar (1,500). The resulting compact design balances broad diagnostic coverage with rapid and cost-effective evaluation, which is important when repeatedly benchmarking a large collection of proprietary and open-source models.

To validate the diversity and fairness of the \benchmark benchmark, we analyze the structural dimensions and visual feature distributions of the curated chart images.

Figure~\ref{fig:dimension_stats} presents the kernel density estimation (KDE) for image widths, heights, and aspect ratios. The wide span of resolutions confirms the structural diversity of the dataset. Importantly, the overlapping distributions across the three sub-tasks (CPA, CRJ Reproduction, and CRJ Edit) indicate consistent data formatting without task-specific size biases.

Furthermore, Figure~\ref{fig:tsne_stats} visualizes the deep visual features of the chart images using t-SNE. The highly intertwined distribution of data points confirms that the charts across all three sub-tasks share a unified and highly diverse visual domain. The absence of task-specific visual clusters indicates consistent visual complexity throughout the benchmark.

\clearpage
\subsection{Chart-Type Evaluation Results}
\label{app:chart_type_results}

Table~\ref{tab:chart_types} breaks down Chart Perception Alignment performance across nine chart types. At the model-averaged level, 3D charts obtain the lowest score (26.6), while radar charts (32.8) are also more challenging than common bar and scatter charts. Nevertheless, the average score remains below 50 for every chart type, ranging from 26.6 to 41.9. This broad performance gap indicates that \benchmark does not derive its difficulty from only a few complex chart families; instead, fine-grained errors are distributed across all evaluated chart types.

\begin{supplementtable}
    \caption{\textbf{Evaluation results on Chart Perception Alignment Tasks.} The comparison results on diverse chart types.}
    \label{tab:chart_types}
    \small
    \renewcommand{\arraystretch}{0.82}
    \setlength{\tabcolsep}{2.8pt}
    \begin{tabular}{@{}l|ccccccccc|c@{}}
        \toprule
        \textbf{Model} & \textbf{Bar} & \textbf{Scatter} & \textbf{Area} & \textbf{Radar} & \textbf{Box} & \textbf{Line} & \textbf{3d} & \textbf{Density} & \textbf{Hist} & \textbf{Avg.} \\
        \midrule
        \multicolumn{11}{l}{\textbf{Proprietary}} \\
        \textbf{Gemini-3-Pro} & \textbf{97.2} & \textbf{97.3} & \textbf{93.1} & \textbf{94.7} & \textbf{98.2} & \textbf{86.0} & \textbf{78.7} & \textbf{97.9} & \textbf{95.3} & \textbf{93.2} \\
        GPT-5.2 & 86.9 & 82.7 & 74.1 & 77.2 & 76.4 & 64.0 & 70.2 & 80.9 & 72.1 & 76.1 \\
        Seed1.6-VL & 67.3 & 80.0 & 58.6 & 61.4 & 69.1 & 52.0 & 42.6 & 59.6 & 67.4 & 62.0 \\
        Claude-Sonnet-4.5 & 52.3 & 62.7 & 65.5 & 59.6 & 56.4 & 40.0 & 42.6 & 59.6 & 44.2 & 53.7 \\
        \midrule
        \multicolumn{11}{l}{\textbf{Open-Source LMMs (thinking)}} \\
        MiMo-VL-SFT & 43.9 & 53.3 & 43.1 & 49.1 & 45.5 & 42.0 & 34.0 & 38.3 & 39.5 & 43.2 \\
        Qwen3-VL-30B & 37.4 & 32.0 & 32.8 & 17.5 & 38.2 & 18.0 & 6.4 & 40.4 & 32.6 & 28.4 \\
        MiMo-VL-RL & 18.7 & 41.3 & 25.9 & 21.1 & 32.7 & 18.0 & 27.7 & 25.5 & 30.2 & 26.8 \\
        LLaVA-Critic-R1 & 5.6 & 9.3 & 10.3 & 1.8 & 5.5 & 6.0 & 4.3 & 6.4 & 4.7 & 6.0 \\
        ThinkLite-VL-7B & 0.0 & 6.7 & 1.7 & 3.5 & 9.1 & 4.0 & 10.6 & 8.5 & 9.3 & 5.9 \\
        \midrule
        \multicolumn{11}{l}{\textbf{Open-Source LMMs (non-thinking)}} \\
        Qwen2-VL-7B & 15.0 & 12.0 & 32.8 & 7.0 & 14.5 & 18.0 & 0.0 & 12.8 & 9.3 & 13.5 \\
        Qwen2-VL-72B & 44.9 & 50.7 & 34.5 & 24.6 & 29.1 & 30.0 & 23.4 & 29.8 & 39.5 & 34.1 \\
        DeepSeek-VL & 0.0 & 0.0 & 0.0 & 0.0 & 0.0 & 0.0 & 0.0 & 0.0 & 0.0 & 0.0 \\
        Qwen2.5-VL-7B & 2.8 & 5.3 & 3.4 & 0.0 & 1.8 & 0.0 & 2.1 & 8.5 & 9.3 & 3.7 \\
        Qwen2.5-VL-72B & 61.7 & 66.7 & 58.6 & 45.6 & 54.5 & 40.0 & 61.7 & 53.2 & 62.8 & 56.1 \\
        GLM-4V & 37.4 & 47.0 & 38.8 & 25.0 & 25.5 & 29.3 & 19.5 & 17.0 & 34.3 & 30.4 \\
        MiMo-VL-SFT (NoThink) & 36.4 & 69.3 & 41.4 & 36.8 & 36.4 & 46.0 & 27.7 & 57.4 & 39.5 & 43.4 \\
        MiMo-VL-RL (NoThink) & 34.6 & 60.0 & 34.5 & 26.3 & 29.1 & 34.0 & 31.9 & 46.8 & 51.2 & 38.7 \\
        Kimi-VL & 37.4 & 34.7 & 37.9 & 42.1 & 30.9 & 32.0 & 21.3 & 36.2 & 25.6 & 33.1 \\
        InternVL2.5-38B & 19.6 & 28.0 & 15.5 & 10.5 & 34.5 & 14.0 & 19.1 & 10.6 & 20.9 & 19.2 \\
        InternVL2.5-8B & 14.0 & 16.0 & 5.2 & 31.6 & 18.2 & 8.0 & 21.3 & 2.1 & 18.6 & 15.0 \\
        Molmo & 5.6 & 5.3 & 3.4 & 1.8 & 9.1 & 12.0 & 10.6 & 12.8 & 7.0 & 7.5 \\
        Molmo2 & 24.3 & 20.0 & 17.2 & 33.3 & 21.8 & 26.0 & 19.1 & 19.1 & 34.9 & 24.0 \\
        Qwen3-VL-8B & 55.1 & 58.7 & 41.4 & 36.8 & 43.6 & 20.0 & 21.3 & 40.4 & 25.6 & 38.1 \\
        Qwen3-VL-32B & 79.4 & 81.3 & 65.5 & 64.9 & 69.1 & 64.0 & 31.9 & 76.6 & 55.8 & 65.4 \\
        InternVL3.5-8B & 9.3 & 16.0 & 10.3 & 15.8 & 21.8 & 14.0 & 21.3 & 12.8 & 14.0 & 15.0 \\
        InternVL3.5-38B & 56.1 & 53.3 & 41.4 & 64.9 & 38.2 & 38.0 & 42.6 & 38.3 & 34.9 & 45.3 \\
        \midrule
        \textbf{Average} & \textbf{36.3} & \textbf{41.9} & \textbf{34.1} & \textbf{32.8} & \textbf{35.0} & \textbf{29.0} & \textbf{26.6} & \textbf{34.3} & \textbf{33.8} & \textbf{33.8} \\
        \bottomrule
    \end{tabular}
\end{supplementtable}

\clearpage

\section{Human Study}
\label{app:human_evaluation}
\begin{figure}[t!]
    \centering
    \includegraphics[width=0.7\linewidth]{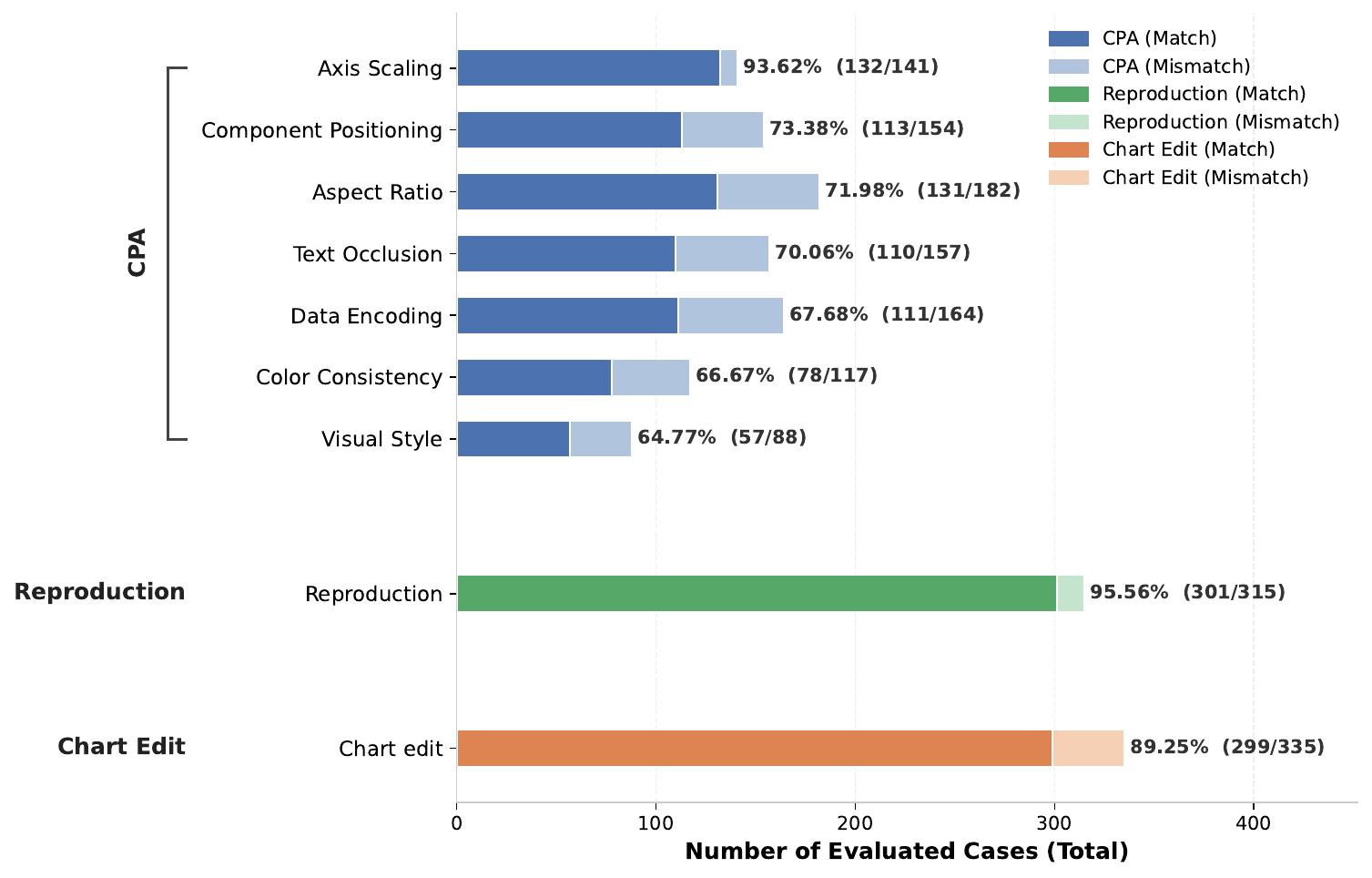}
    \caption{\textbf{Human-Benchmark Alignment Results.} Match rates across various dimensions. The gap in certain CPA categories highlights the benchmark's superior sensitivity to subtle defects compared to non-expert human participants.}
    \label{fig:human_study_results}
\end{figure}

To validate the alignment between our benchmark's deterministic criteria and human judgment, we organized a human study involving 10 undergraduate students from the School of Computer Science. Depending on the task type, participants performed different evaluation protocols:
\begin{itemize}
    \item \textbf{CPA Task:} Participants conducted a preference-based comparison. For each instance, they were presented with two charts and asked to choose from four options: \textit{Image A is better}, \textit{Image B is better}, \textit{Both are Good}, or \textit{Both are Bad}.
    \item \textbf{CRJ Tasks (Reproduction \& Editing):} Participants performed a binary quality assessment. They were asked to strictly judge whether the generated chart met the task requirements, leading to a final verdict of \textit{Accept} or \textit{Reject}.
\end{itemize}
The human-benchmark alignment results are shown in Fig.~\ref{fig:human_study_results}.

As shown in Fig.~\ref{fig:human_study_results}, while our benchmark achieves high alignment in the Reproduction (95.56\%) and Chart Editing (89.25\%) tasks, certain dimensions in the CPA task exhibit lower match rates, such as Visual Style (64.77\%) and Color Consistency (66.67\%). Qualitative review suggests that when processing large-scale data, non-expert participants are susceptible to fatigue and often overlook fine-grained visual defects, leading to a ``lenience bias'' where they accept charts with subtle flaws. 

Instead of signaling a flaw, this discrepancy underscores that our criterion-guided expert verification process yields a more discriminative and consistent gold standard than non-expert intuition. The reliability and clarity of our benchmark's criteria are further evidenced by the high performance of frontier models such as Gemini-3-Pro on the CPA task, demonstrating that while these subtle visual defects may elude a casual observer, they can still be identified by careful expert review and state-of-the-art multimodal judges.

\section{Downstream Utility and Bias Mitigation}
\label{app:downstream_utility}

\subsection{Downstream Utility with METAL}
\label{app:metal_utility}

To evaluate whether performance on \benchmark predicts practical reward-model utility, we randomly sampled 100 chart-to-code instances from ChartMimic and evaluated them with the METAL chart-to-code agent. Each of the five evaluated LMMs served as the reward model in a separate METAL run, providing feedback used by the agent to revise its generated charts. We then computed the average RM-driven score gain over the 100 sampled instances.

\begin{supplementtable}
    \caption{Relationship between CPA performance and downstream reward-model utility in METAL on 100 randomly sampled ChartMimic instances.}
    \label{tab:metal_downstream_utility}
    \small
    \setlength{\tabcolsep}{12pt}
    \renewcommand{\arraystretch}{1.05}
    \begin{tabular}{@{}lcc@{}}
        \toprule
        \textbf{Model} & \textbf{CPA Score (\%)} & \textbf{METAL Score Gain} \\
        \midrule
        Gemini-3-Pro      & \textbf{95.0} & \textbf{+7.3090} \\
        GPT-5.2           & 79.7 & +2.1713 \\
        Qwen3-VL-32B      & 68.9 & +1.5224 \\
        InternVL3.5-38B   & 47.0 & +1.7064 \\
        LLaVA-Critic-R1   & 7.8  & $-$1.4112 \\
        \bottomrule
    \end{tabular}
\end{supplementtable}

Across the five reward models, CPA Score strongly correlates with downstream Score Gain (Pearson $r=0.859$). Gemini-3-Pro obtains both the highest CPA Score and the largest improvement in METAL, whereas LLaVA-Critic-R1 has the lowest CPA Score and produces a negative score gain. These results provide evidence that \benchmark captures capabilities that matter when an LMM is used to guide chart refinement, rather than measuring diagnostic performance in isolation.

\subsection{Few-Shot Mitigation of Lenience Bias}
\label{app:few_shot_lenience}

We further examine whether in-context demonstrations can mitigate the lenience bias observed on CRJ chart editing. Table~\ref{tab:few_shot_specificity} compares zero-shot and four-shot specificity for the same five models.

\begin{supplementtable}
    \caption{Specificity on CRJ chart editing under zero-shot and four-shot evaluation. $\Delta$ is the absolute change in percentage points.}
    \label{tab:few_shot_specificity}
    \small
    \setlength{\tabcolsep}{10pt}
    \renewcommand{\arraystretch}{1.05}
    \begin{tabular}{@{}lccc@{}}
        \toprule
        \textbf{Model} & \textbf{0-shot Spec. (\%)} & \textbf{4-shot Spec. (\%)} & \textbf{$\Delta$ (pp)} \\
        \midrule
        Gemini-3-Pro      & 64.0 & \textbf{84.3} & +20.3 \\
        GPT-5.2           & 69.6 & \textbf{82.5} & +12.9 \\
        Qwen3-VL-32B      & 54.6 & \textbf{58.1} &  +3.5 \\
        InternVL3.5-38B   & 27.9 & \textbf{32.2} &  +4.3 \\
        LLaVA-Critic-R1   & \textbf{32.6} & 32.0 &  $-$0.6 \\
        \bottomrule
    \end{tabular}
\end{supplementtable}

Four of the five models achieve higher specificity in the four-shot setting, with an average absolute improvement of 8.08 percentage points. The gains are particularly large for Gemini-3-Pro and GPT-5.2, while LLaVA-Critic-R1 does not benefit from the demonstrations. Thus, few-shot prompting can reduce lenience bias for capable judges, but the improvement is not universal across models.

\section{Case Studies and Dataset Gallery}
\label{app:error_analysis}

\subsection{Case Studies}
\begin{tcolorbox}[
    colback=gray!5!white, 
    colframe=MyDarkBlue, 
    title=Typical Cases Analysis,
    arc=4pt, 
    boxrule=0.8pt, 
    fonttitle=\bfseries,
    width=\linewidth,
    breakable  
]
\small

\noindent
\begin{minipage}{\linewidth}
    \centering
    \begin{minipage}{0.47\linewidth}
        \centering
        \includegraphics[width=\linewidth]{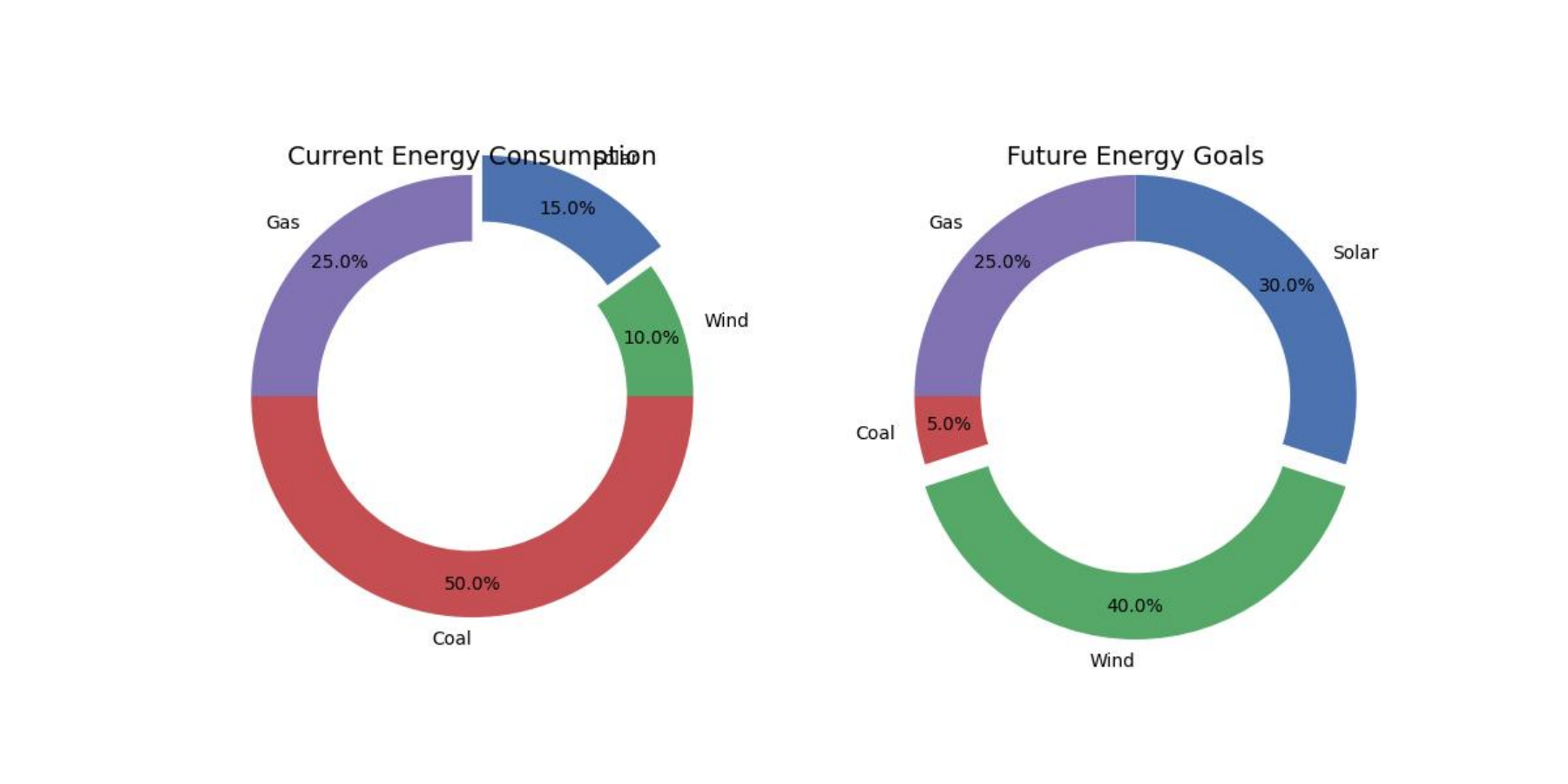}
        \\ \scriptsize \textit{Image A}
    \end{minipage}
    \hspace{0.04\linewidth} 
    \begin{minipage}{0.47\linewidth}
        \centering
        \includegraphics[width=\linewidth]{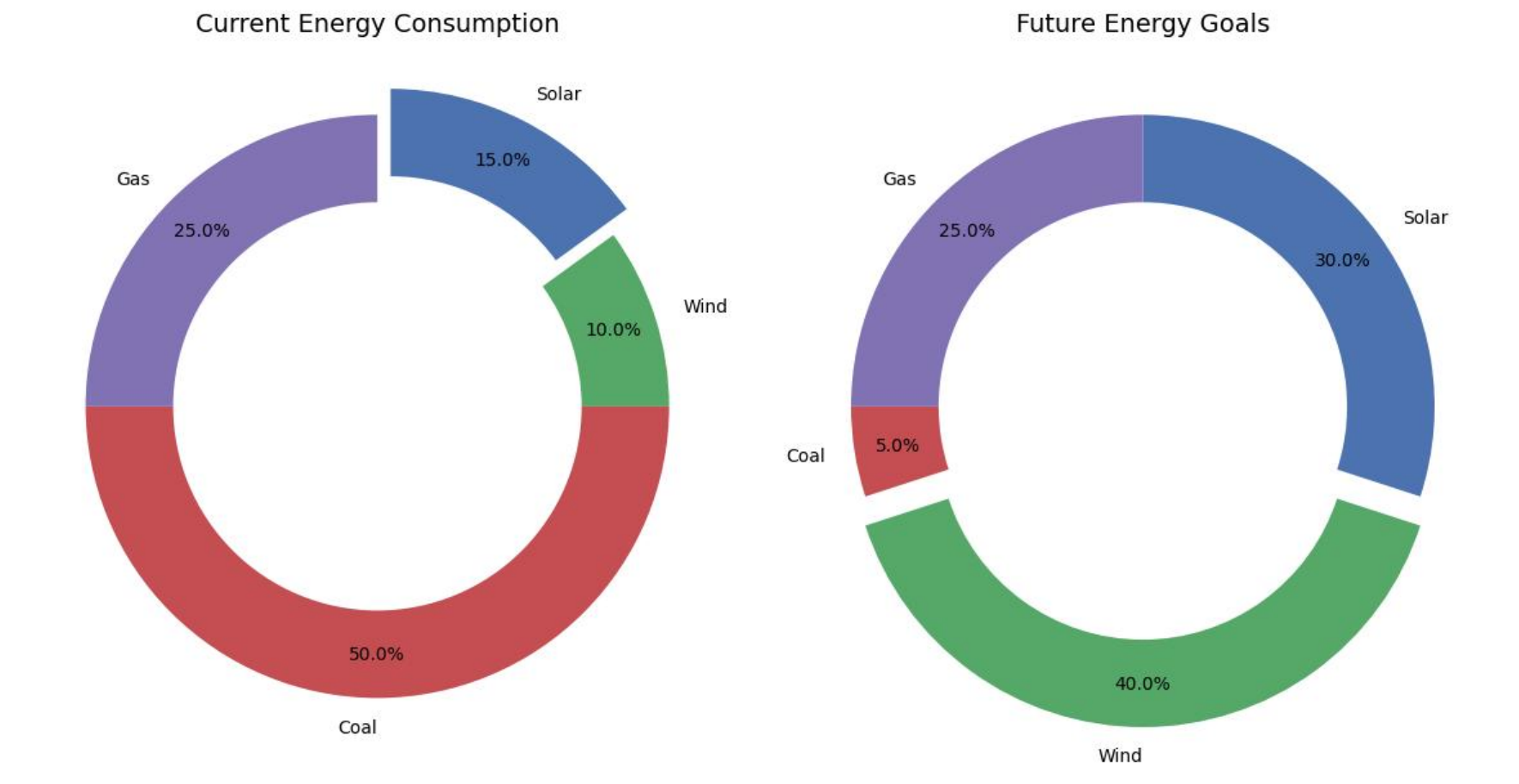}
        \\ \scriptsize \textit{Image B}
    \end{minipage}
\end{minipage}
\vspace{8pt} 
\begin{minipage}{\linewidth}
    \scriptsize
    \textbf{Error Analysis: Perception}  \\
    \textbf{Model:} Qwen3-VL-30B \hfill \textbf{Golden Answer:} Image B \\
    \vspace{2pt}
    
    \textbf{Model Answer:} ``[Answer]: Image A. [Reason]: Image A has no text overlap unlike Image B where `Wind' label partially overlaps the chart.'' \\
     \textbf{Analysis:} ``Image A has overlap between the `Current Energy Consumption' title and the 15.0 \% segment label, causing occlusion and reduced readability, while Image B is free of such occlusion and offers better readability.''
\end{minipage}
 \vspace{0.8em}\hrule\vspace{0.8em}

\begin{minipage}{\linewidth}
    \centering
    \begin{minipage}{0.47\linewidth}
        \centering
        \includegraphics[width=\linewidth]{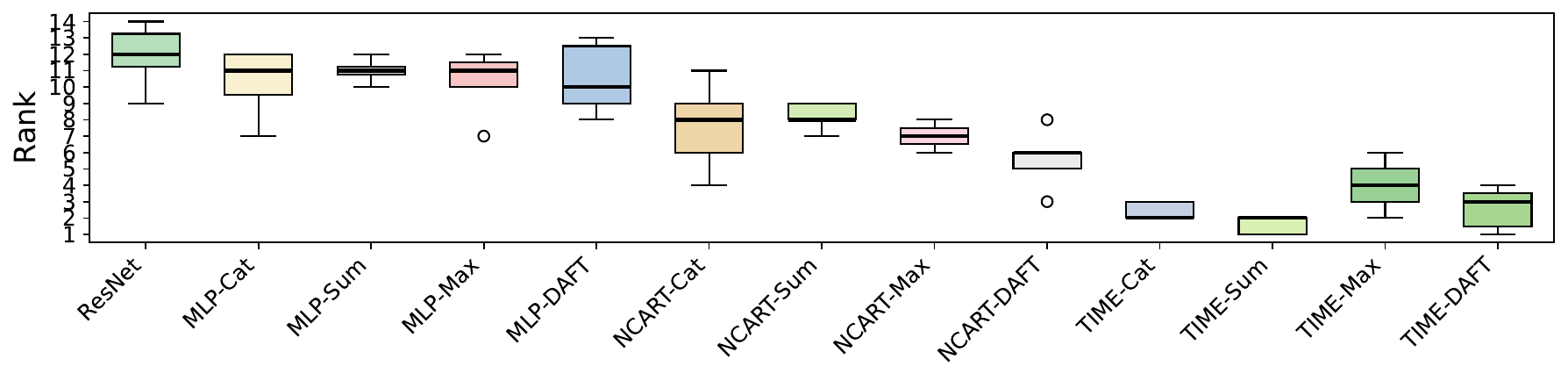}
        \\ \scriptsize \textit{Image A}
    \end{minipage}
    \hspace{0.04\linewidth} 
    \begin{minipage}{0.47\linewidth}
        \centering
        \includegraphics[width=\linewidth]{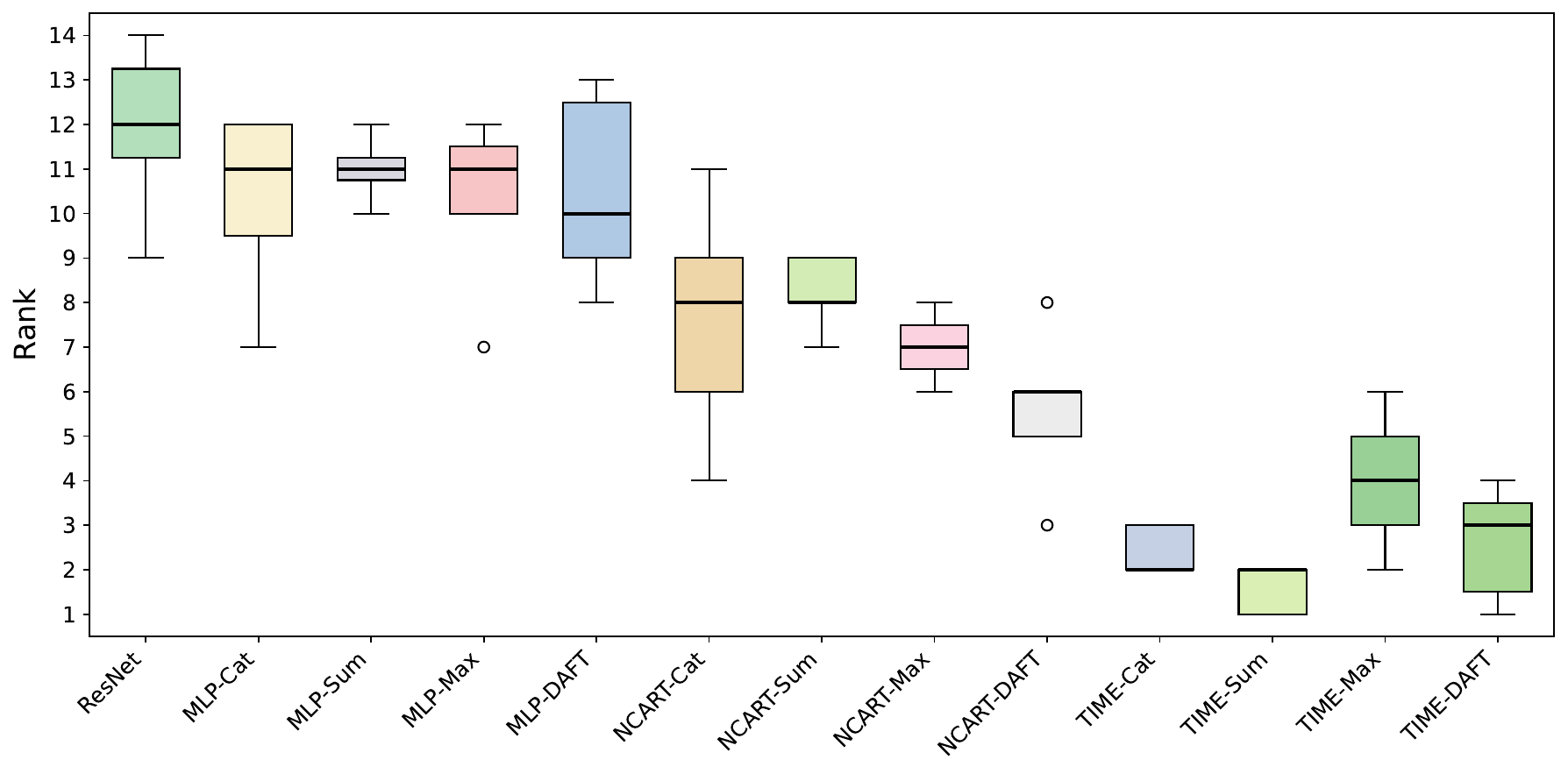}
        \\ \scriptsize \textit{Image B}
    \end{minipage}
\end{minipage}
\vspace{8pt} 
\begin{minipage}{\linewidth}
    \scriptsize
    \textbf{Error Analysis: Position Bias}  \\
    \textbf{Model:} InternVL3.5-38B \hfill \textbf{Golden Answer:} Image B \\
    \vspace{2pt}
    
    \textbf{Model Answer:} ``[Answer]: Image A. [Reason]: Layout \& Visibility: Image A has better spacing and alignment, reducing occlusion and improving readability'' \\
    \vspace{2pt}
    
    \textbf{Analysis:} ``Image A has severe overlap between the vertical `Rank' y-axis label and its tick labels, reducing readability. Image B avoids this overlap, with all axis text clearly visible.''
\end{minipage}
 \vspace{0.8em}\hrule\vspace{0.8em}

\begin{minipage}{\linewidth}
    \centering
    \begin{minipage}{0.47\linewidth}
        \centering
        \includegraphics[width=\linewidth]{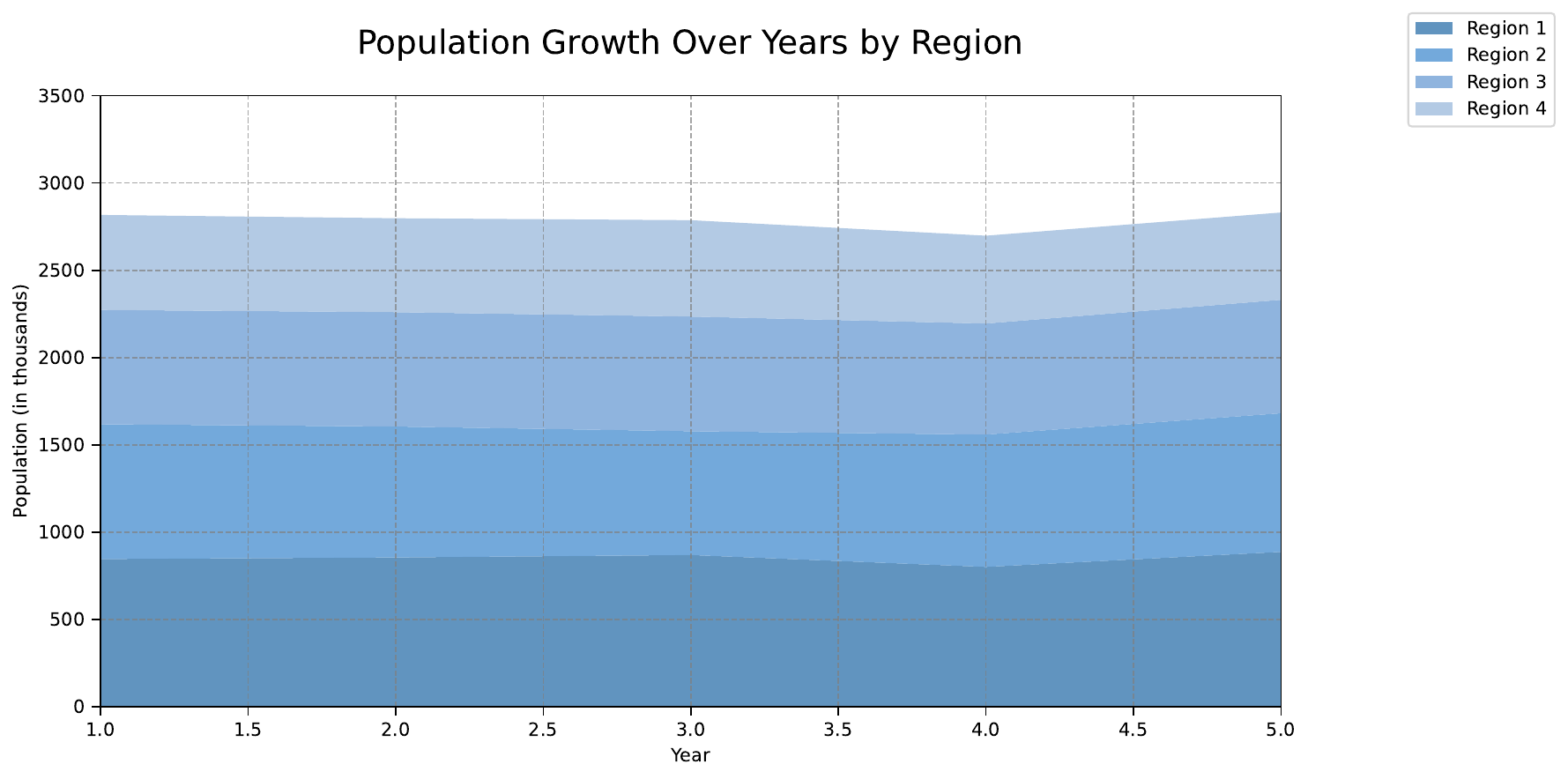}
        \\ \scriptsize \textit{Image A}
    \end{minipage}
    \hspace{0.04\linewidth} 
    \begin{minipage}{0.47\linewidth}
        \centering
        \includegraphics[width=\linewidth]{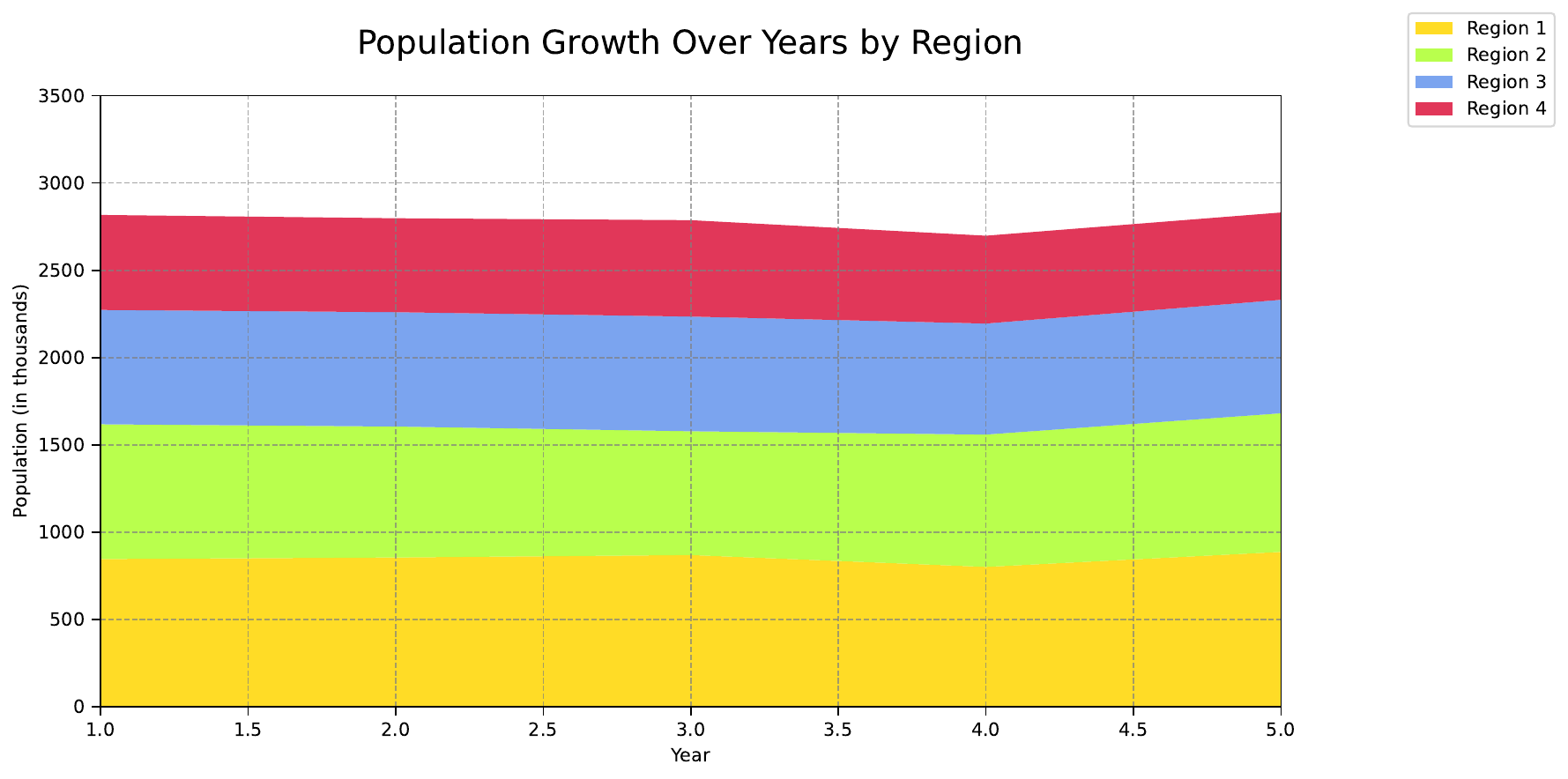}
        \\ \scriptsize \textit{Image B}
    \end{minipage}
\end{minipage}
\vspace{8pt} 
\begin{minipage}{\linewidth}
    \scriptsize
    \textbf{Error Analysis: Reference}  \\
    \textbf{Model:} Qwen2.5-VL-72B \hfill \textbf{Golden Answer:} Image B \\
    \vspace{2pt}
    
    \textbf{Model Answer:} ``[Answer]: Image A. [Reason]: Data Integrity: Image A uses distinct shades, avoiding color confusion. Layout \& Visibility: Legends do not overlap data. Aesthetics: Cleaner gradient transitions enhance readability.'' \\
    \vspace{2pt}
    
    \textbf{Analysis:} ``Image B uses monochromatic blue shades, while Image A uses distinct colors and correct spelling for clarity.''
\end{minipage}
 \vspace{0.8em}\hrule\vspace{0.8em}

\begin{minipage}{\linewidth}
    \centering
    \begin{minipage}{0.47\linewidth}
        \centering
        \includegraphics[width=\linewidth]{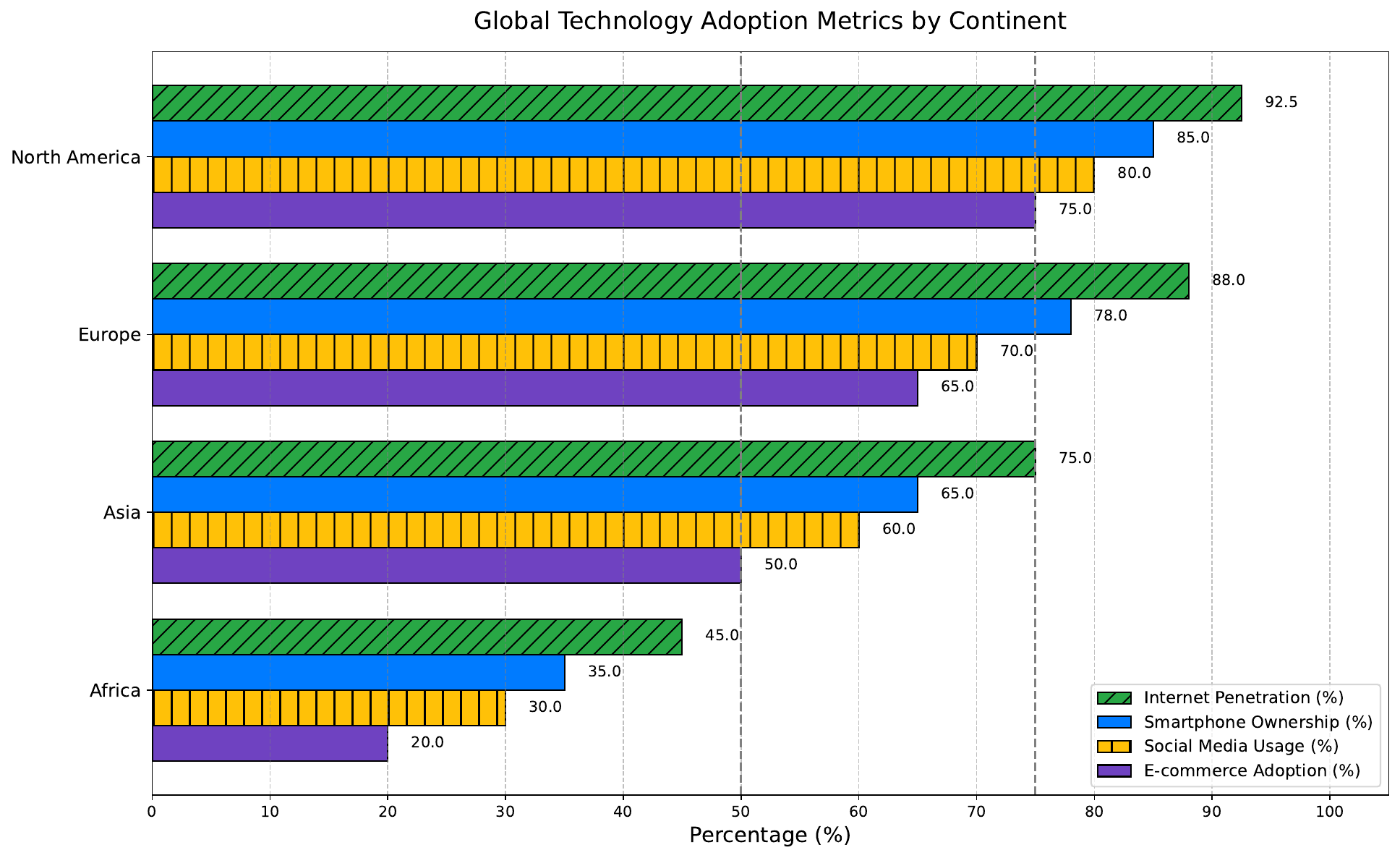}
        \\ \scriptsize \textit{Image A}
    \end{minipage}
    \hspace{0.04\linewidth} 
    \begin{minipage}{0.47\linewidth}
        \centering
        \includegraphics[width=\linewidth]{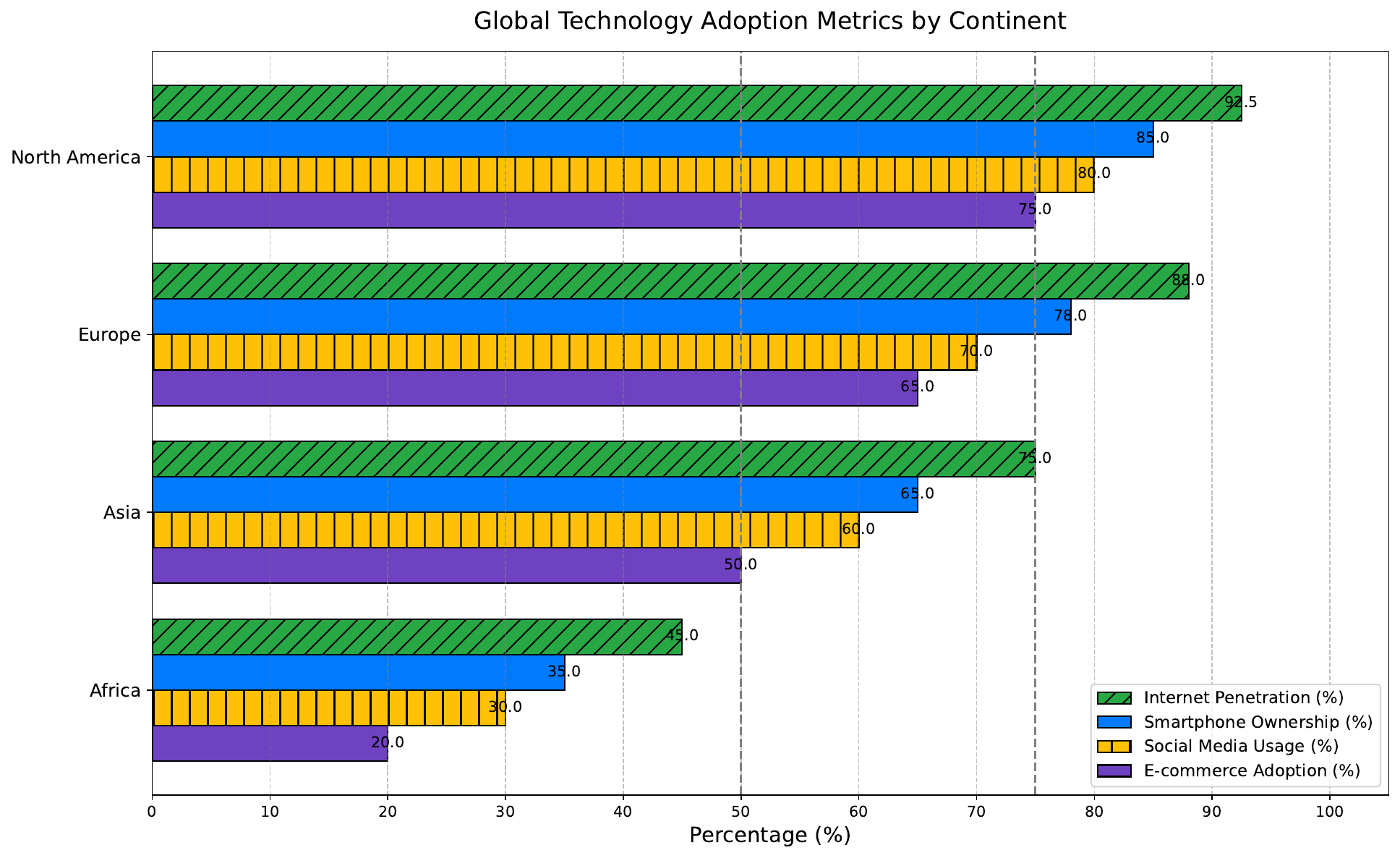}
        \\ \scriptsize \textit{Image B}
    \end{minipage}
\end{minipage}
\vspace{8pt} 
\begin{minipage}{\linewidth}
    \scriptsize
    \textbf{Error Analysis: Hallucination}  \\
    \textbf{Model:} Qwen3-VL-32B \hfill \textbf{Golden Answer:} Image A \\
    \vspace{2pt}
    
    \textbf{Model Answer:} ``[Answer]: Image B. [Reason]: Layout \& Visibility: Data labels are positioned inside bars without overlapping, ensuring clear readability and no occlusion, unlike Image A where labels extend beyond bars and may interfere with visual clarity.'' \\
    \vspace{2pt}
    
    \textbf{Analysis:} ``In Image B, the percentage labels are partially occluded by the bars, reducing readability. Image A places all labels clearly outside the bars, ensuring full visibility''
\end{minipage}
 \vspace{0.8em}\hrule\vspace{0.8em}

\begin{minipage}{\linewidth}
    \centering
    \begin{minipage}{0.47\linewidth}
        \centering
        \includegraphics[width=\linewidth]{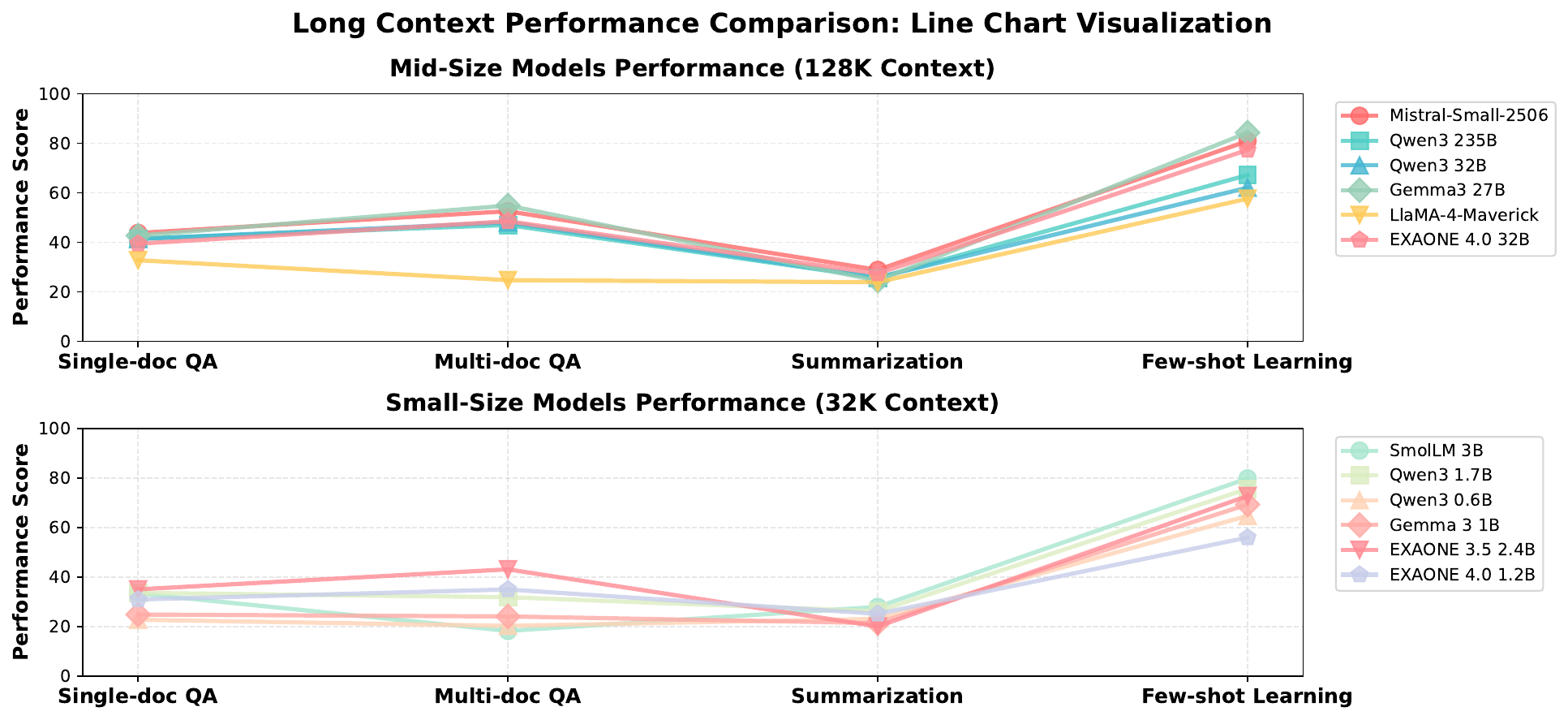}
        \\ \scriptsize \textit{Image A}
    \end{minipage}
    \hspace{0.04\linewidth} 
    \begin{minipage}{0.47\linewidth}
        \centering
        \includegraphics[width=\linewidth]{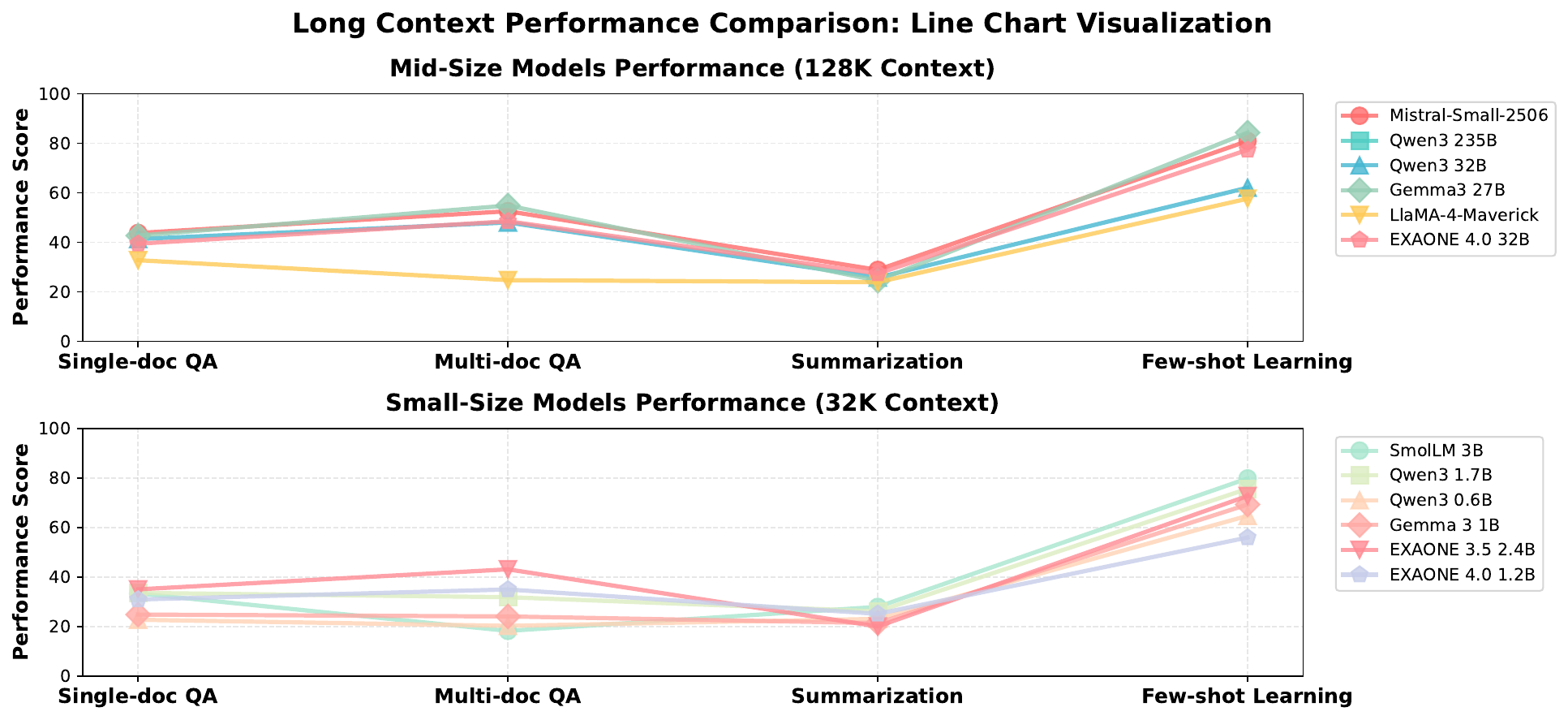}
        \\ \scriptsize \textit{Image B}
    \end{minipage}
\end{minipage}
\vspace{8pt} 
\begin{minipage}{\linewidth}
    \scriptsize
    \textbf{Good Case}  \\
    \textbf{Model:} Gemini-3-Pro \hfill \textbf{Golden Answer:} Image A \\
    \vspace{2pt}
    
   \textbf{Model Answer:} ``[Answer]: Image A
[Reason]: \textit{Data Integrity}: Image A correctly displays the Qwen3 235B data point (cyan square) in the \enquote{Few-shot Learning} section of the top subplot, whereas Image B is missing this data point.
- \textit{Data Integrity}: Prevent data omission.'' \\
    \vspace{2pt}
    
    \textbf{Analysis:} ``Precise identification of data omission; the model successfully detects the missing specific data series, prioritizing completeness in its evaluation.''
\end{minipage}
 \vspace{0.8em}\hrule\vspace{0.8em}


\begin{minipage}{\linewidth}
    \centering
    \begin{minipage}{0.47\linewidth}
        \centering
        \includegraphics[width=\linewidth]{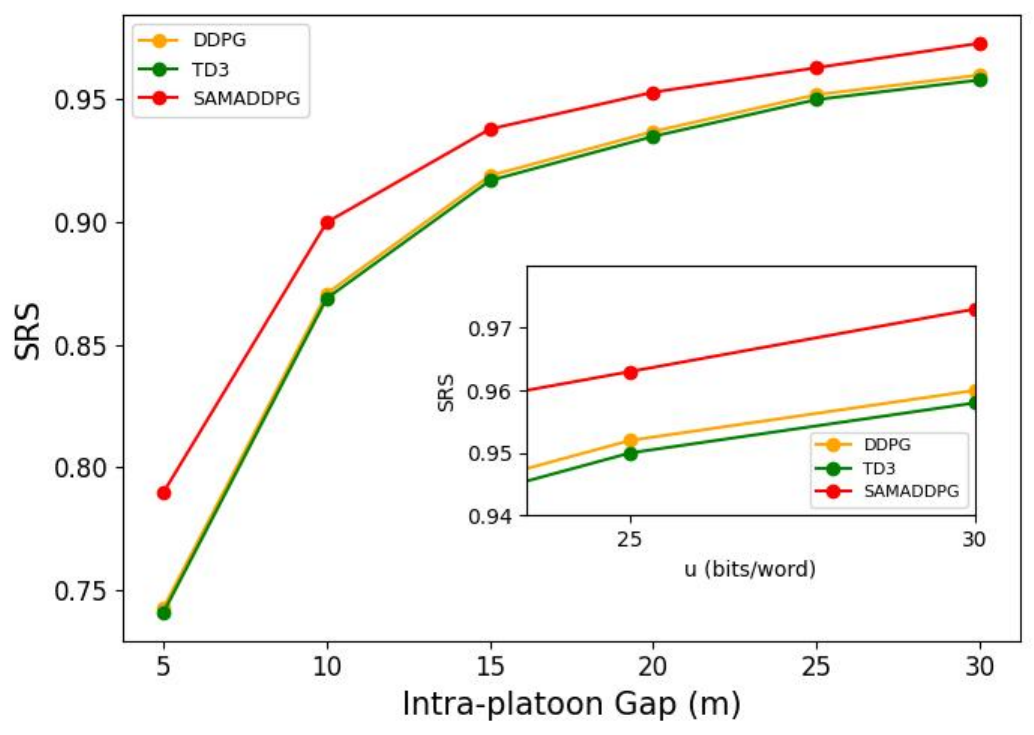}
        \\ \scriptsize \textit{Reference Figure}
    \end{minipage}
    \hspace{0.04\linewidth} 
    \begin{minipage}{0.47\linewidth}
        \centering
        \includegraphics[width=\linewidth]{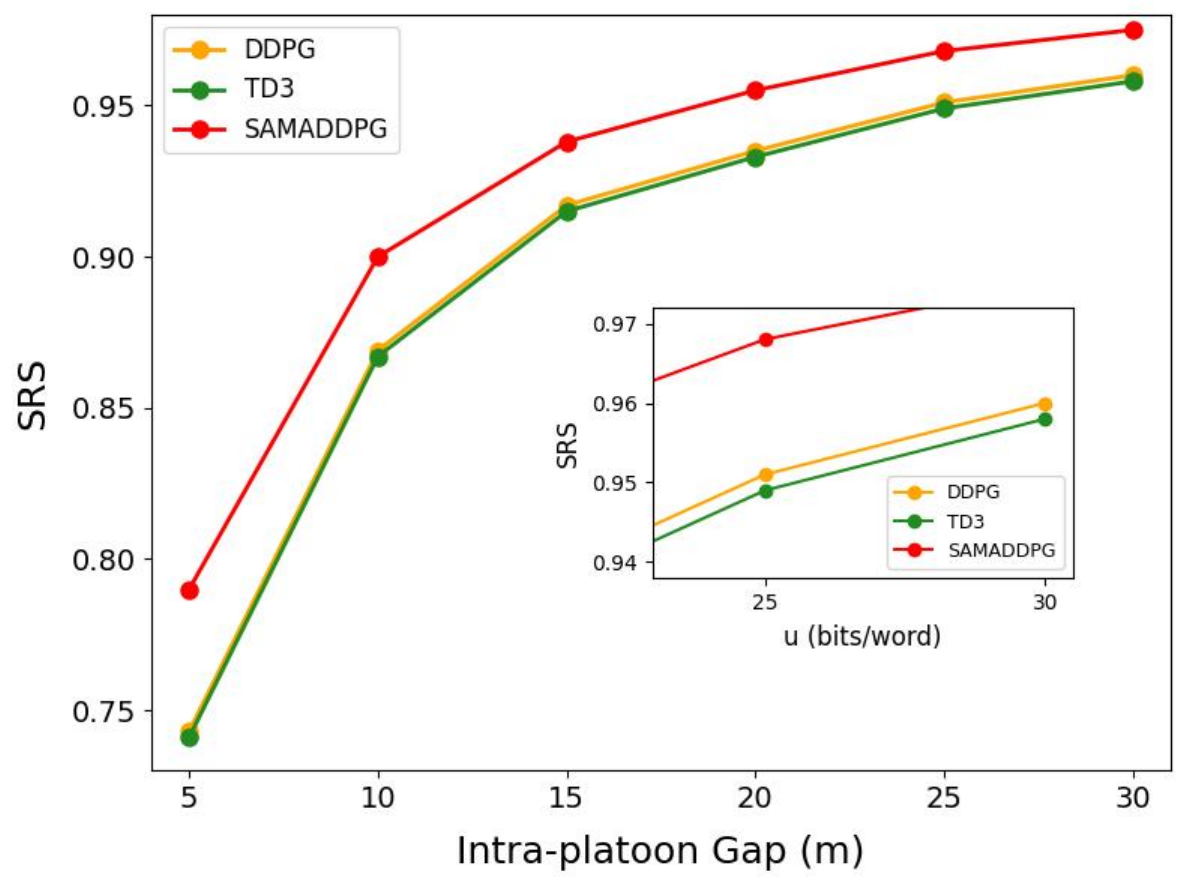}
        \\ \scriptsize \textit{Generated Figure}
    \end{minipage}
\end{minipage}

\vspace{8pt} 

\begin{minipage}{\linewidth}
    \scriptsize
    \textbf{Error Analysis: Lenience Bias or Perception}  \\
    \textbf{Model:} GPT-5.2 \hfill \textbf{Golden Answer:} Reject \\
    \vspace{2pt}
    
    \textbf{Model Answer:} ``[Answer]: Accept. [Reason]: Overall Fidelity: Data values, colors, markers, legend placement, inset, labels, and layout closely match the ground truth with no visible defects or missing elements.'' \\
    \vspace{2pt}
    
    \textbf{Analysis:} ``The generated image contains critical subplot data mismatch, failing to accurately replicate the ground truth’s data integrity.
''
\end{minipage}
 \vspace{0.8em}\hrule\vspace{0.8em}

\begin{minipage}{\linewidth}
    \centering
    \begin{minipage}{0.47\linewidth}
        \centering
        \includegraphics[width=\linewidth]{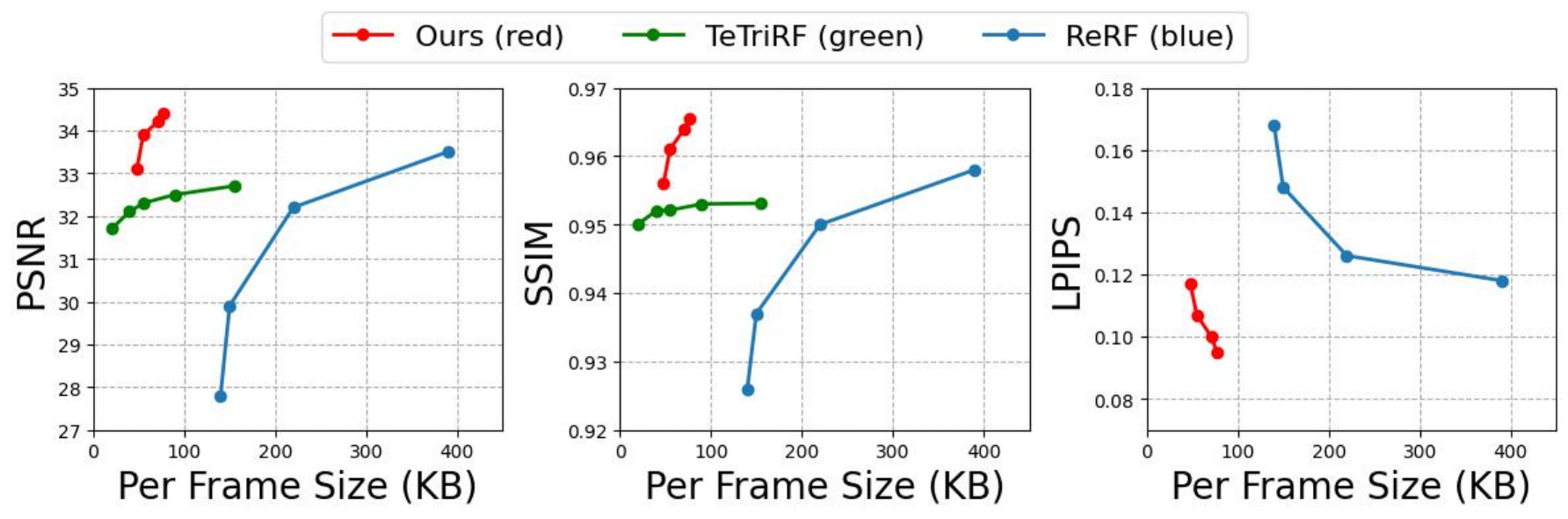}
        \\ \scriptsize \textit{Reference Figure}
    \end{minipage}
    \hspace{0.04\linewidth} 
    \begin{minipage}{0.47\linewidth}
        \centering
        \includegraphics[width=\linewidth]{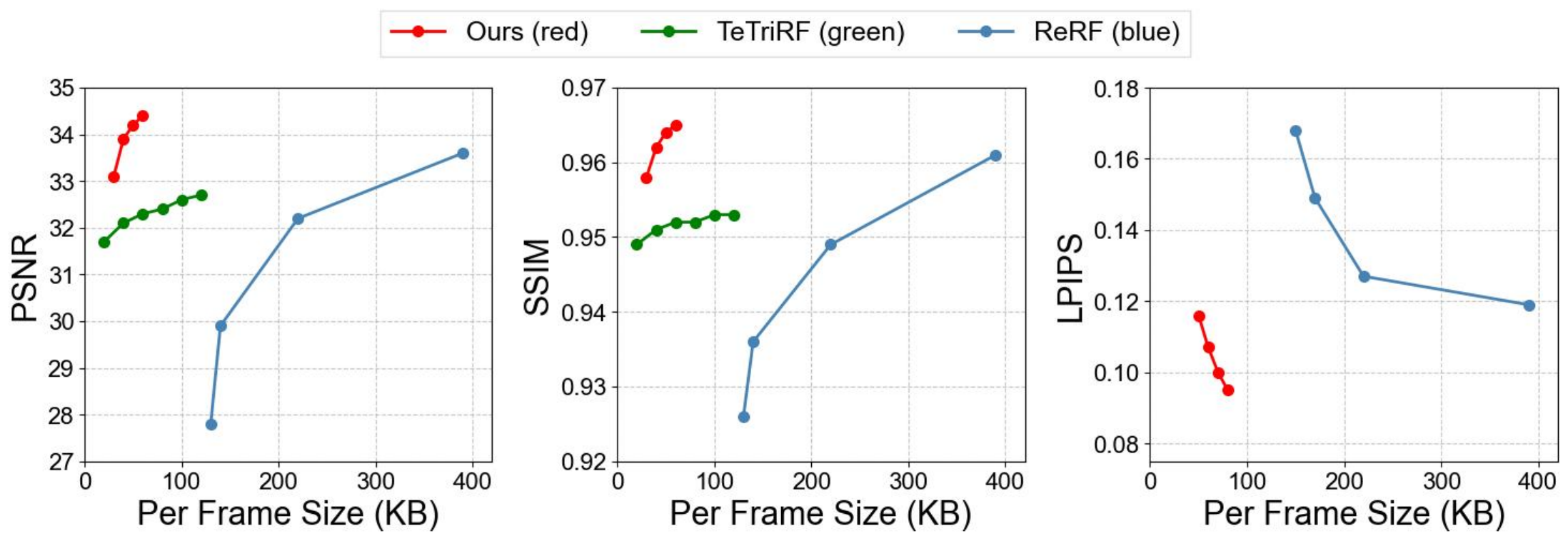}
        \\ \scriptsize \textit{Generated Figure}
    \end{minipage}
\end{minipage}

\vspace{8pt} 

\begin{minipage}{\linewidth}
    \scriptsize
    \textbf{Error Analysis: Lenience Bias or Perception}  \\
    \textbf{Model:} Seed1.6-VL \hfill \textbf{Golden Answer:} Reject \\
    \vspace{2pt}
    
    \textbf{Model Answer:} ``[Answer]: Accept. [Reason]: Data Accuracy: All values, trends, and visual elements (colors, markers, layout) match exactly. No defects observed.'' \\
    \vspace{2pt}
    
    \textbf{Analysis:} ``The generated image has an incorrect number of data points (6 instead of 5) for the TeTriRF (green) line in the PSNR subplot, failing to match the ground truth data.''
\end{minipage}
 \vspace{0.8em}\hrule\vspace{0.8em}

\begin{minipage}{\linewidth}
    \centering
    \begin{minipage}{0.47\linewidth}
        \centering
        \includegraphics[width=\linewidth]{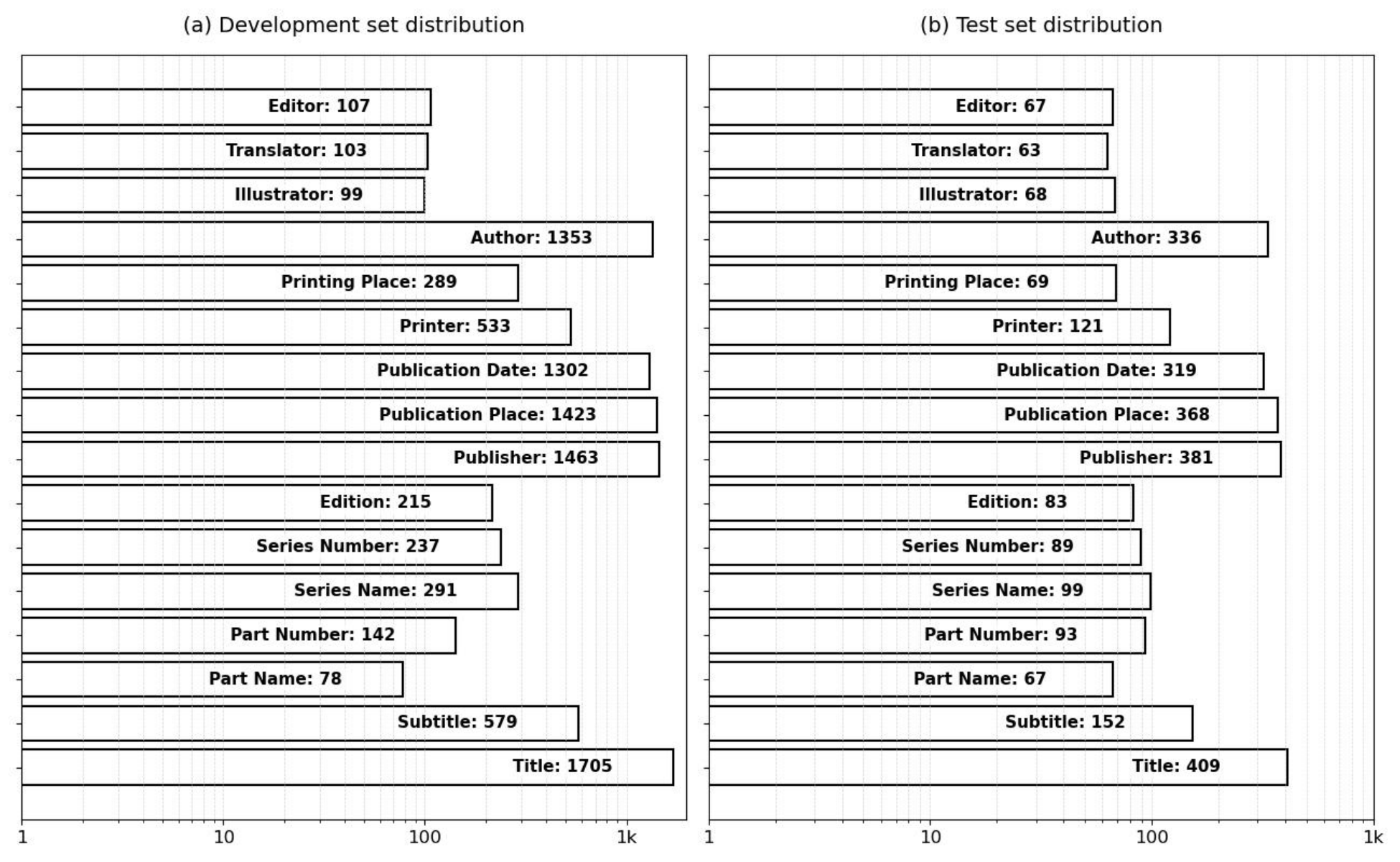}
        \\ \scriptsize \textit{Reference Figure}
    \end{minipage}
    \hspace{0.04\linewidth} 
    \begin{minipage}{0.47\linewidth}
        \centering
        \includegraphics[width=\linewidth]{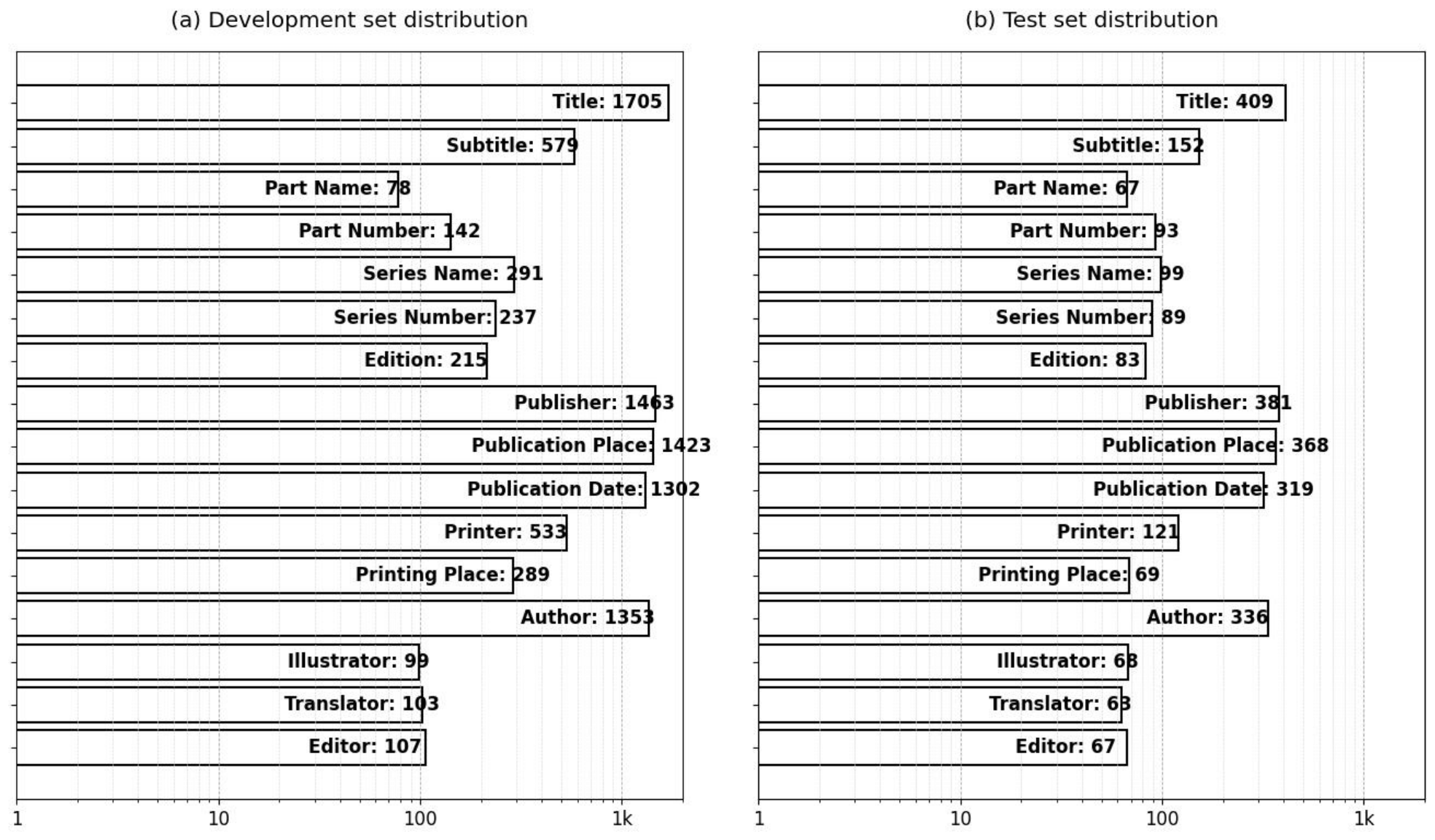}
        \\ \scriptsize \textit{Generated Figure}
    \end{minipage}
\end{minipage}

\vspace{8pt} 

\begin{minipage}{\linewidth}
    \scriptsize
    \textbf{Error Analysis: Lenience Bias or Perception}  \\
    \textbf{Model:} Qwen2.5-VL-72B \hfill \textbf{Golden Answer:} Reject \\
    \vspace{2pt}
    
    \textbf{Model Answer:} ``"[Answer]: Accept. [Reason]: Data Accuracy: Values and trends match exactly; visual elements, text, and quality are consistent with no defects."'' \\
    \vspace{2pt}
    
    \textbf{Analysis:} ``The generated image reverses the vertical label order and contains text overlap/occlusion, compromising the original visual hierarchy and readability despite preserving data values.''
\end{minipage}
 \vspace{0.8em}\hrule\vspace{0.8em}

\begin{minipage}{\linewidth}
    \centering
    \begin{minipage}{0.47\linewidth}
        \centering
        \includegraphics[width=\linewidth]{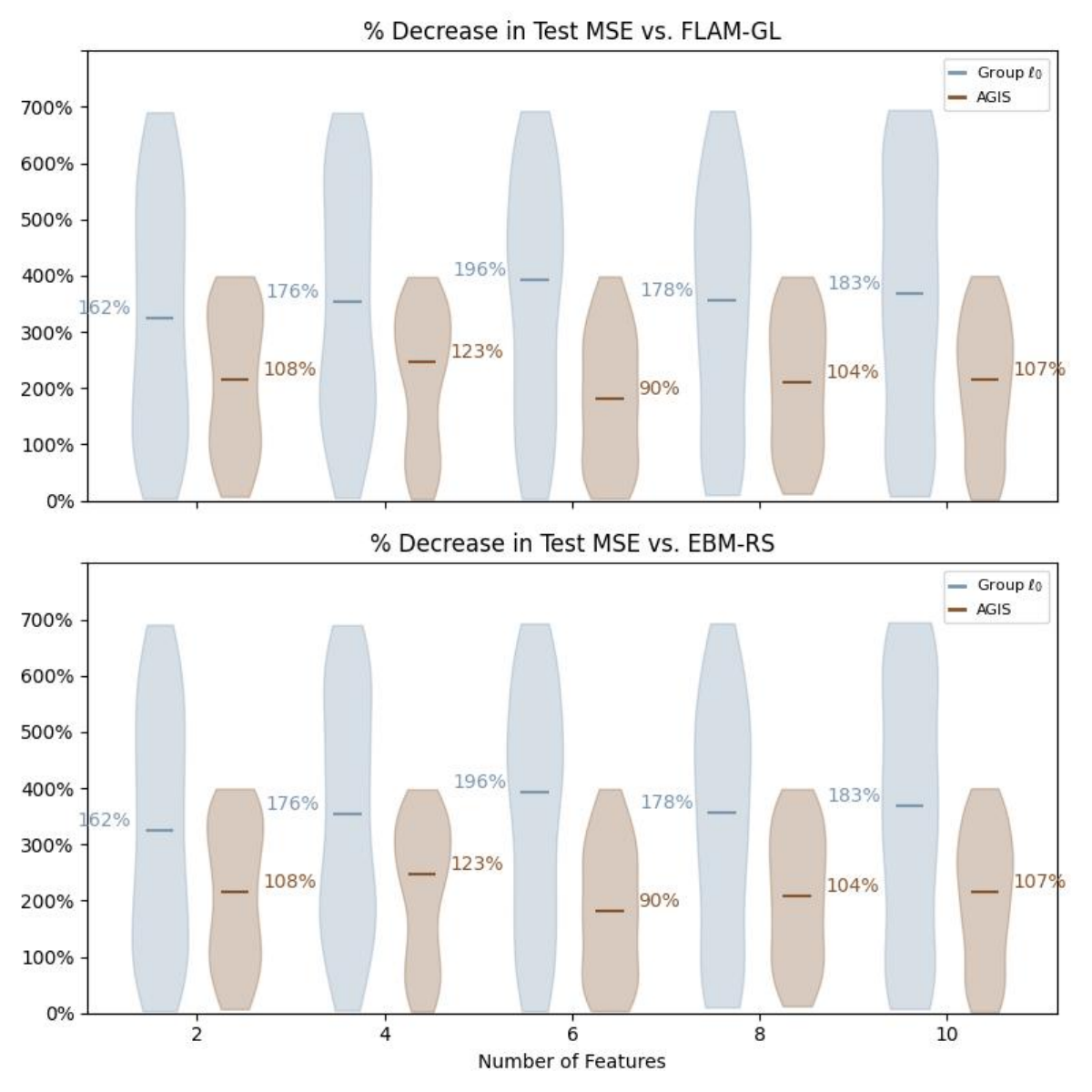}
        \\ \scriptsize \textit{Reference Figure}
    \end{minipage}
    \hspace{0.04\linewidth} 
    \begin{minipage}{0.47\linewidth}
        \centering
        \includegraphics[width=\linewidth]{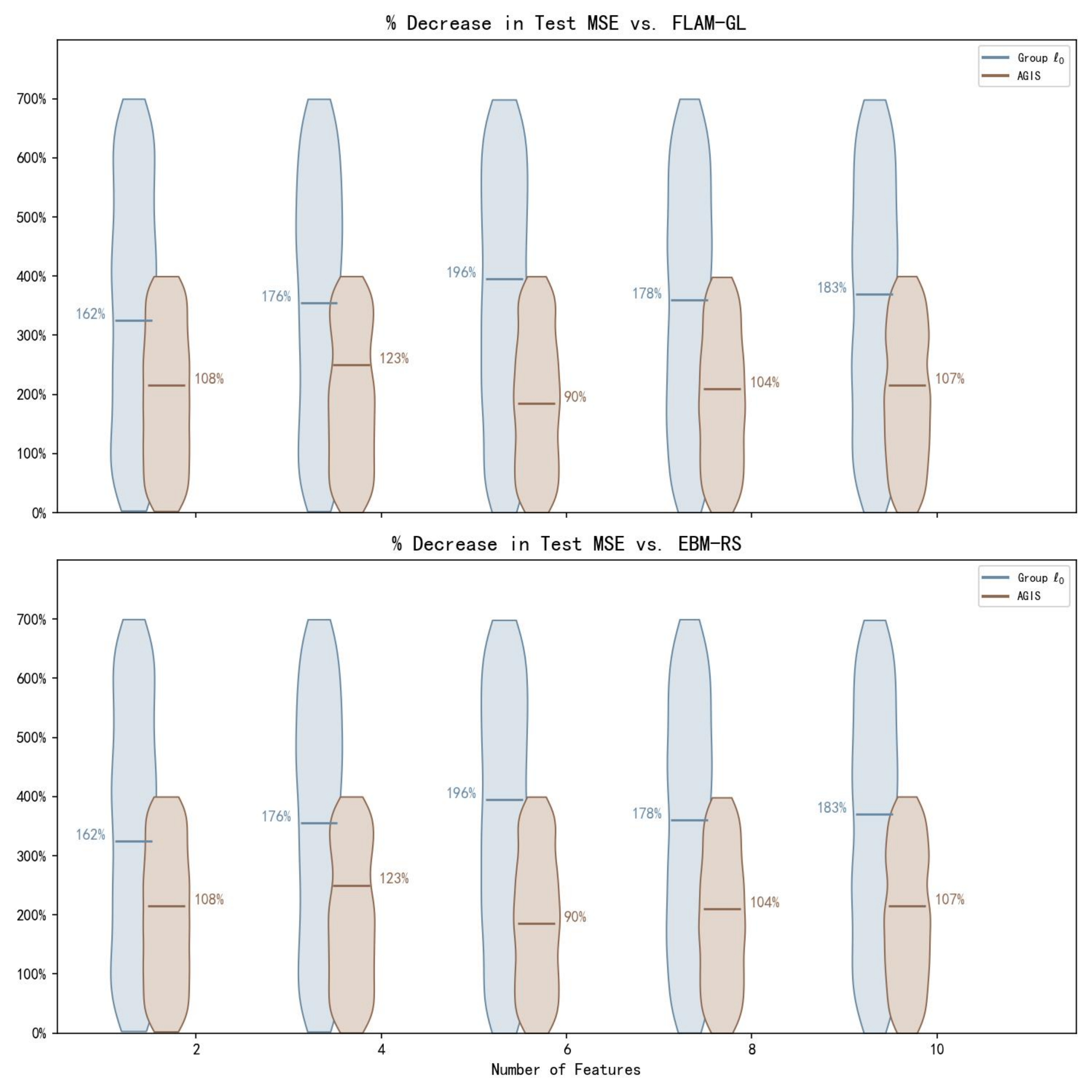}
        \\ \scriptsize \textit{Generated Figure}
    \end{minipage}
\end{minipage}

\vspace{8pt} 

\begin{minipage}{\linewidth}
    \scriptsize
    \textbf{Error Analysis: Perception}  \\
    \textbf{Model:} InternVL3.5-38B \hfill \textbf{Golden Answer:} Reject \\
    \vspace{2pt}
    
    \textbf{Model Answer:} ``[Answer]: Accept. [Reason]: Data Accuracy: All values and trends match the original. Visual Elements: Colors, markers, and line styles are consistent. Text: Titles, labels, and ticks are correct. Visual Quality: No defects like text overlapping or occlusion.'' \\
    \vspace{2pt}
    
\textbf{Analysis:} ``The generated image has severe overlap between the two violin plots (Group $\ell_0$ and AGIS), obscuring data distributions and annotations, unlike the ground truth which maintains clear separation.''
\end{minipage}
 \vspace{0.8em}\hrule\vspace{0.8em}

\begin{minipage}{\linewidth}
    \centering
    \begin{minipage}{0.47\linewidth}
        \centering
        \includegraphics[width=\linewidth]{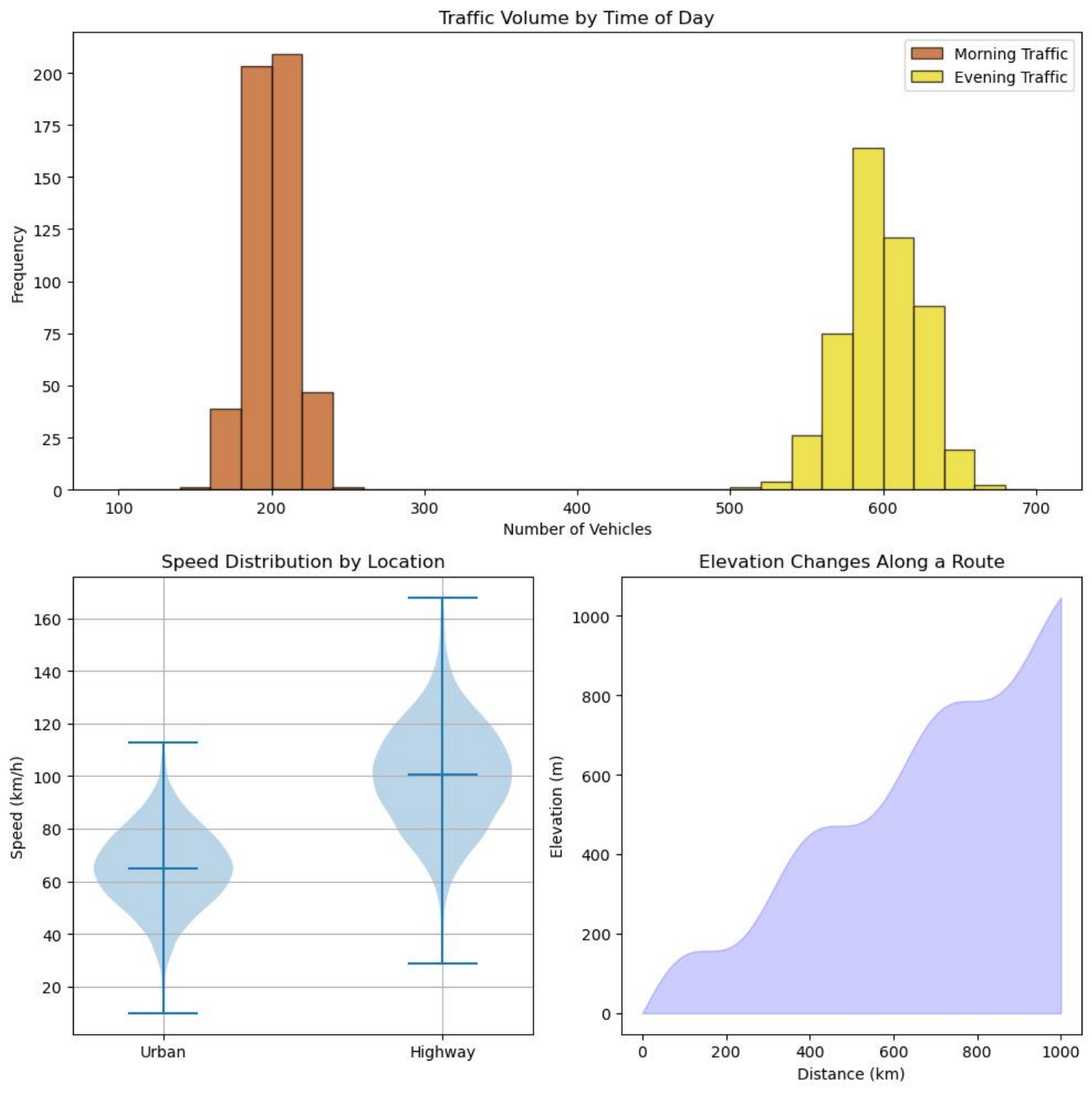}
        \\ \scriptsize \textit{Reference Figure}
    \end{minipage}
    \hspace{0.04\linewidth} 
    \begin{minipage}{0.47\linewidth}
        \centering
        \includegraphics[width=\linewidth]{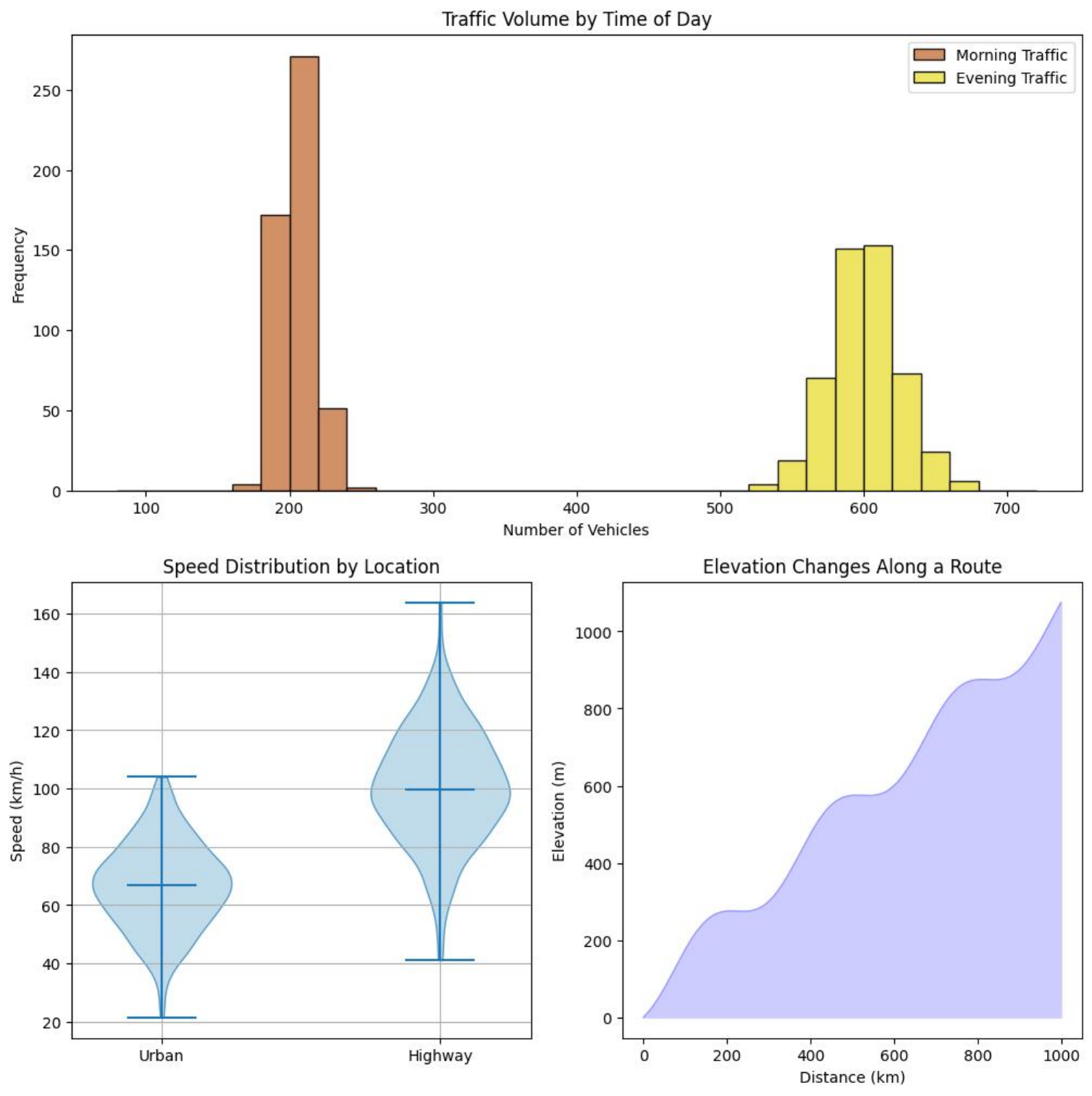}
        \\ \scriptsize \textit{Generated Figure}
    \end{minipage}
\end{minipage}

\vspace{8pt} 

\begin{minipage}{\linewidth}
    \scriptsize
    \textbf{Good Case}  \\
    \textbf{Model:} Qwen3-VL-32B \hfill \textbf{Golden Answer:} Reject \\
    \vspace{2pt}
    
    \textbf{Model Answer:} ``[Answer]: Reject. [Reason]: Data Accuracy: Morning Traffic bar heights and Evening Traffic bin frequencies differ from ground truth; visual elements and text are otherwise consistent.'' \\
    \vspace{2pt}
    
    \textbf{Analysis:} ``Precise detection of data corruption; the model correctly identifies numerical discrepancies in bar heights and frequencies despite overall visual similarity.''
\end{minipage}
 \vspace{0.8em}\hrule\vspace{0.8em}


\begin{tcolorbox}[
    colback=blue!5!white, 
    colframe=MyDarkBlue!50, 
    arc=2pt, 
    boxrule=0.5pt, 
    left=2pt, right=2pt, top=2pt, bottom=2pt
]
    \scriptsize \textbf{User Instruction:} \textit{Move the legend to the top left corner inside the chart.}
\end{tcolorbox}

\vspace{4pt}

\begin{minipage}{\linewidth}
    \centering
    \begin{minipage}{0.47\linewidth}
        \centering
        \includegraphics[width=\linewidth]{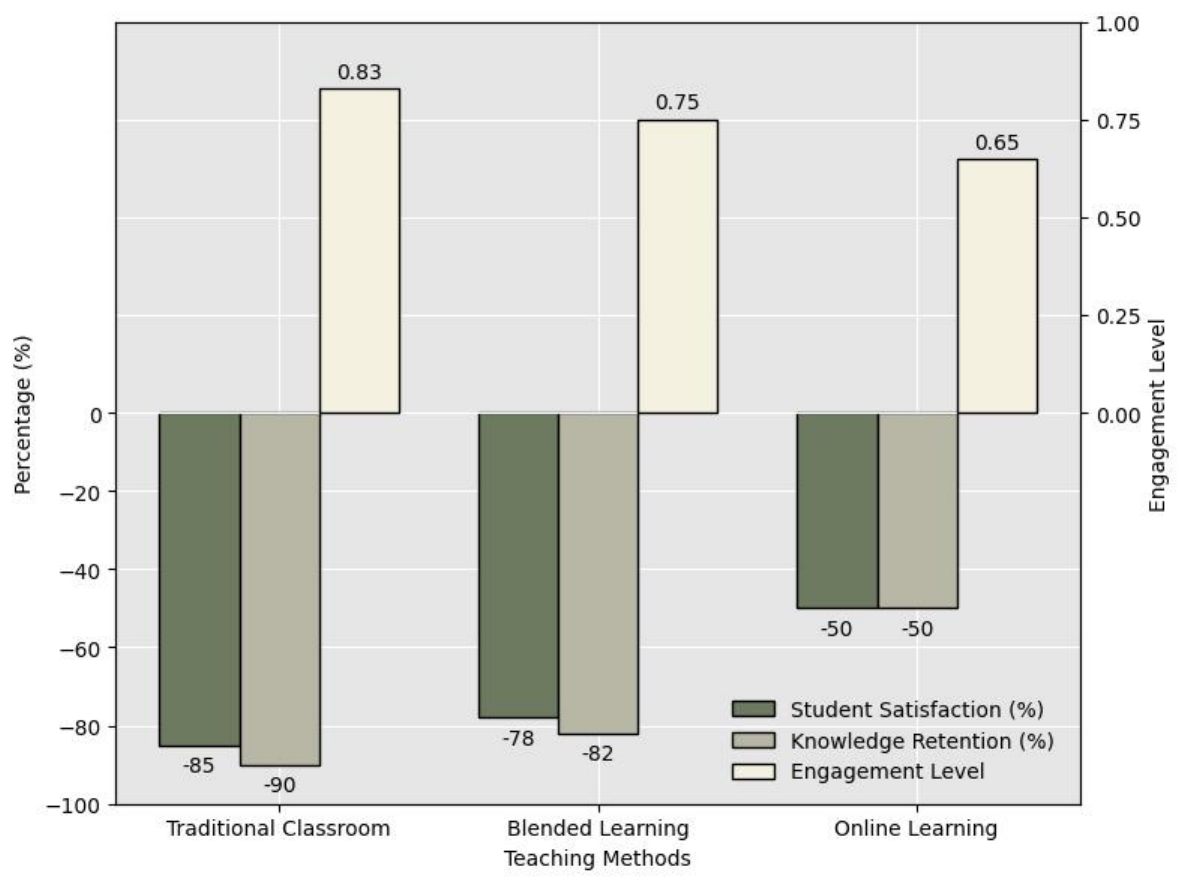}
        \\ \scriptsize \textit{Reference Figure}
    \end{minipage}
    \hspace{0.04\linewidth}
    \begin{minipage}{0.47\linewidth}
        \centering
        \includegraphics[width=\linewidth]{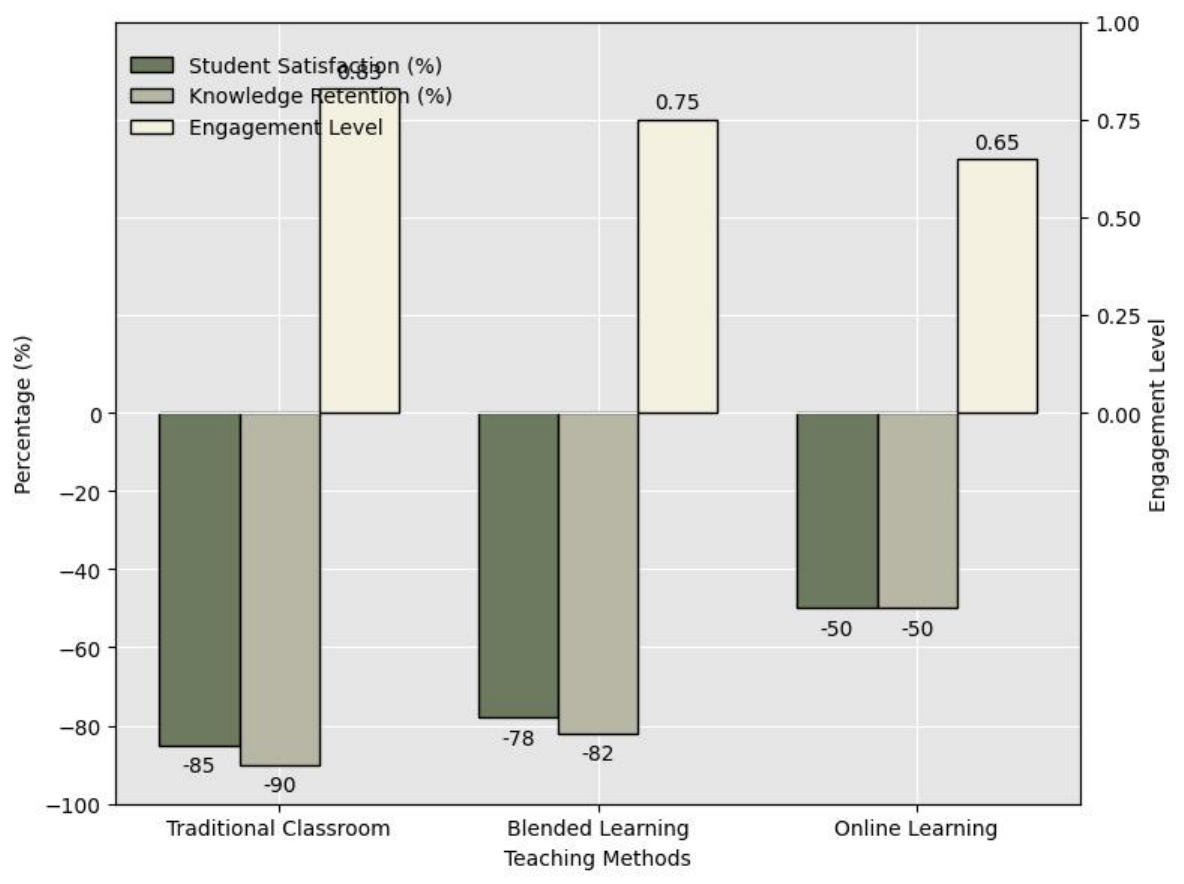}
        \\ \scriptsize \textit{Generated Figure}
    \end{minipage}
\end{minipage}

\vspace{8pt}

\begin{minipage}{\linewidth}
    \scriptsize
    \textbf{Error Analysis: Partial Verification Shortcut} \\
    \textbf{Model:} Gemini-3-Pro \hfill \textbf{Golden Answer:} Reject\\
    \vspace{2pt}
    
    \textbf{Model Answer:} ``[Answer]: Accept. [Reason]: Instruction followed: The legend was successfully relocated to the top left corner of the chart area, preserving data integrity and visual consistency.'' \\
    \vspace{2pt}
    
   \textbf{Analysis:} ``The model exhibits a confirmation bias in its evaluation; because it successfully executed the command to move the legend, it falsely concludes that the edit is successful. It fails to perform a holistic visual quality check, overlooking the fact that the new placement occludes the \enquote{Engagement Level} bar and labels. This reveals a gap between instruction following and visual constraint satisfaction.''
\end{minipage}

\vspace{0.8em}\hrule\vspace{0.8em}

\begin{tcolorbox}[
    colback=blue!5!white, 
    colframe=MyDarkBlue!50, 
    arc=2pt, 
    boxrule=0.5pt, 
    left=2pt, right=2pt, top=2pt, bottom=2pt
]
    \scriptsize \textbf{User Instruction:} \textit{Add data value labels at the center of each data point to indicate the exact values.}
\end{tcolorbox}

\vspace{4pt}

\begin{minipage}{\linewidth}
    \centering
    \begin{minipage}{0.47\linewidth}
        \centering
        \includegraphics[width=\linewidth]{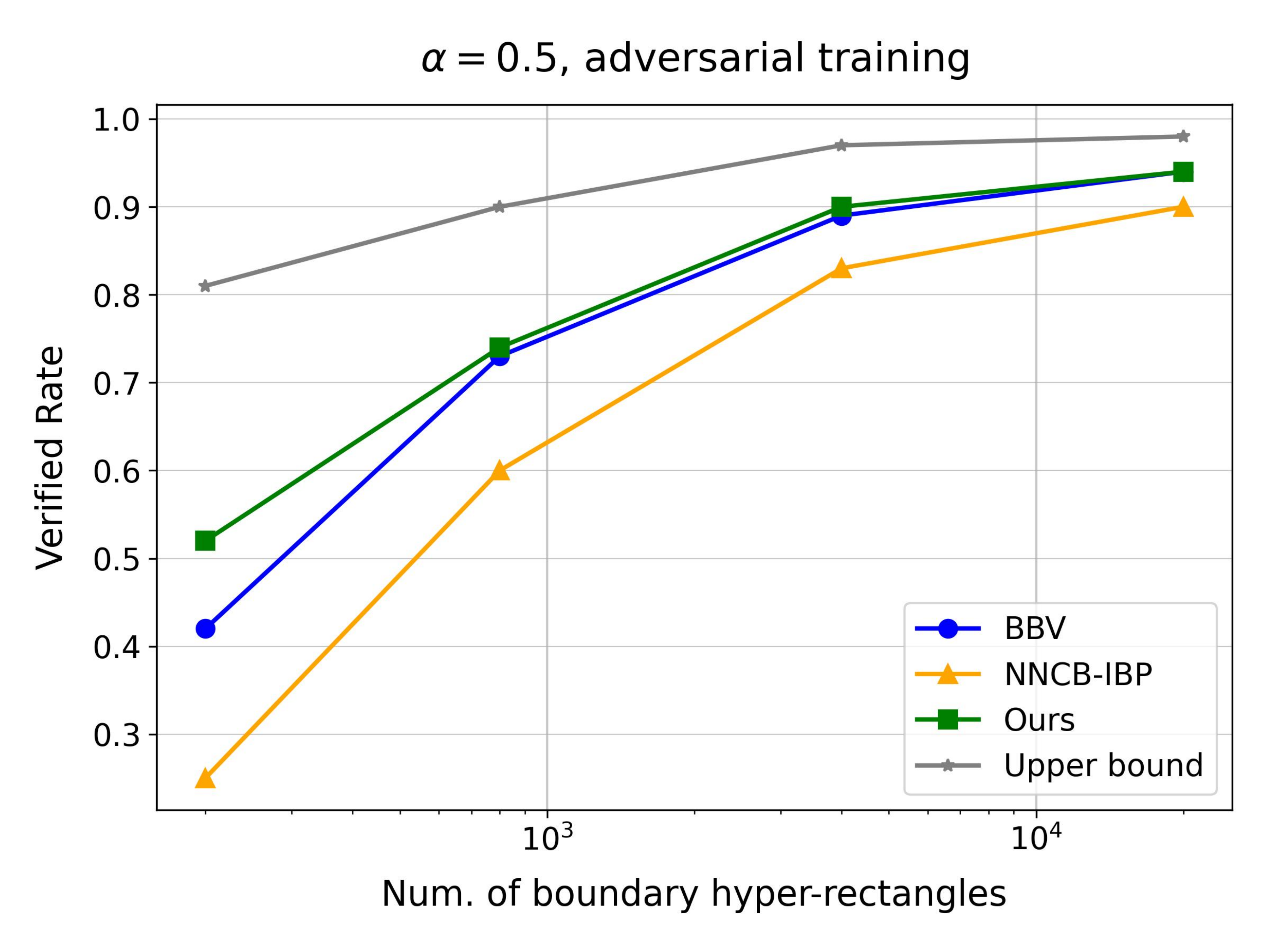}
        \\ \scriptsize \textit{Reference Figure}
    \end{minipage}
    \hspace{0.04\linewidth}
    \begin{minipage}{0.47\linewidth}
        \centering
        \includegraphics[width=\linewidth]{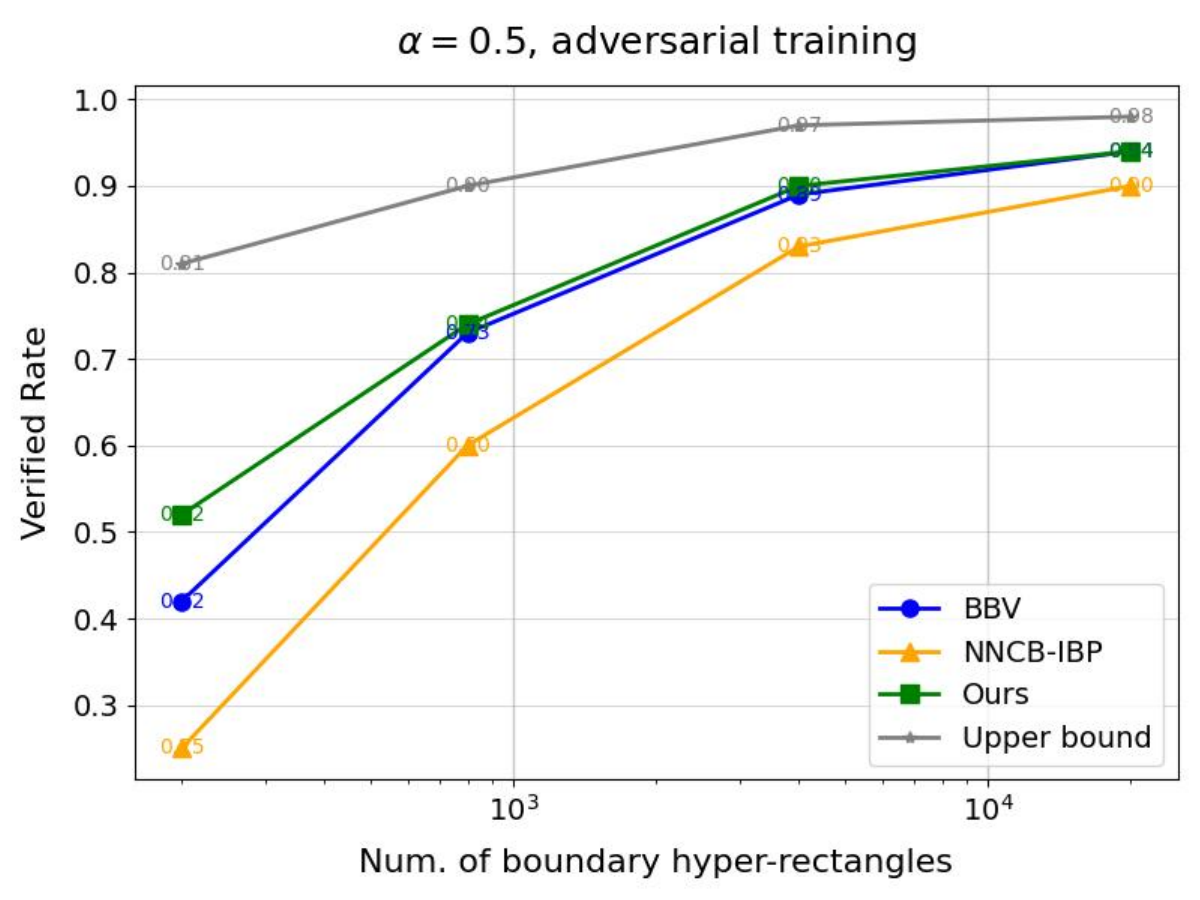}
        \\ \scriptsize \textit{Generated Figure}
    \end{minipage}
\end{minipage}

\vspace{8pt}

\begin{minipage}{\linewidth}
    \scriptsize
    \textbf{Error Analysis: Partial Verification Shortcut} \\
    \textbf{Model:} Seed1.6-VL \hfill \textbf{Golden Answer:} Reject\\
    \vspace{2pt}
    
    \textbf{Model Answer:} ``[Answer]: Accept. [Reason]: Instruction Adherence: Data value labels added at each data point center, preserving data and structure with no new defects.", "query": "Add data value labels at the center of each data point to indicate the exact values.'' \\
    \vspace{2pt}
    
    \textbf{Analysis:} ``While data labels were added as requested, they overlap with data points and lines, making exact values unreadable. This violates the Data Integrity principle and fails to fulfill the visualization goal of clear value display.''
\end{minipage}

\vspace{0.8em}\hrule\vspace{0.8em}

\begin{tcolorbox}[
    colback=blue!5!white, 
    colframe=MyDarkBlue!50, 
    arc=2pt, 
    boxrule=0.5pt, 
    left=2pt, right=2pt, top=2pt, bottom=2pt
]
    \scriptsize \textbf{User Instruction:} \textit{Add a percentage sign to the end of all the numeric labels on top of the bars and make them bold.}
\end{tcolorbox}

\vspace{4pt}

\begin{minipage}{\linewidth}
    \centering
    \begin{minipage}{0.47\linewidth}
        \centering
        \includegraphics[width=\linewidth]{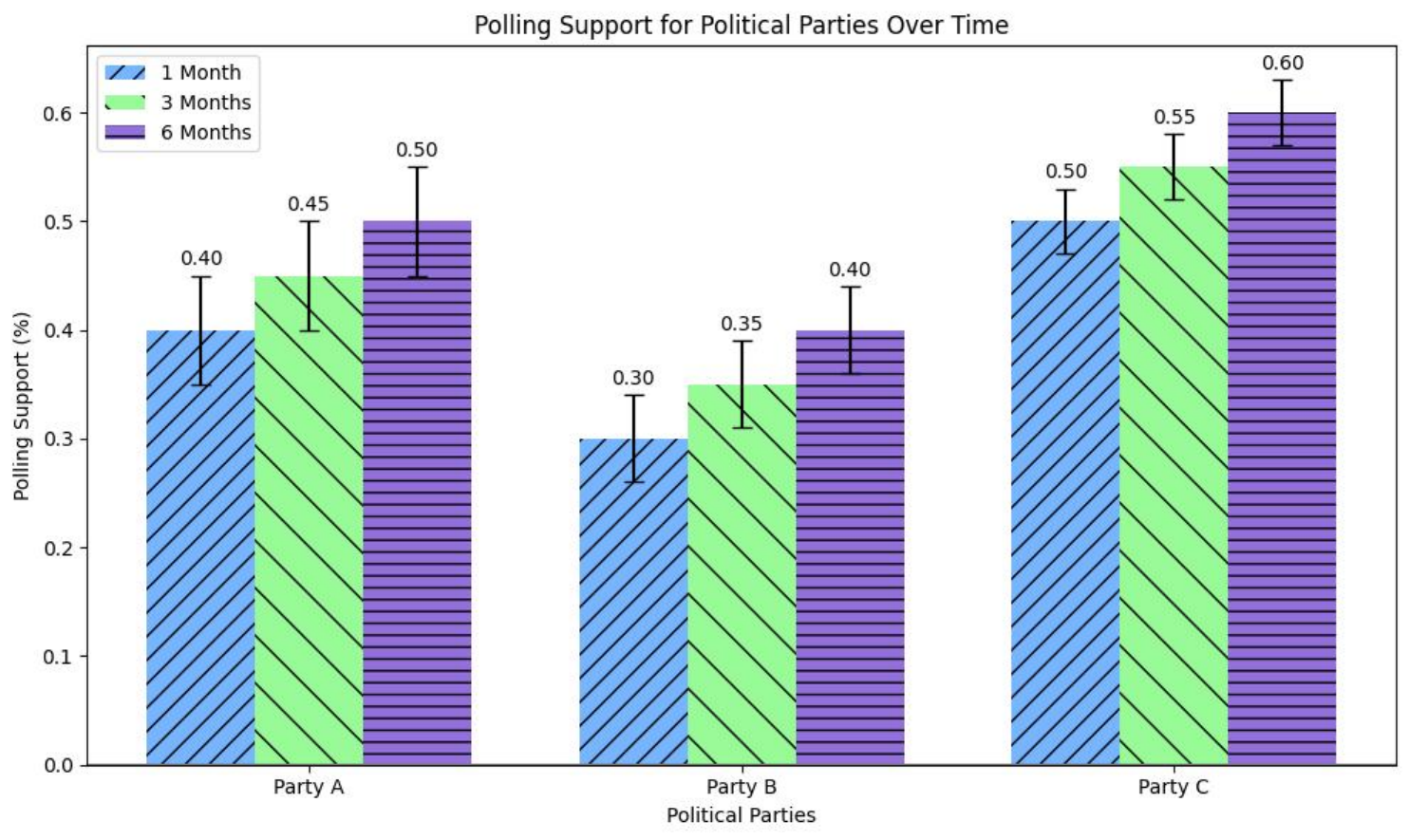}
        \\ \scriptsize \textit{Reference Figure}
    \end{minipage}
    \hspace{0.04\linewidth}
    \begin{minipage}{0.47\linewidth}
        \centering
        \includegraphics[width=\linewidth]{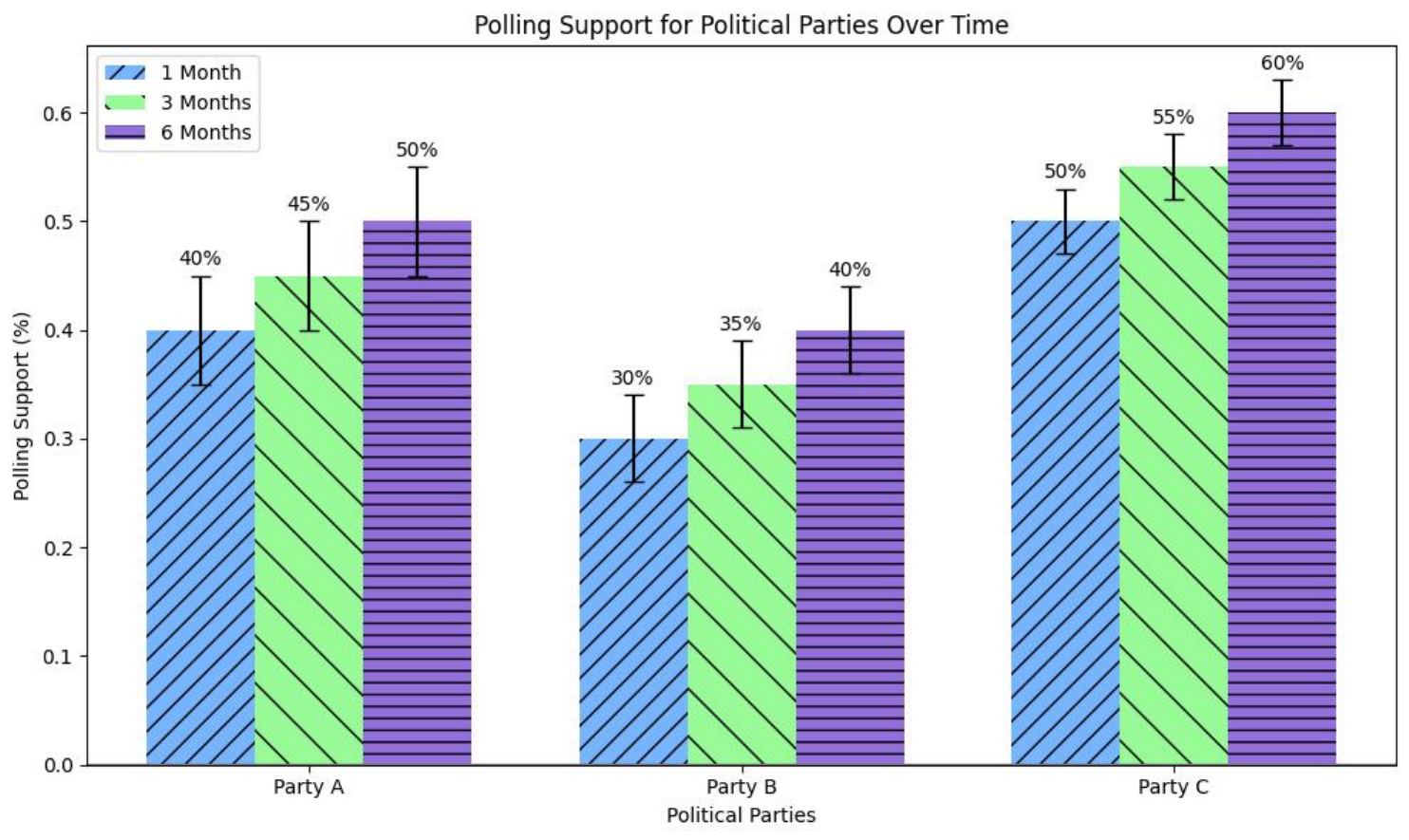}
        \\ \scriptsize \textit{Generated Figure}
    \end{minipage}
\end{minipage}

\vspace{8pt}

\begin{minipage}{\linewidth}
    \scriptsize
    \textbf{Error Analysis: Perception} \\
    \textbf{Model:} Qwen3-VL-32B \hfill \textbf{Golden Answer:} Reject\\
    \vspace{2pt}
    
    \textbf{Model Answer:} "[Answer]: Accept. [Reason]: Instruction Adherence: All numeric labels now end with '\%' and are bolded as requested. Data Integrity: Values unchanged, bars and structure preserved. Visual Quality Check: No new defects. Consistency with Original: All other elements unchanged.'' \\
    \vspace{2pt}
    
    \textbf{Analysis:} ``The model failed to recognize and apply the bold formatting requirement, resulting in incomplete adherence to the user instruction. While other criteria are met, the failure to follow the explicit styling instruction warrants rejection.''
\end{minipage}

\vspace{0.8em}\hrule\vspace{0.8em}

\begin{tcolorbox}[
    colback=blue!5!white, 
    colframe=MyDarkBlue!50, 
    arc=2pt, 
    boxrule=0.5pt, 
    left=2pt, right=2pt, top=2pt, bottom=2pt
]
    \scriptsize \textbf{User Instruction:} \textit{Make the median lines black instead of blue and give the boxes some transparency so we can see the grid lines behind them.}
\end{tcolorbox}

\vspace{4pt}

\begin{minipage}{\linewidth}
    \centering
    \begin{minipage}{0.47\linewidth}
        \centering
        \includegraphics[width=\linewidth]{appendix/0-Figure/error_cases/box_96_fail_ref.pdf}
        \\ \scriptsize \textit{Reference Figure}
    \end{minipage}
    \hspace{0.04\linewidth}
    \begin{minipage}{0.47\linewidth}
        \centering
        \includegraphics[width=\linewidth]{appendix/0-Figure/error_cases/box_96_fail_edit.pdf}
        \\ \scriptsize \textit{Generated Figure}
    \end{minipage}
\end{minipage}

\vspace{8pt}

\begin{minipage}{\linewidth}
    \scriptsize
    \textbf{Error Analysis: Perception} \\
    \textbf{Model:} InternVL3.5-38B \hfill \textbf{Golden Answer:} Reject\\
    \vspace{2pt}
    
    \textbf{Model Answer:} ``[Answer]: Accept. [Reason]: Instruction Adherence: The median lines are now black, and the boxes have transparency, allowing the grid lines to be visible. Data Integrity: The data values and chart structures remain unchanged. Visual Quality Check: No new visual defects are introduced. Consistency with Original: All non-edited parts are consistent with the original image.'' \\
    \vspace{2pt}
    
\textbf{Analysis:} ``The model fails to detect the fine-grained discrepancy between the editing instruction (which specifies `black') and the actual rendered result (which remains `red'). This perception error stems from the subtle color difference, leading the model to incorrectly validate the output as compliant with the instruction.''
\end{minipage}

\vspace{0.8em}\hrule\vspace{0.8em}

\begin{tcolorbox}[
    colback=blue!5!white, 
    colframe=MyDarkBlue!50, 
    arc=2pt, 
    boxrule=0.5pt, 
    left=2pt, right=2pt, top=2pt, bottom=2pt
]
    \scriptsize \textbf{User Instruction:} \textit{Place the title at the bottom of the chart, below the axis labels.}
\end{tcolorbox}

\vspace{4pt}

\begin{minipage}{\linewidth}
    \centering
    \begin{minipage}{0.47\linewidth}
        \centering
        \includegraphics[width=\linewidth]{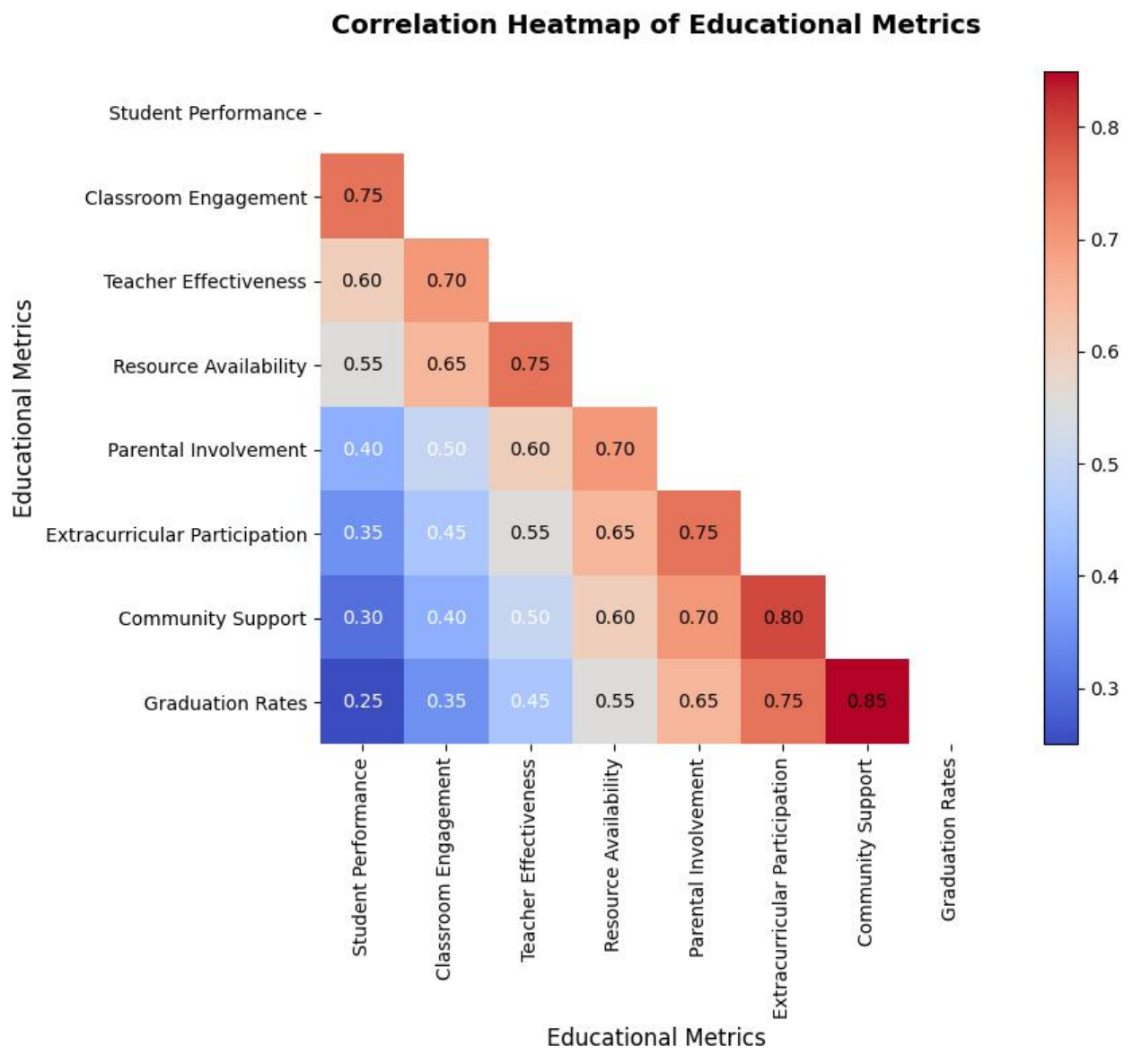}
        \\ \scriptsize \textit{Reference Figure}
    \end{minipage}
    \hspace{0.04\linewidth}
    \begin{minipage}{0.47\linewidth}
        \centering
        \includegraphics[width=\linewidth]{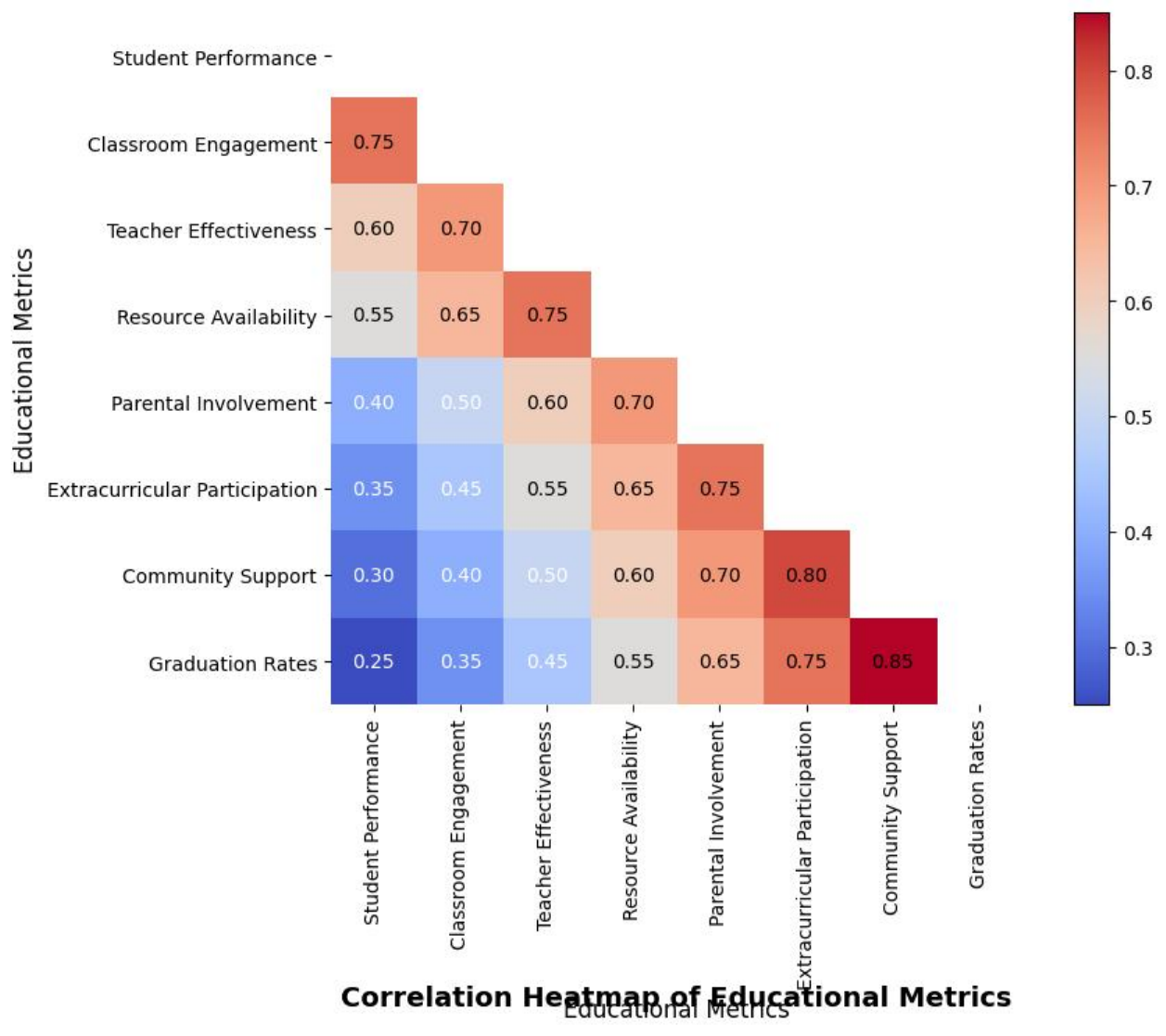}
        \\ \scriptsize \textit{Generated Figure}
    \end{minipage}
\end{minipage}

\vspace{8pt}

\begin{minipage}{\linewidth}
    \scriptsize
    \textbf{Good Case} \\
    \textbf{Model:} GPT-5.2 \hfill \textbf{Golden Answer:} Reject\\
    \vspace{2pt}
    
    \textbf{Model Answer:} ``[Answer]: Reject. [Reason]: Visual Defect: The title was moved to the bottom but overlaps with the x-axis label and appears clipped, introducing a new visual quality issue not present in the original.'' \\
    \vspace{2pt}
    
   \textbf{Analysis:} ``The model demonstrates spatial reasoning and  layout validation capabilities. Instead of simply confirming the execution of the instruction (moving the title), it performs a holistic check and identifies a \enquote{Side Effect}: the new position causes a collision with the x-axis label. This indicates the model has moved beyond simple instruction following to a higher-level understanding of visual quality constraints.''
\end{minipage}

\vspace{0.8em}\hrule\vspace{0.8em}
 
\end{tcolorbox}

\subsection{Dataset Gallery}
\begin{figure}[htbp]
    \centering
    \includegraphics[width=0.76\textwidth]{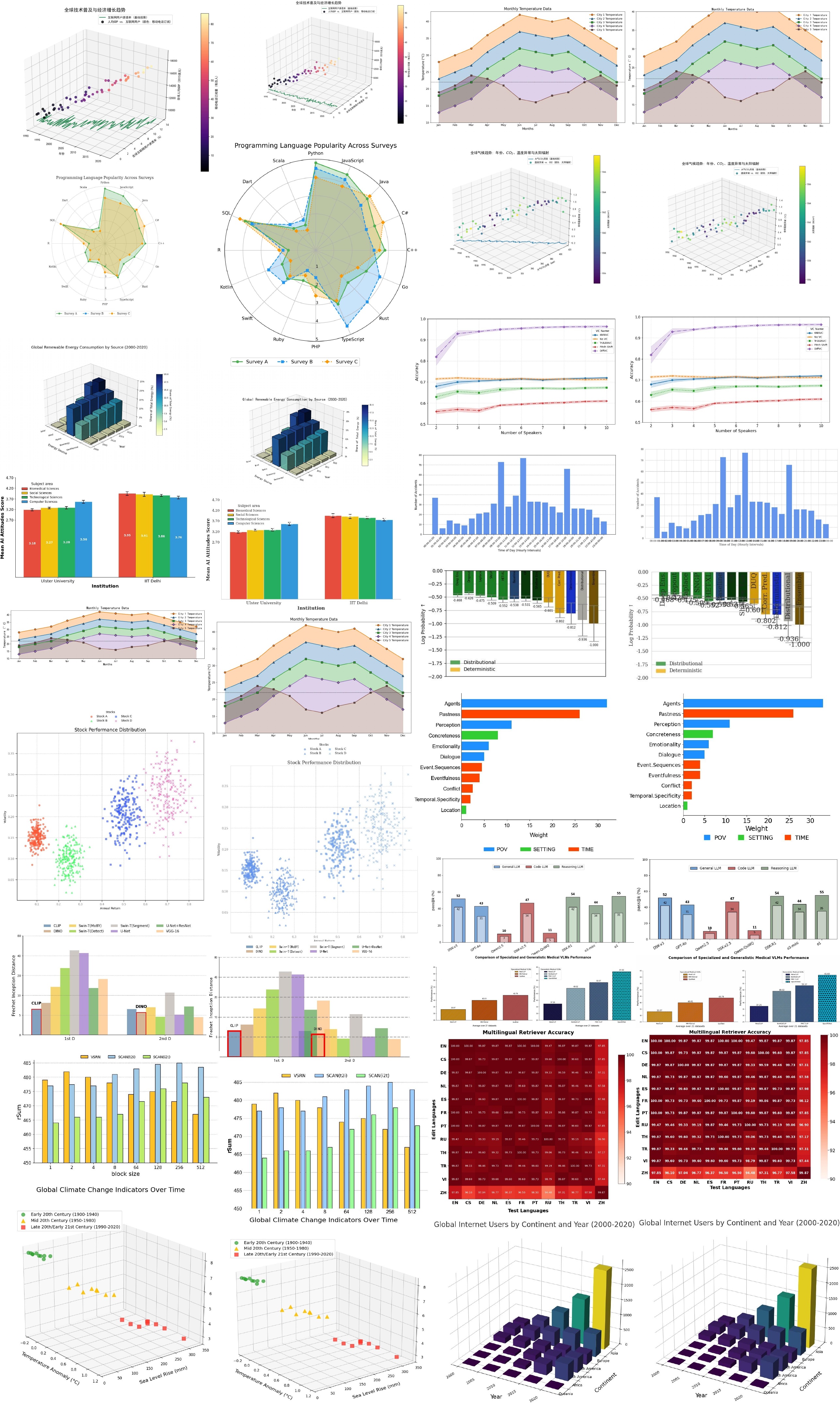}
    \caption{Selected samples from the \benchmark dataset, illustrating various chart types and their corresponding flaw injections.}
    \label{fig:dataset_gallery_overview}
\end{figure}

Figure~\ref{fig:dataset_gallery_overview} visualizes a sampled subset of \benchmark, demonstrating its chart diversity and fine-grained flaw injections across multiple dimensions.

\clearpage

\section{Experimental Setup and Inference Statistics}
\label{app:exp_setup}

\subsection{Model Sources and Hyperparameters}
To ensure the reproducibility of our evaluation, we meticulously document the sources and configurations of all evaluated large multimodal models (LMMs). As detailed in Table~\ref{tab:model_sources}, the models are accessed either via official API endpoints or downloaded directly from HuggingFace. 

Regarding inference configurations (Table~\ref{tab:model_hyperparameters}), we utilize standard deterministic decoding parameters, setting the temperature to 0.1 and top-p to 0.9 as the default baseline for all task evaluations. However, for models evaluated using Test-Time computation (e.g., Majority Voting), we intentionally increase the temperature (T=0.6) and top-p (P=0.95) to encourage response diversity and broader exploration of the reasoning space.

\begin{supplementtable}
    \centering
    \caption{\textbf{Model Versions and Hyperparameter Configurations.} Default settings use Temp=0.1 and Top-P=0.9, while Test-Time setups use higher values to encourage diversity.}
    \label{tab:model_hyperparameters}
    \renewcommand{\arraystretch}{1.15}
    \setlength{\tabcolsep}{4pt}
    \resizebox{\textwidth}{!}{
    \begin{tabular}{l l c c c c}
        \toprule
        \textbf{Model} & \textbf{Version / HF Checkpoint} & \textbf{Max Tokens} & \textbf{Temp.} & \textbf{Top-P} & \textbf{Test-Time Setup} \\
        \midrule
        \multicolumn{6}{c}{\textbf{Proprietary Multimodal Large Language Models}} \\
        \midrule
        GPT-5.2 & gpt-5.2-2025-12-11 & 512 & 0.1 & 0.9 & T=0.6, P=0.95 \\
        Gemini-3-Pro & gemini-3-pro-20251118 & 512 & 0.1 & 0.9 & T=0.6, P=0.95 \\
        Claude-Sonnet-4.5 & claude-4.5-sonnet-20250929 & 512 & 0.1 & 0.9 & - \\
        Seed1.6-VL & seed1.6-VL-20250625 & 512 & 0.1 & 0.9 & T=0.6, P=0.95 \\
        \midrule
        \multicolumn{6}{c}{\textbf{Open-Source Models (Thinking / RL-aligned)}} \\
        \midrule
        MiMo-VL-SFT &  XiaomiMiMo/MiMo-VL-7B-SFT & 1024 & 0.1 & 0.9 & - \\
        MiMo-VL-RL &  XiaomiMiMo/MiMo-VL-7B-RL & 1024 & 0.1 & 0.9 & - \\
        Qwen3-VL-30B & Qwen/Qwen3-VL-30B-A3B-Think & 1024 & 0.1 & 0.9 & - \\
        LLaVA-Critic-R1 & lmms-lab/llava-critic-r1 & 1024 & 0.1 & 0.9 & - \\
        ThinkLite-VL-7B & russwang/ThinkLite-VL & 1024 & 0.1 & 0.9 & - \\
        \midrule
        \multicolumn{6}{c}{\textbf{Open-Source Models (Non-thinking)}} \\
        \midrule
        Qwen2.5-VL-72B & Qwen/Qwen2.5-VL-72B-Instruct & 512 & 0.1 & 0.9 & T=0.6, P=0.95 \\
        Qwen2.5-VL-7B & Qwen/Qwen2.5-VL-7B-Instruct & 512 & 0.1 & 0.9 & - \\
        Qwen2-VL-72B & Qwen/Qwen2-VL-72B-Instruct & 512 & 0.1 & 0.9 & - \\
        Qwen2-VL-7B & Qwen/Qwen2-VL-7B-Instruct & 512 & 0.1 & 0.9 & - \\
        InternVL3.5-38B & OpenGVLab/InternVL3\_5-38B & 1024 & 0.1 & 0.9 & T=0.6, P=0.95 \\
        InternVL3.5-8B & OpenGVLab/InternVL3\_5-8B & 512 & 0.1 & 0.9 & - \\
        InternVL2.5-38B & OpenGVLab/InternVL2\_5-38B & 512 & 0.1 & 0.9 & - \\
        InternVL2.5-8B & OpenGVLab/InternVL2\_5-8B & 512 & 0.1 & 0.9 & - \\
        Qwen3-VL-32B & Qwen/Qwen3-VL-32B-Instruct & 512 & 0.1 & 0.9 & T=0.6, P=0.95 \\
        Qwen3-VL-8B & Qwen/Qwen3-VL-8B-Instruct & 512 & 0.1 & 0.9 & - \\
        DeepSeek-VL & deepseek-ai/deepseek-vl-7b-chat & 512 & 0.1 & 0.9 & - \\
        GLM-4V & zai-org/glm-4.6v-Flash & 512 & 0.1 & 0.9 & - \\
        Kimi-VL & moonshotai/Kimi-VL & 512 & 0.1 & 0.9 & - \\
        Molmo & allenai/Molmo-7B-D-0924 & 512 & 0.1 & 0.9 & - \\
        Molmo2 & allenai/Molmo2-7B & 512 & 0.1 & 0.9 & - \\
        MiMo-VL-SFT(NoThink) &  XiaomiMiMo/MiMo-VL-7B-SFT & 512 & 0.1 & 0.9 & - \\
        MiMo-VL-RL(NoThink) &  XiaomiMiMo/MiMo-VL-7B-RL & 512 & 0.1 & 0.9 & - \\
        \bottomrule
    \end{tabular}
    }
\end{supplementtable}

\begin{supplementtable}
    \centering
    \caption{\textbf{Release Time and Model Sources.} A detailed list of the proprietary and open-source LMMs evaluated in our benchmark.}
    \label{tab:model_sources}
    \renewcommand{\arraystretch}{1.15}
    \setlength{\tabcolsep}{6pt}
    \resizebox{\textwidth}{!}{
    \begin{tabular}{l c l}
        \toprule
        \textbf{Model} & \textbf{Release Time} & \textbf{Source / URL} \\
        \midrule
        \multicolumn{3}{c}{\textbf{Proprietary Models}} \\
        \midrule
        Gemini-3-Pro & 2025-11-18 & \url{https://deepmind.google/models/gemini/pro/} \\
        GPT-5.2 & 2025-12-11 & \url{https://openai.com/index/introducing-gpt-5-2/} \\
        Claude-Sonnet-4.5 & 2025-09-29 & \url{https://www.anthropic.com/news/claude-sonnet-4-5} \\
        Seed1.6-VL & 2025-06-25 & \url{https://seed.bytedance.com/en/seed1_6} \\
        \midrule
        \multicolumn{3}{c}{\textbf{Open-Source Models}} \\
        \midrule
        MiMo-VL-SFT & 2025-08-10 & \url{https://huggingface.co/XiaomiMiMo/MiMo-VL-7B-SFT-2508} \\
        MiMo-VL-RL & 2025-08-10 & \url{https://huggingface.co/XiaomiMiMo/MiMo-VL-7B-RL-2508} \\
        LLaVA-Critic-R1 & 2025-07-19 & \url{https://huggingface.co/lmms-lab/LLaVA-Critic-R1-7B} \\
        ThinkLite-VL-7B & 2025-04-18 & \url{https://huggingface.co/russwang/ThinkLite-VL-7B} \\
        Qwen3-VL-30B & 2025-10-03 & \url{https://huggingface.co/Qwen/Qwen3-VL-30B-A3B-Thinking} \\
        Qwen3-VL-32B & 2025-10-23 & \url{https://huggingface.co/Qwen/Qwen3-VL-32B-Instruct} \\
        Qwen3-VL-8B & 2025-10-23 & \url{https://huggingface.co/Qwen/Qwen3-VL-8B-Instruct} \\
        Qwen2.5-VL-72B & 2025-01-26 & \url{https://huggingface.co/Qwen/Qwen2.5-VL-72B-Instruct} \\
        Qwen2.5-VL-7B & 2025-01-26 & \url{https://huggingface.co/Qwen/Qwen2.5-VL-7B-Instruct} \\
        Qwen2-VL-72B & 2024-09-18 & \url{https://huggingface.co/Qwen/Qwen2-VL-72B-Instruct} \\
        Qwen2-VL-7B & 2024-09-18 & \url{https://huggingface.co/Qwen/Qwen2-VL-7B-Instruct} \\
        InternVL3.5-38B & 2025-08-25 & \url{https://huggingface.co/OpenGVLab/InternVL3_5-38B} \\
        InternVL3.5-8B & 2025-08-25 & \url{https://huggingface.co/OpenGVLab/InternVL3_5-8B} \\
        InternVL2.5-38B & 2024-11-21 & \url{https://huggingface.co/OpenGVLab/InternVL2_5-38B} \\
        InternVL2.5-8B & 2024-11-21 & \url{https://huggingface.co/OpenGVLab/InternVL2_5-8B} \\
        DeepSeek-VL & 2024-03-09 & \url{https://huggingface.co/deepseek-ai/deepseek-vl-7b-base} \\
        GLM-4V & 2025-12-08 & \url{https://huggingface.co/zai-org/GLM-4.6V-Flash} \\
        Kimi-VL & 2024-08-20 & \url{https://huggingface.co/moonshotai/Kimi-VL-A3B-Thinking} \\
        Molmo & 2024-09-25 & \url{https://huggingface.co/allenai/Molmo-7B-D-0924} \\
        Molmo2 & 2025-12-16 & \url{https://huggingface.co/allenai/Molmo2-8B} \\
        \bottomrule
    \end{tabular}
    }
\end{supplementtable}

\subsection{Inference Length and Truncation Analysis}
\label{app:inference_length}
\begin{supplementtable}
    \centering
    \caption{\textbf{Inference Output Length and Parse Error Statistics.} The average output length is measured in characters. Parse errors denote failures in following strict format instructions.}
    \label{tab:inference_stats}
    \renewcommand{\arraystretch}{1.15}
    \setlength{\tabcolsep}{3.5pt} 
    \resizebox{\textwidth}{!}{
    \begin{tabular}{l | c c | c c | c c}
        \toprule
        \multirow{2}{*}{\textbf{Model}} & \multicolumn{2}{c|}{\textbf{Task 1 (CPA)}} & \multicolumn{2}{c|}{\textbf{Task 2 (Reproduction)}} & \multicolumn{2}{c}{\textbf{Task 3 (Edit)}} \\
        \cmidrule(lr){2-3} \cmidrule(lr){4-5} \cmidrule(lr){6-7}
        & \textbf{Avg. Len.} & \textbf{Parse Error} & \textbf{Avg. Len.} & \textbf{Parse Error} & \textbf{Avg. Len.} & \textbf{Parse Error} \\
        \midrule
        \multicolumn{7}{c}{\textit{Proprietary Models}} \\
        \midrule
        GPT-5.2 & 206 & 0.00\% & 212 & 0.00\% & 212 & 0.00\% \\
        Gemini-3-Pro & 97 & 0.20\% & 95 & 0.00\% & 120 & 0.00\% \\
        Claude-Sonnet-4.5 & 999 & 10.27\% & 1755 & 0.00\% & 300 & 0.00\% \\
        Seed1.6-VL & 152 & 1.10\% & 139 & 0.00\% & 137 & 0.00\% \\
        \midrule
        \multicolumn{7}{c}{\textit{Open-Source Models (Thinking / RL-aligned)}} \\
        \midrule
        MiMo-VL-SFT & 937 & 18.25\% & 156 & 0.00\% & 164 & 0.00\% \\
        MiMo-VL-RL & 2593 & 61.81\% & 209 & 0.00\% & 196 & 0.00\% \\
        Qwen3-VL-30B & 146 & 0.00\% & 174 & 0.00\% & 176 & 0.00\% \\
        LLaVA-Critic-R1 & 108 & 3.39\% & 108 & 0.00\% & 156 & 0.00\% \\
        ThinkLite-VL-7B & 214 & 0.00\% & 153 & 0.00\% & 187 & 0.00\% \\
        \midrule
        \multicolumn{7}{c}{\textit{Open-Source Models (Non-thinking)}} \\
        \midrule
        Qwen2.5-VL-72B & 170 & 0.00\% & 146 & 0.00\% & 191 & 0.00\% \\
        Qwen2.5-VL-7B & 97 & 7.78\% & 132 & 0.00\% & 126 & 0.00\% \\
        Qwen2-VL-72B & 145 & 0.00\% & 235 & 0.00\% & 274 & 0.00\% \\
        Qwen2-VL-7B & 140 & 1.30\% & 199 & 0.00\% & 228 & 0.00\% \\
        InternVL3.5-38B & 146 & 0.00\% & 456 & 0.00\% & 316 & 0.00\% \\
        InternVL3.5-8B & 2397 & 46.36\% & 220 & 0.00\% & 249 & 0.00\% \\
        InternVL2.5-38B & 144 & 0.00\% & 116 & 0.00\% & 179 & 0.00\% \\
        InternVL2.5-8B & 217 & 0.00\% & 188 & 0.00\% & 362 & 0.00\% \\
        Qwen3-VL-32B & 252 & 0.00\% & 238 & 0.00\% & 262 & 0.00\% \\
        Qwen3-VL-8B & 196 & 0.00\% & 193 & 0.00\% & 186 & 0.00\% \\
        DeepSeek-VL & 136 & 0.00\% & 270 & 0.00\% & 274 & 0.00\% \\
        GLM-4V & 1327 & 25.67\% & 233 & 0.00\% & 199 & 0.00\% \\
        Kimi-VL & 232 & 0.00\% & 249 & 0.00\% & 147 & 0.00\% \\
        Molmo & 148 & 0.00\% & 217 & 0.00\% & 207 & 0.00\% \\
        Molmo2 & 179 & 0.00\% & 197 & 0.00\% & 230 & 0.00\% \\
        MiMo-VL-SFT (NoThink) & 250 & 0.00\% & 214 & 0.00\% & 243 & 0.00\% \\
        MiMo-VL-RL (NoThink) & 266 & 0.00\% & 231 & 0.00\% & 255 & 0.00\% \\
        \bottomrule
    \end{tabular}
    }
\end{supplementtable}

Table~\ref{tab:inference_stats} reports the average output length (measured in characters) and the parse error rate across all three evaluation tasks. A response is strictly flagged as a parse error when the final verdict cannot be extracted. This primarily occurs due to two reasons: (1) the model fails to adhere to the strict output formatting instructions (e.g., missing the required decision tags), or (2) the model falls into excessive over-thinking loops, leading to premature sequence truncation (hitting the \texttt{max\_tokens} limit) before the final judgment is generated.

Notably, while an allocated \texttt{max\_tokens} limit of up to 1024 is theoretically more than sufficient for this judging task, our statistics reveal an intriguing phenomenon in the complex reasoning task (Task 1: CPA). Several RL post-trained thinking models (e.g., MiMo-VL-RL) exhibit exceptionally long outputs coupled with high parse error rates due to truncation. Furthermore, we observed that InternVL3.5-8B natively generated extensive \texttt{<think>} tags specifically on the CPA task, frequently becoming trapped in repetitive reasoning cycles. Consequently, we explicitly set its \texttt{max\_tokens} to 1024 to align with the evaluation settings of other deep-thinking models.

Our qualitative analysis reveals that these truncation-induced parse errors highlight a critical vulnerability in current RL post-trained thinking models. When handling complex chart evaluations (CPA), their advanced logical reasoning capabilities lack a sufficiently strong foundation in fine-grained visual perception. Without robust visual grounding to confidently anchor their deductions, these models spiral into endless, ungrounded thinking loops. 

Conversely, in the Chart Reproduction and Editing tasks (CRJ), the absence of long truncation does not imply improved competence. Instead, it exposes an entirely different failure mode: an \textit{unexpected lenience bias introduced by RL training}, as discussed in our main text. In these generation-verification scenarios, the models frequently bypass rigorous step-by-step visual reasoning. Driven by this inherent bias, they prematurely ``Accept'' the flawed charts without conducting the necessary logical deduction, thereby outputting significantly shorter, yet fundamentally unreliable, verdicts. The striking contrast between endless over-thinking in CPA and superficial premature acceptance in CRJ further confirms the severe misalignment between their sophisticated reasoning mechanisms and their visual perception limits.

\end{document}